\documentclass{article}
\usepackage{graphicx}
\usepackage{tikz}
\usetikzlibrary{positioning, arrows.meta, shapes.geometric, fit, backgrounds, calc}
\usepackage{xcolor}
\usepackage{float}
\usepackage{url}
\usepackage{longtable}
\usepackage{booktabs}
\usepackage{amsmath}
\usepackage{bbm}
\usepackage[most]{tcolorbox}
\usepackage{fvextra}
\usepackage{mathtools}
\usepackage{amssymb}
\usepackage{pgfplots}
\usepackage{adjustbox}
\usepackage{placeins}
\usepackage{needspace}
\usepgfplotslibrary{patchplots} 
\usepgfplotslibrary{statistics}
\pgfplotsset{compat=1.18}
\usepackage{pgfplotstable}
\usepgfplotslibrary{polar}
\usetikzlibrary{patterns,matrix,shapes.arrows}
\usepackage{caption}
\usepackage{subcaption}

\providecolor{radarDream}{HTML}{76B900}  
\providecolor{radarPi}{HTML}{F5A623}     
\providecolor{radarN17}{HTML}{3B82F6}    
\providecolor{radarMolmo}{HTML}{E74C3C}  
\providecolor{radarN16}{HTML}{8B5CF6}    

\definecolor{componentPrompt}{HTML}{DCEAF7}
\definecolor{componentGrade}{HTML}{D9F0EC}
\definecolor{componentArena}{HTML}{F5E8C8}
\definecolor{componentNeutral}{HTML}{F4F5F6}
\definecolor{componentStroke}{HTML}{59636E}

\tikzset{
  component/.style={
    draw=componentStroke,
    rounded corners=2pt,
    line width=0.6pt,
    align=center,
    font=\sffamily\small,
    inner sep=4pt
  },
  flowarrow/.style={->,>=latex,draw=componentStroke,line width=0.7pt}
}
\pgfplotsset{
  paperplot/.style={
    tick label style={font=\sffamily\scriptsize},
    label style={font=\sffamily\small},
    legend style={font=\sffamily\scriptsize},
    grid style={draw=gray!18,line width=0.3pt},
    axis line style={draw=gray!65,line width=0.4pt},
    tick style={draw=gray!65}
  }
}

\newcommand{\systemname}{\textsc{ScaleEvals}}

\usepackage{marginnote}
\newif\ifshowcomments
\showcommentsfalse

\newif\ifhighlightrev
\highlightrevfalse
\newcommand{\revised}[1]{\ifhighlightrev{\color{blue}#1}\else#1\fi}

\usepackage[final]{corl_2026} 

\hypersetup{
  pdftitle={ScaleEvals: Scaling Generalist Policy Evaluation through Reset-Free Autonomous Play},
  pdfauthor={Amitoj Sandhu, Ethan Sick, Omkaar Buddhikot, Nadia Laswi, Wesley Durbano, Nadun Ranawaka Arachchige, You Liang Tan},
  pdfsubject={Proceedings of the 10th Conference on Robot Learning (CoRL 2026)}
}

\title{ScaleEvals: Scaling Generalist Policy Evaluation through Reset-Free Autonomous Play}

\author{
  Amitoj Sandhu, Ethan Sick, Omkaar Buddhikot, Nadia Laswi, Wesley Durbano, \\
  \textbf{Nadun Ranawaka$^{\dagger}$, You Liang Tan$^{\dagger}$} \\
  {\scriptsize NVIDIA, $^{\dagger}$Project Leads}
}

\begin{document}

\maketitle

\begin{figure}[htbp]
  \centering
  \includegraphics[width=0.92\linewidth]{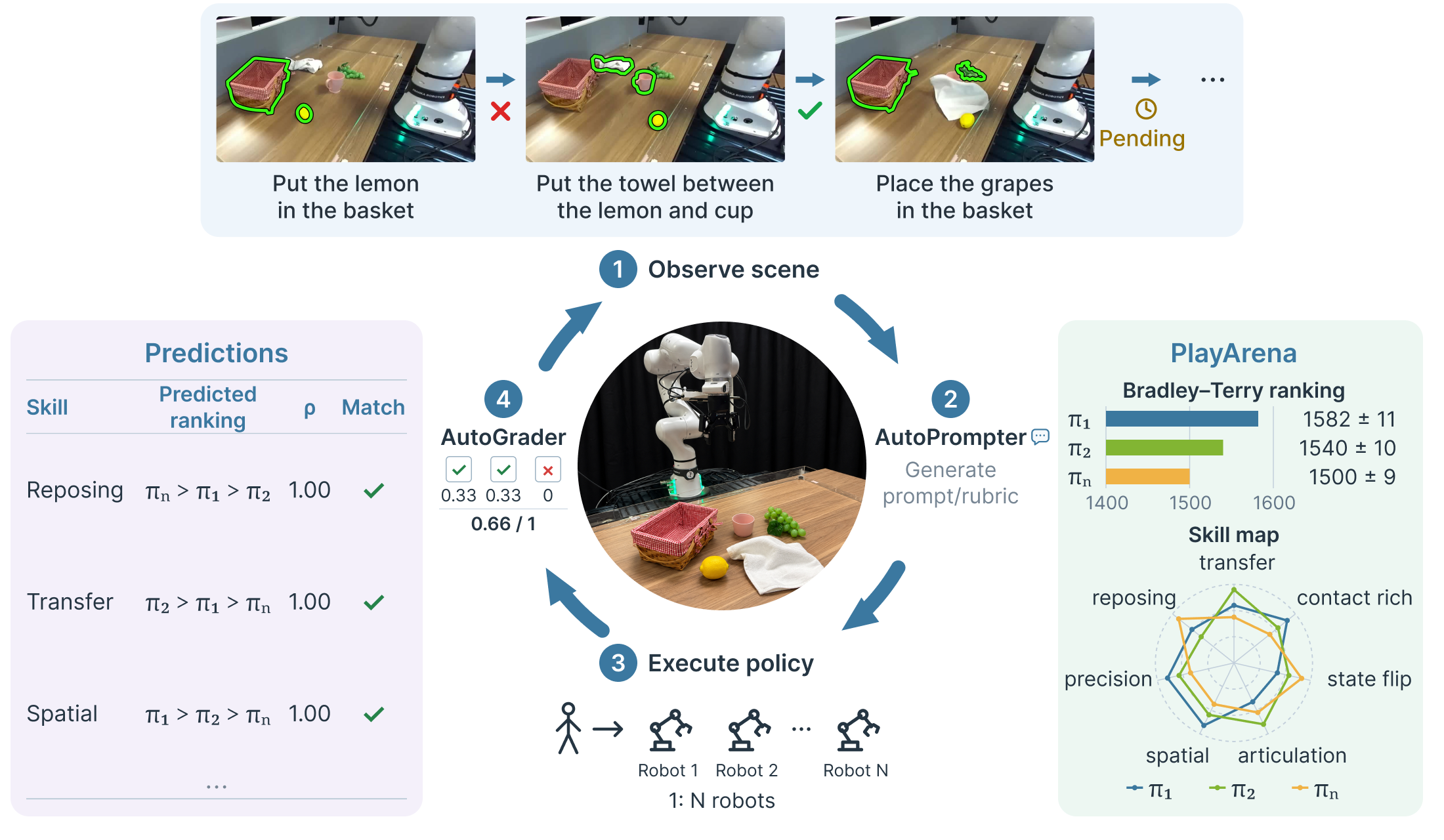}
  \caption{\textbf{Reset-free continuous-play evaluation.} AutoPrompter generates grounded tasks and rubrics from continuously evolving scene, the policy executes, and AutoGrader scores the rollouts. Aggregated performance yields policy rankings and reveals skill-conditioned diagnostics.}
\end{figure}

\begin{abstract}
Real-world evaluation of generalist robot policies is bottlenecked by manual
resets, human grading, and small fixed task suites with limited diagnostic
coverage. We present \systemname{}, a scalable reset-free protocol that evaluates
policies through continuous physical play across a robot fleet. A VLM AutoPrompter generates scene-grounded tasks with difficulty estimates, language prompts, and skill tags, enabling performance analysis across diverse rollouts. A VLM AutoGrader assesses task progress, achieving 90.7\% agreement with human graders. Scores are aggregated via a soft-win Bradley–Terry model, while tag-stratified ANOVA provides interpretable capability diagnostics.
Across 3,000+ physical autonomous rollouts on DROID and YAM, \systemname{} achieves
23.0 net rollouts/hr per station while reducing active human labor by 98\%. By ranking five SOTA generalist policies,
\systemname{} correlates with held-out manual evaluation ranking on DROID
(\(r{=}0.94,\rho{=}0.90,p{=}0.018\)), and remains internally consistent on a
second embodiment, YAM (\(r{=}0.839,\rho{=}0.80\)).
Beyond aggregate rankings, we observe \emph{behavioral compositionality}:
recombining verb-, skill-, and object-conditioned evidence predicts
policy order on unseen prompts. Across 10 tasks and 240 prospective
physical rollouts, we recover three-policy orderings of 96\% pairwise concordance, demonstrating \systemname{} can predict not just which policy leads, but where rankings reverse.

\textbf{Project:} \href{https://scaleevals.github.io/}{scaleevals.github.io}
\end{abstract}

\keywords{Generalist Robot Policy Evaluation, Scalability, Benchmark}

\section{Introduction}

Real-world evaluation is a bottleneck for generalist robot policy development. 
Current physical evaluation protocols typically rely on fixed task suites, manual scene resets, and human grading, making them costly to scale and limiting feedback to sparse success rates. 
Simulation-based evaluation provides a scalable alternative, but sim-to-real correlation can depend on embodiment, camera setup, task distribution, and contact regime~\cite{Li2024,yang2026robolabhighfidelitysimulationbenchmark}. 
Physical evaluation therefore remains essential, especially for manipulation tasks involving clutter, contact-rich interaction, articulated objects, and evolving scene state~\cite{Zhou2025}.

Beyond throughput, current evaluation protocols provide limited diagnostic resolution. 
Aggregate task progress and success rates can obscure whether a policy fails because of spatial reasoning, precision, contact-rich interaction, object state changes, or long-horizon sequencing. 
Fixed task suites may also overrepresent particular objects, relations, language forms, or interaction modes. 
Recent diagnostic benchmarks such as INT-ACT~\cite{Fang2025} highlight this issue by showing that VLA policies can often infer the correct high-level intent while failing to execute it reliably.

We propose \systemname{}, a reset-free framework for scalable real-world evaluation of generalist robot policies. 
Instead of evaluating policies through isolated reset-required trials, \systemname{} runs each robot station as a continuous play stream: tasks are generated from the current scene, policies act in the evolving environment, and outcomes are automatically graded and aggregated. 
This turns real-world evaluation from a sequence of manually reset benchmark trials into an autonomous evaluation loop. Summary of contributions:

\begin{enumerate}
  \item \textbf{Continuous play as evaluation:} \systemname{} generates scene-aware tasks online, with each rollout starting from the scene state left by previous rollouts rather than from a manually reset configuration. Unlike reset-required suites that control initial states, ScaleEvals evaluates language-conditioned steering through evolving scenes. It complements rather than replaces controlled spatial- and initial-state-generalization tests.

  \item \textbf{Automated task generation and grading:} A VLM-based AutoPrompter generates grounded instructions, milestone rubrics, difficulty annotations, and skill tags. A VLM-based AutoGrader scores rollout videos at subtask granularity, producing task-progress estimates without task-specific fine-tuning.

  \item \textbf{Matched ranking and tag-conditioned diagnostics:} PlayArena matches heterogeneous rollouts by difficulty and task--scene similarity, aggregates progress-derived pairwise comparisons with a soft-win Bradley--Terry (BT) model, and uses tag-stratified analyses to expose relative policy strengths and weaknesses across skill axes.
\end{enumerate}

\begin{figure}[t]
  \centering
  \includegraphics[width=\linewidth]{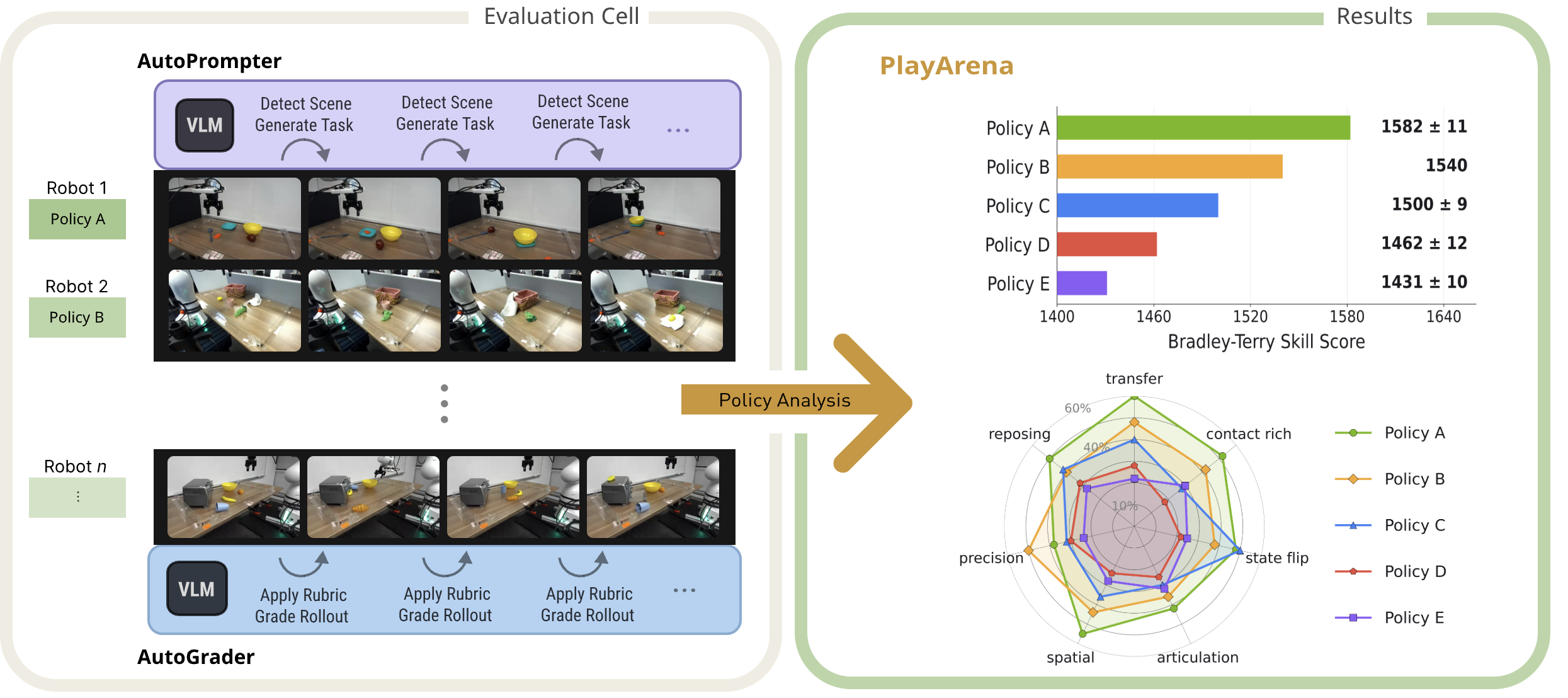}
  \caption{\textbf{Deployment and aggregation architecture.} Each evaluation
  cell runs policy streams through AutoPrompter and AutoGrader. PlayArena
  accumulates matched rollout evidence across cells to produce a global
  leaderboard and capability profiles.}
  \label{fig:SystemOverview}
\end{figure}

Across five generalist policies and more than 3,000 real-world trajectories on DROID and YAM, \systemname{} achieves 23.0 net rollouts/h per station compared with 14.8 rollouts/h for the manual baseline, while reducing active human involvement from 4 hours to 5 minutes per 100 rollouts and maintaining 90.7\% AutoGrader agreement with human rubric judgments.
BT rankings correlate strongly with manual-evaluation mean task progress (Pearson \(r{=}0.94,p{<}0.05\)). Skill-tag-conditioned BT rankings support pre-evaluation prompt generation and rank prediction (Spearman \(\rho{=}0.88\) before AutoGrader audit). \systemname{} collects comparable signal at higher net throughput while reducing active human time by 98\%, and scales across embodiments. Finally, \systemname{} reveals how fixed manual suites can confound policy assessment.
These results demonstrate that a reset-free paradigm facilitates scale as an autonomous, continuously running system that reliably assesses generalizability and produces interpretable policy diagnostics.

\section{Related Work}

\paragraph{Simulation-based evaluation.} 
Simulation-based benchmarks improve evaluation throughput by replacing costly physical rollouts with reproducible simulated trials. Systems such as SIMPLER, RoboLab, and PolaRiS report encouraging sim-real alignment across manipulation settings, IsaacLab task suites, and neural scene reconstructions \cite{Li2024,yang2026robolabhighfidelitysimulationbenchmark, Jain2025}. Diagnostic probing suites such as INT-ACT further show that high-level intent understanding can outpace execution reliability \cite{Fang2025}. World-model approaches such as Ctrl-World and Interactive World Simulator provide controllable environments for policy evaluation and training \cite{Guo2026,Wang2026}. Despite these gains, transfer correlations are sensitive to embodiment, sensor setup, and contact dynamics; even small physics mismatches can change behavior, especially for contact-rich, articulated, or deformable interactions. Prior work (e.g., \textit{Genesis}) reports strong but still bounded sim-to-real agreement \cite{Genesis}. World models can also accumulate rollout drift during multi-task, contact-rich, occluded, or long-horizon interactions. We therefore treat simulation and world models as useful proxies, while focusing on scalable real-world evaluation as the direct measure of physical generalization.
\systemname{} targets this gap by scaling physical evaluation itself: it continuously probes policies in evolving real scenes without manual resets or per-rollout human intervention.

\paragraph{Autonomous play and Self Improvement.}
Predictive proxies of post-training performance (e.g., low validation MSE) are often unreliable in real-world deployment \cite{Hu2025}. Autonomous play has therefore become a practical route for collecting diverse interaction experience and improving policies. PlayWorld shows that play-derived dynamics improve downstream RL performance \cite{Yin2026}, and recent work demonstrates autonomous improvement of instruction-following skills through foundation-model-driven loops \cite{Zhou2024}. Related efforts scale egocentric robot-data collection and online post-training across tasks, environments, and robot fleets \cite{Punamiya2026,Zheng2026,Pan2026}. In our setting, we use autonomous play for evaluation rather than policy updates: continuously evolving scenes generate broad behavior coverage, while skill tags provide a structured basis for fine-grained diagnostics.

\paragraph{Automated evaluations.} Real-world evaluations require costly human assessment and intervention. Existing automated methodologies include comparing final observation to a goal-state image using visual similarity or fine-tuning a VLM per task for success classification \cite{Zhou2025, Liang2026}. Reward modeling methods such as Robometer leverage a fine-tuned VLM that runs inference concurrently to output continuous task progress in scalar form \cite{liang2026robometer}, which can be opaque in interpretability, computationally intensive, and susceptible to out-of-distribution trajectories. Recent work has highlighted the importance of graduating from binary success rates into partial completions through deploying cost-aware, VLM-based autograders proposed as an open source effort \cite{ElMallah2025}.
Recent autonomous evaluation systems, such as AutoEval, reduce human intervention but still rely on per-task models and reset policies for binary success detection \cite{Zhou2025}. Benchmarks such as RoboChallenge scale via API access, but remain tied to fixed reset-required suites \cite{Yakefu2025}. Fixed suites are useful for controlled comparison, yet are more susceptible to benchmark-specific fitting; continuous reset-free streams do not remove distributional bias, but can reduce opportunities for memorization and improve coverage of naturally evolving scene states.

\paragraph{Pairwise ranking and preference-based evaluation.}
Pairwise-comparison models such as Bradley--Terry provide a principled way to
aggregate incomplete and imbalanced comparisons into global rankings~\cite{Bradley1952,Liusie2024}.
They have been widely used in preference-based evaluation and robot learning, including
trajectory-preference optimization~\cite{liang2026robometer} and large-scale policy leaderboards
such as RoboArena~\cite{Atreya2025}. These methods are attractive for robot evaluation
because policies need not be evaluated on identical fixed suites, and pairwise outcomes
can be pooled across diverse trials. However, existing leaderboard-style rankings primarily
summarize aggregate capability: they say which policy is stronger overall, but can take time to converge into stable rankings and provide limited insight into which interaction modes, language factors, or physical skills drive the difference. As seen with LLM's, performance can exhibit a "jagged frontier" across capabilities~\cite{sun2026}. 

A/B testing is effective when policies are evaluated on shared, controlled task distributions~\cite{Snyder2025}, but reset-free
streams produce different prompts and evolving scene states,
making direct comparisons vulnerable to task-difficulty confounding.
Exhaustive pairwise testing of \(K\) policies also requires
\(K(K-1)/2\) comparisons. \systemname{} instead matches
heterogeneous rollouts by task--scene similarity, skill tags,
and estimated difficulty, then aggregates weighted soft wins
through Bradley--Terry~\cite{Bradley1952}. This produces a
joint policy ranking without identical prompts, while reducing,
but not eliminating, task-distribution confounding. Similarly, while world model simulators attempt to replace physical execution, world models suffer from compounding rollout drift during contact-rich interactions, occlusion, and long-horizon retries, making physical play-at-scale a valuable diagnostic methodology. \revised{\textsc{ScaleEvals} is complementary to leaderboard efforts such as RoboArena~\cite{Atreya2025}: it serves as a local evaluation engine for testing in-development checkpoints against RoboArena-listed open-source policies while minimizing human effort, helping practitioners iterate toward more competitive submissions before committing to a distributed leaderboard evaluation.}

\systemname{} builds on this ranking view but adapts it to reset-free, continuously generated
real-world evaluations. Instead of requiring exact task matches, PlayArena forms weighted
comparisons between semantically related rollouts using task, scene, difficulty, and skill-tag
metadata. We convert AutoGrader progress scores into soft pairwise outcomes and fit a
weighted Bradley--Terry model to estimate global policy strengths. To move beyond a single
leaderboard number, we further stratify rollouts by skill tag and use one-way analysis of variance (ANOVA) to test
whether policy identity explains task-progress variation within each capability slice. This
combination yields both a global ranking and interpretable diagnostics over skills such as
spatial reasoning, precision, articulation, contact-rich manipulation, and reposing.

\section{ScaleEvals Overview}

\subsection{AutoPrompter: Continuous-Play Task Generation}
\label{sec:autoprompter}

Fixed benchmarks cannot cover the diversity of scenes, object states, and
manipulation skills encountered by generalist policies. \systemname{} addresses
this with AutoPrompter, a few-shot VLM-based task generator that converts the
current scene into a grounded instruction, grading rubric, and task metadata.
At each rollout, AutoPrompter detects visible objects and proposes a feasible
manipulation task for the current scene. It records the motion verb, primary
object, estimated difficulty, and primary/secondary skill tags, and generates a
milestone rubric of yes/no visual checks for AutoGrader. After the policy acts,
AutoPrompter conditions on the updated scene and task history to generate the
next instruction, enabling continuous reset-free evaluation. These metadata make generated tasks auditable and comparable across policies. PlayArena uses them to match related rollouts with estimated difficulty and analyze performance by capability (Additional details in Sec.~\ref{sec:LLMdetails}).

\revised{Since tasks are grounded in the objects present, scene composition directly drives behavioral variety: across 2{,}486 rollouts, AutoPrompter produced 26 distinct action verbs over 32 objects, and the objects in-scene largely determine which skills are exercised (e.g., a drawer elicits articulation). The resulting prompts span all seven capability tags---state flip (38\%), spatial reasoning (31\%), transfer (28\%), articulation (18\%), precision (13\%), contact-rich manipulation (12\%), and reposing (11\%); percentages exceed 100\% because a single task may carry multiple tags. Operators can therefore increase variety simply by swapping objects into the scene, while keeping 5--7 props per scene, which we find optimal for throughput and policy performance (excess distractors degrade performance). Candidate tasks are executed only after automated checks for object visibility and safety, and failing tasks are regenerated automatically without human filtering.}
AutoPrompter is configurable through the scene object inventory, safe-to-lift and
heavy-object constraints, recent task history, few-shot examples, and permitted
action/task rules. Users can therefore shift the generated task distribution
without model fine-tuning by changing scene contents or these prompt-level controls.
Appendix Figure~\ref{fig:autoprompter-output-distribution} visualizes the
multi-label skill-tag distribution of accepted prompts.

\subsection{AutoGrader: Completion Progress from Multimodal Evidence}
\label{sec:autograder}

\systemname{} uses the off-the-shelf VLM Gemini 3 Flash as a cost-effective, fine-tune-free AutoGrader. We also evaluated Nemotron and Gemini Pro backbones and selected Gemini 3 Flash for its balance of criterion-level accuracy, macro F1, and cost. Per trajectory, the frame rate (fps), batch size, and nth frame extraction can be configured prior to submitting an array of frames to the VLM. We group frames into temporally ordered batches, which are passed to the VLM with explicit batch identifiers. This provides a structured representation for temporal reasoning (e.g., referencing when an action occurs) and improves auditability by allowing graders to localize decisions to specific segments of the trajectory. Through a Visual Question Answering (VQA) style of subtask grading provided by Autoprompter, the VLM must answer yes/no per subtask. For example, for a ``Place the marker in a cup task" the AutoPrompter will generate subtasks in VQA format such as i)\textit{ Did the robot grasp the marker?} ii) \textit{Did the robot pick up the marker?} iii) \textit{Was the marker placed in the cup?} for a normalized score ranging [0,1]. We have found 3 fps, batch size of 10 frames, sending every other frame, preserves $90.7\%$ accuracy at the criterion level.

\subsection{PlayArena: Matched Ranking and Capability Diagnostics}
\label{sec:bt}

PlayArena aggregates reset-free rollout streams into policy rankings and skill-level diagnostics. Since AutoPrompter generates tasks online rather than replaying a fixed suite, policies may encounter different but related prompts. For each pair of policies, we construct a bipartite similarity matrix over their rollouts using instruction difficulty and hybrid task--scene similarity. The score favors shared task factors, including motion verbs and skill tags, and shared scene factors, including the primary manipulated object. We then apply the Hungarian algorithm to obtain the globally optimal one-to-one assignment under this similarity score, rather than greedily pairing each rollout with its nearest match. Higher-quality matches receive larger weights in the downstream Bradley--Terry ranking. Matching is fully automated and requires no manual pairing. Empirically, above 100 rollouts per policy, the induced rank order is insensitive to the matching hyperparameters, and 100\% of rollouts obtain at least one match.

Let \(m\in\mathcal{M}\) index a matched comparison between policies \(\pi_i\) and \(\pi_j\), and let \(P_i^{(m)},P_j^{(m)}\in[0,100]\) be their AutoGrader task-progress percentages. We convert each matched comparison into a fractional preference target,
\[
y_{ij}^{(m)}
=
\frac{P_i^{(m)}+\epsilon}
     {P_i^{(m)}+P_j^{(m)}+2\epsilon},
\qquad
\epsilon=0.01 .
\]
This target assigns more preference mass to the policy with higher task progress, maps equal progress to a tie-like value near \(0.5\), and handles mutual-zero comparisons. Since \(P\) is measured in percentage points, \(\epsilon\) acts only as a numerical guard.

We fit policy strengths with a weighted soft-win Bradley--Terry model~\cite{Bradley1952}:
\[
\mathcal{L}(\theta)
=
-\sum_{m\in\mathcal{M}}
\omega_m
\left[
y_{ij}^{(m)}\log \sigma(\theta_i-\theta_j)
+
(1-y_{ij}^{(m)})
\log\!\left(1-\sigma(\theta_i-\theta_j)\right)
\right]
+
\lambda \lVert \theta\rVert_2^2 .
\]
Here, \(\mathcal{M}\) is the set of matched comparisons, \(\omega_m\) is the match weight, \(\lambda\) is the \(L_2\) regularization coefficient, and \(i,j\) denote the two policies compared in match \(m\). We report the resulting strengths on an Elo-like scale with a fixed anchor policy fixed to reduce ambiguity of Bradley--Terry scores.

For diagnostics, PlayArena stratifies task progress by skill tag and runs one-way ANOVA within each tag to test whether policy identity explains variation in task progress. We report \(F\)-statistics, \(p\)-values, Benjamini--Hochberg FDR-corrected significance, and effect size \(\eta^2\). We analyze seven primary skill tags: \textit{transfer, spatial reasoning, state flip, articulation, precision, contact-rich manipulation, and reposing}. Because prompts can contain multiple tags, a single rollout can contribute to multiple capability slices. Thus, PlayArena produces both a global policy ranking and tag-conditioned capability profiles.

\section{Experiments}

Our experiments serve to answer the following: (1)\textit{ How do reset-free play-based evaluations and rankings compare to traditional, human and reset-required approaches?} (2) \textit{How sample-efficient is \systemname{}?} (3)\textit{ Can \systemname{} enable diagnostic insights and predictive qualities through classical statistics beyond performance ranking?}

\subsection{Experimental Setup}



\paragraph{Policies.}
We compare five generalist policies: GR00T N1.6, GR00T N1.7, DreamZero, MolmoAct2, and $\pi_{0.5}$ \cite{2025pi05visionlanguageactionmodelopenworld, gr00tn1openfoundation, Ye2026, duan2026molmoact2}. This selection emulates academic labs comparing the generalizability of policy checkpoints. Across independent \systemname{} sessions, each policy receives identical initial scene setup and time budgets to limit computational advantage. All policies are evaluated on the DROID robot platform \cite{Khazatsky2025}.

\paragraph{Scenes and Setups.}
We evaluate in tabletop and room-scale scenes with varying object sets (\revised{5--7} per scene). To minimize human intervention, acrylic barriers confine the objects to the area navigable by the robot. Remote intervention, autonomous monitoring, workspace exclusion zones, and velocity/force thresholds constrain execution. Should the Franka arms encounter hardware issues, we implemented a slack-bot to notify the operator for intervention. Backgrounds can vary above cubicle dividers, the introduced noise serves as a proxy for real-world scene variance, counterbalanced by significant samples within-policy.

\begin{figure}[htbp]
  \centering
  \includegraphics[width=\linewidth]{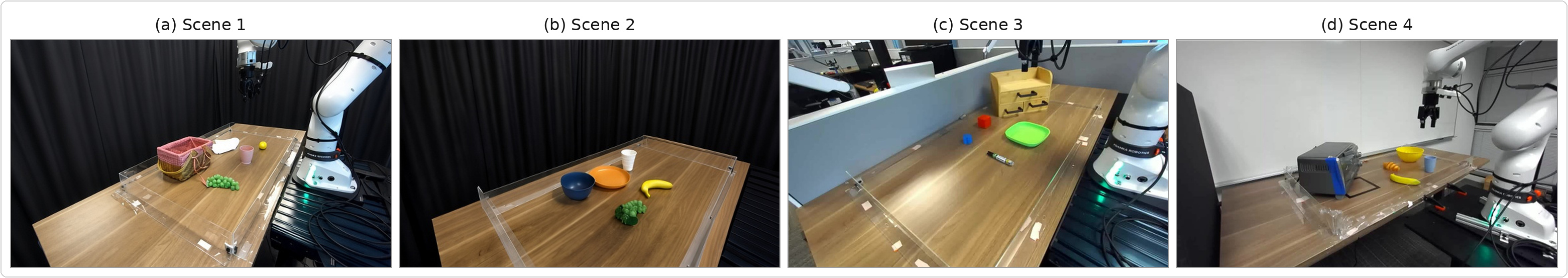}
  \caption{\textbf{Evaluation scenes.} Four tabletop and room-scale setups with
  different object sets. Panels show the original evaluation setups and object configurations used for physical rollouts.}
  \label{fig:evaluation-scenes}
\end{figure}



\Needspace{10\baselineskip}
\subsection{Results}
\paragraph{AutoGrader.}
We audit AutoGrader on 525 randomly sampled trajectories, corresponding to
1,454 rubric-level VQA judgments sampled across difficulty buckets and skill
tags. AutoGrader achieves 90.7\% criterion-level agreement with human graders,
supporting its use as a scalable measurement primitive with subtask resolution.
Residual errors skew toward false positives under depth ambiguity, occlusion,
and uncertain contact state; Figure~\ref{fig:graderHeatmap} and
Section~\ref{sec:limitations} provide further analysis.
\FloatBarrier

%
%

\begin{table}[!t]
  \centering
  \small
  \setlength{\tabcolsep}{4pt}
  \renewcommand{\arraystretch}{1.02}
  \begin{tabular*}{\linewidth}{@{\extracolsep{\fill}}l c c c c c c@{}}
    \toprule
    & \multicolumn{4}{c}{\textbf{ScaleEvals ($n{=}350$ per policy)}}
    & \multicolumn{2}{c}{\textbf{Manual Eval Suite (up to $n{=}40$/policy)}} \\
    \cmidrule(lr){2-5} \cmidrule(lr){6-7}
    \textbf{Policy} & \textbf{$n$} & \textbf{BT}\,$\uparrow$
    & \textbf{TP\,\%}\,$\uparrow$ & \textbf{SR\,\%}\,$\uparrow$
    & \textbf{TP\,\%}\,$\uparrow$ & \textbf{SR\,\%}\,$\uparrow$ \\
    \midrule
    DreamZero             & 350 & \textbf{1580}\,$\pm$\,16 & \textbf{38.5} & \textbf{13.1}  & \textbf{54.4} & \textbf{31.6}  \\
    $\pi_{0.5}$ (anchor)  & 350 & 1539                     & 30.7          & \phantom{0}7.7 & 45.0          & 12.5           \\
    GR00T N1.7            & 350 & 1500\,$\pm$\,17          & 23.2          & \phantom{0}3.4 & 43.9          & 10.0           \\
    MolmoAct2             & 350 & 1456\,$\pm$\,16          & 17.3          & \phantom{0}2.6 & 27.0          & \phantom{0}5.0 \\
    GR00T N1.6            & 350 & 1453\,$\pm$\,16          & 15.5          & \phantom{0}0.6 & 34.3          & 12.5           \\
    \bottomrule
  \end{tabular*}
  \caption{\textbf{Policy performance under equal-$N$ ($n{=}350$ per policy).}
  \textbf{ScaleEvals} reports bootstrap-median ($K{=}200$) Bradley--Terry (BT), mean task progress (TP), and success rate (SR) on a per-skill-balanced experiment. $\pi_{0.5}$ is anchored at $\text{BT}{=}1539$ on the Elo scale ($400/\ln 10$; Hessian SE shown). \textbf{Manual Eval Suite} reports mean TP and SR on held-out human-graded reset-required tasks (up to $40$ rollouts per policy; $198$ observed rollouts, DreamZero lacks two trials for a task).}
  \label{tab:headline-leaderboard}
\end{table}

\begin{figure}[!t]
  \centering
  \begin{tikzpicture}
    \begin{axis}[
      width=0.72\linewidth, height=4.8cm,
      xlabel={Trajectories per policy $B$},
      ylabel={Bradley--Terry Elo (bootstrap median)},
      xmin=20, xmax=395,
      ymin=1370, ymax=1660,
      xtick={25,50,75,100,150,200,250,300,350},
      ytick={1400,1450,1500,1550,1600,1650},
      yticklabel style={/pgf/number format/1000 sep={}},
      grid=both,
      major grid style={dashed, gray!20, line width=0.3pt},
      axis line style={draw=black!60, line width=0.4pt},
      tick label style={font=\scriptsize},
      label style={font=\scriptsize},
      legend style={
        font=\scriptsize, draw=black!30,
        fill=white, fill opacity=0.92, text opacity=1,
        at={(1.02,1.00)}, anchor=north west,
        cells={anchor=west}, row sep=-1pt, inner sep=2pt,
      },
      legend cell align=left,
    ]

      \addplot[fill=radarDream, fill opacity=0.13, draw=none, forget plot] coordinates {
        (25,1453.1)  (50,1518.8)  (75,1553.5)  (100,1572.8) (125,1565.7) (150,1563.6)
        (175,1574.3) (200,1571.0) (225,1566.8) (250,1573.8) (275,1574.0) (300,1575.5)
        (325,1576.9) (350,1579.3) (379,1581.0)
        (379,1589.5) (350,1595.8) (325,1586.5) (300,1586.4) (275,1591.5) (250,1592.5)
        (225,1593.4) (200,1597.9) (175,1598.6) (150,1598.2) (125,1591.4) (100,1609.4)
        (75,1627.2)  (50,1630.3)  (25,1639.9)
      };
      \addplot[color=radarDream, mark=*, mark size=1.4pt, line width=1pt]
        coordinates {(25,1555.4) (50,1574.7) (75,1573.8) (100,1592.0) (125,1573.9)
                     (150,1572.9) (175,1585.9) (200,1585.7) (225,1580.3) (250,1586.9)
                     (275,1585.5) (300,1581.6) (325,1581.2) (350,1585.8) (379,1585.8)};
      \addlegendentry{DreamZero \ ($B^\dagger{=}50$)}

      \addplot[fill=radarPi, fill opacity=0.13, draw=none, forget plot] coordinates {
        (25,1452.2)  (50,1500.0)  (75,1500.0)  (100,1513.0) (125,1502.1) (150,1512.5)
        (175,1525.8) (200,1527.1) (225,1518.2) (250,1523.6) (275,1524.3) (300,1529.6)
        (325,1527.0) (350,1532.9) (379,1530.3)
        (379,1544.0) (350,1543.6) (325,1546.7) (300,1547.0) (275,1551.8) (250,1546.0)
        (225,1555.9) (200,1552.1) (175,1558.6) (150,1552.8) (125,1527.7) (100,1553.1)
        (75,1597.1)  (50,1600.1)  (25,1561.8)
      };
      \addplot[color=radarPi, mark=diamond*, mark size=1.6pt, line width=1pt]
        coordinates {(25,1500.0) (50,1525.7) (75,1553.8) (100,1536.1) (125,1517.0)
                     (150,1532.8) (175,1533.8) (200,1537.9) (225,1539.3) (250,1534.3)
                     (275,1536.8) (300,1537.5) (325,1535.8) (350,1538.4) (379,1538.1)};
      \addlegendentry{$\pi_{0.5}$ \ ($B^\dagger{=}50$)}

      \addplot[color=radarN17, mark=triangle*, mark size=1.5pt, line width=1pt]
        coordinates {(25,1500.0) (50,1500.0) (75,1500.0) (100,1500.0) (125,1500.0)
                     (150,1500.0) (175,1500.0) (200,1500.0) (225,1500.0) (250,1500.0)
                     (275,1500.0) (300,1500.0) (325,1500.0) (350,1500.0) (379,1500.0)};
      \addlegendentry{GR00T N1.7 \ ($B^\dagger{=}25$)}

      \addplot[fill=radarMolmo, fill opacity=0.13, draw=none, forget plot] coordinates {
        (25,1440.0)  (50,1391.4)  (75,1426.5)  (100,1450.1) (125,1436.4) (150,1449.7)
        (175,1445.3) (200,1454.0) (225,1448.2) (250,1451.8) (275,1447.1) (300,1454.8)
        (325,1451.7) (350,1453.4) (379,1454.8)
        (379,1465.8) (350,1470.5) (325,1463.4) (300,1466.0) (275,1467.5) (250,1471.5)
        (225,1469.5) (200,1479.3) (175,1479.6) (150,1478.5) (125,1469.2) (100,1485.1)
        (75,1492.0)  (50,1500.0)  (25,1514.3)
      };
      \addplot[color=radarMolmo, mark=pentagon*, mark size=1.7pt, line width=1pt]
        coordinates {(25,1500.0) (50,1446.0) (75,1458.2) (100,1466.3) (125,1447.3)
                     (150,1466.5) (175,1459.7) (200,1466.6) (225,1459.4) (250,1461.5)
                     (275,1457.4) (300,1458.8) (325,1457.3) (350,1459.3) (379,1461.6)};
      \addlegendentry{MolmoAct2 \ ($B^\dagger{=}50$)}

      \addplot[fill=radarN16, fill opacity=0.13, draw=none, forget plot] coordinates {
        (25,1371.7)  (50,1388.1)  (75,1438.4)  (100,1432.6) (125,1426.9) (150,1441.1)
        (175,1446.8) (200,1448.0) (225,1439.3) (250,1442.5) (275,1443.4) (300,1445.4)
        (325,1445.4) (350,1449.0) (379,1453.0)
        (379,1462.4) (350,1465.2) (325,1460.0) (300,1459.7) (275,1461.8) (250,1463.8)
        (225,1462.5) (200,1468.2) (175,1470.8) (150,1476.1) (125,1470.0) (100,1484.5)
        (75,1496.8)  (50,1500.0)  (25,1519.2)
      };
      \addplot[color=radarN16, mark=square*, mark size=1.3pt, line width=1pt]
        coordinates {(25,1499.8) (50,1461.5) (75,1460.2) (100,1463.9) (125,1440.4)
                     (150,1451.5) (175,1455.0) (200,1459.2) (225,1449.3) (250,1453.4)
                     (275,1454.8) (300,1454.0) (325,1454.3) (350,1457.9) (379,1457.5)};
      \addlegendentry{GR00T N1.6 \ ($B^\dagger{=}50$)}

      \addplot[black!55, line width=0.7pt, dashed, forget plot]
        coordinates {(100, 1370) (100, 1660)};
      \node[anchor=south, font=\tiny, text=black!75, align=center, fill=white, fill opacity=0.85, text opacity=1, inner sep=1.5pt]
        at (axis cs:100, 1655) {full ranking\\ stable from $B{=}100$};
    \end{axis}
  \end{tikzpicture}
  \caption{\revised{\textbf{ScaleEvals rank convergence (reset-free only).}
  Bradley--Terry Elo per policy versus rollout budget $B$ under per-tag-balanced sub-sampling, computed entirely from reset-free ScaleEvals play (no manual data). For this convergence visualization, every bootstrap fit is recentered to GR00T N1.7${=}1500$; this additive shift does not change ordering and differs from the $\pi_{0.5}{=}1539$ anchor used in the leaderboard. The policy ordering becomes stable by $B{=}100$ trajectories per policy, and subsequent estimates remain within approximately $\pm 25$ Elo.}}
  \label{fig:Comprehensive-convergence}
\end{figure}
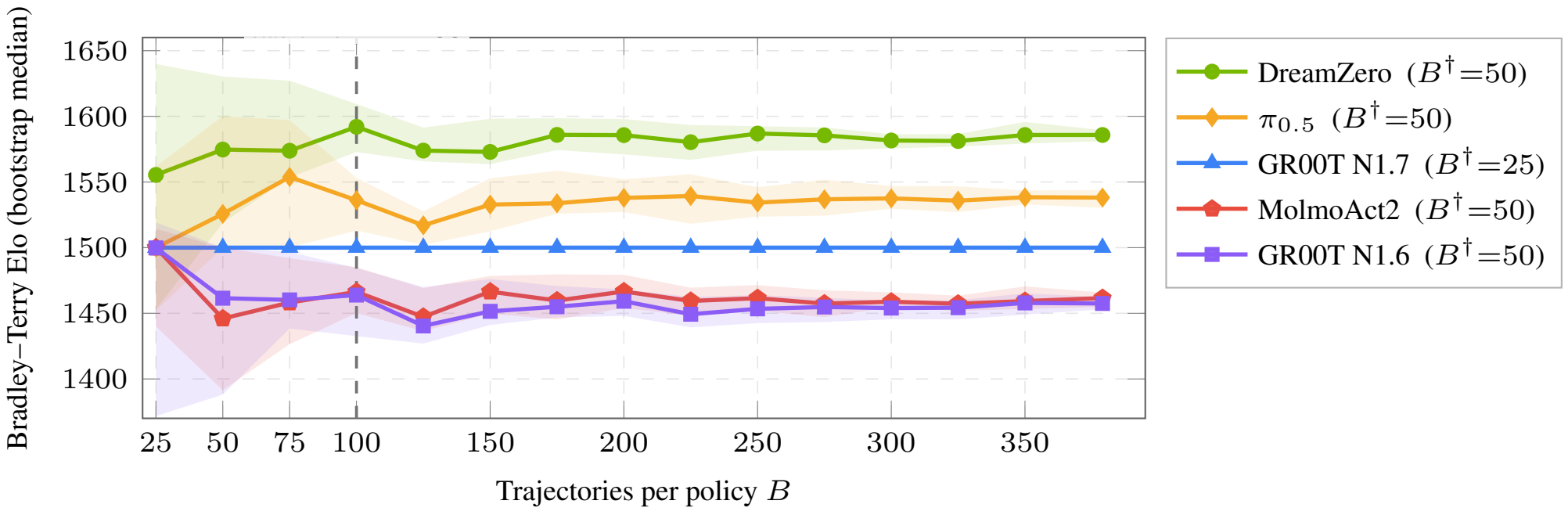

\begin{figure}[!t]
  \centering
  \pgfplotsset{
    /pgfplots/calibrationpanel/.style={
      width=0.95\linewidth, height=3.5cm,
      xlabel style={font=\scriptsize, yshift=1pt},
      ylabel style={font=\scriptsize, yshift=-1pt},
      title style={font=\scriptsize, yshift=-3pt, align=center},
      tick label style={font=\scriptsize},
      label style={font=\scriptsize},
      grid=both,
      major grid style={dashed, gray!20, line width=0.3pt},
      axis line style={draw=black!60, line width=0.4pt},
      tick style={draw=black!60},
      enlarge x limits=0.04, enlarge y limits=0.04,
    },
  }
  \begin{minipage}[t]{0.49\linewidth}
    \vspace{0pt}\centering
    \begin{tikzpicture}
      \begin{axis}[calibrationpanel,
        xlabel={Manual Eval Mean TP (\%)},
        ylabel={ScaleEvals BT Ranking (Elo)},
        title={ \ \ $r{=}0.94,\,p{=}0.018$},
        xmin=20, xmax=65, ymin=1430, ymax=1620,
        xtick={20,30,40,50,60},
        ytick={1450,1500,1550,1600},
        yticklabel style={/pgf/number format/1000 sep={}},
      ]
        \addplot[black!75, line width=0.7pt, forget plot]
          coordinates {(20, 1403.98) (65, 1622.66)};
        \addplot[only marks, mark=*, mark size=3pt,
          color=radarDream, mark options={fill=radarDream,draw=black,line width=0.3pt}]
          coordinates {(54.4, 1580.3)};
        \addplot[only marks, mark=diamond*, mark size=3.5pt,
          color=radarPi, mark options={fill=radarPi,draw=black,line width=0.3pt}]
          coordinates {(45.0, 1538.8)};
        \addplot[only marks, mark=triangle*, mark size=3.5pt,
          color=radarN17, mark options={fill=radarN17,draw=black,line width=0.3pt}]
          coordinates {(43.9, 1500.0)};
        \addplot[only marks, mark=square*, mark size=3pt,
          color=radarN16, mark options={fill=radarN16,draw=black,line width=0.3pt}]
          coordinates {(34.3, 1452.8)};
        \addplot[only marks, mark=pentagon*, mark size=4pt,
          color=radarMolmo, mark options={fill=radarMolmo,draw=black,line width=0.5pt}]
          coordinates {(27.0, 1456.3)};
      \end{axis}
    \end{tikzpicture}
  \end{minipage}\hfill
  \begin{minipage}[t]{0.49\linewidth}
    \vspace{0pt}\centering
    \begin{tikzpicture}
      \begin{axis}[calibrationpanel,
        xlabel={Manual Eval Mean TP (\%)},
        ylabel={ScaleEvals Mean TP (\%)},
        title={\ $r{=}0.92,\,p{=}0.029$},
        xmin=20, xmax=65, ymin=12, ymax=42,
        xtick={20,30,40,50,60},
        ytick={15,20,25,30,35,40},
      ]
        \addplot[black!75, line width=0.7pt, forget plot]
          coordinates {(20, 7.68) (65, 45.01)};
        \addplot[only marks, mark=*, mark size=3pt,
          color=radarDream, mark options={fill=radarDream,draw=black,line width=0.3pt}]
          coordinates {(54.4, 38.5)};
        \addplot[only marks, mark=diamond*, mark size=3.5pt,
          color=radarPi, mark options={fill=radarPi,draw=black,line width=0.3pt}]
          coordinates {(45.0, 30.7)};
        \addplot[only marks, mark=triangle*, mark size=3.5pt,
          color=radarN17, mark options={fill=radarN17,draw=black,line width=0.3pt}]
          coordinates {(43.9, 23.2)};
        \addplot[only marks, mark=square*, mark size=3pt,
          color=radarN16, mark options={fill=radarN16,draw=black,line width=0.3pt}]
          coordinates {(34.3, 15.5)};
        \addplot[only marks, mark=pentagon*, mark size=4pt,
          color=radarMolmo, mark options={fill=radarMolmo,draw=black,line width=0.5pt}]
          coordinates {(27.0, 17.3)};
      \end{axis}
    \end{tikzpicture}
  \end{minipage}
  \caption{\revised{\textbf{ScaleEvals task-progress signal agrees with the held-out manual suite.}
  Reset-free ScaleEvals plotted against the held-out manual evaluation suite ($n{=}40$ per policy). \emph{Left}: manual mean TP versus ScaleEvals BT rank ($r{=}0.94,\,p{=}0.018$). \emph{Right}: manual mean TP versus raw ScaleEvals mean TP ($r{=}0.92,\,p{=}0.029$). Reset-free measurements preserve the relative policy signal of reset-required evaluation while using far broader task coverage. Task-level manual progress and initial frames are reported in Table~\ref{tab:F6-detailed}.}}
  \label{fig:Comprehensive-calibration}
\end{figure}
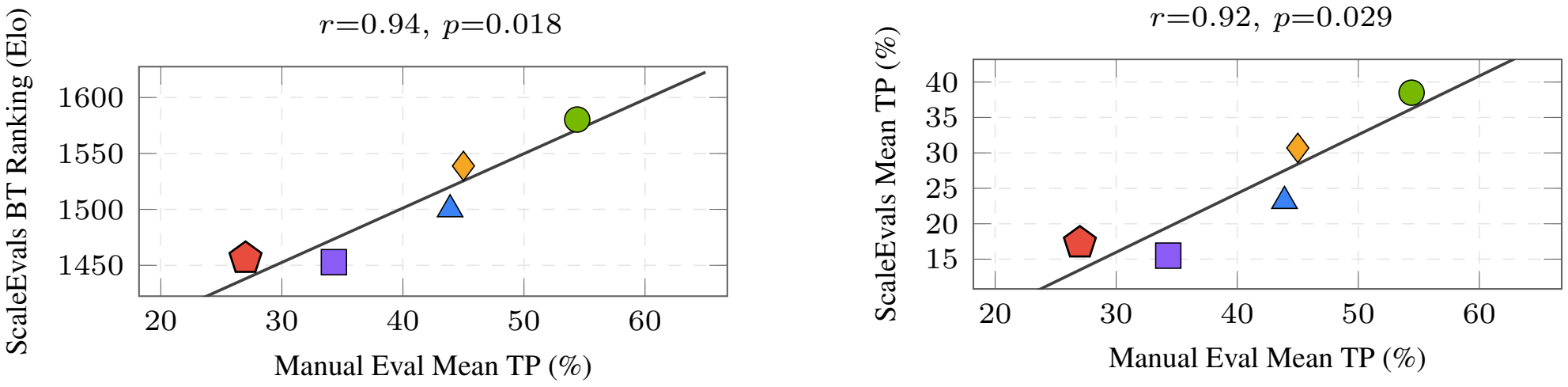

\begin{figure}[!t]
  \centering
  \includegraphics[width=\linewidth]{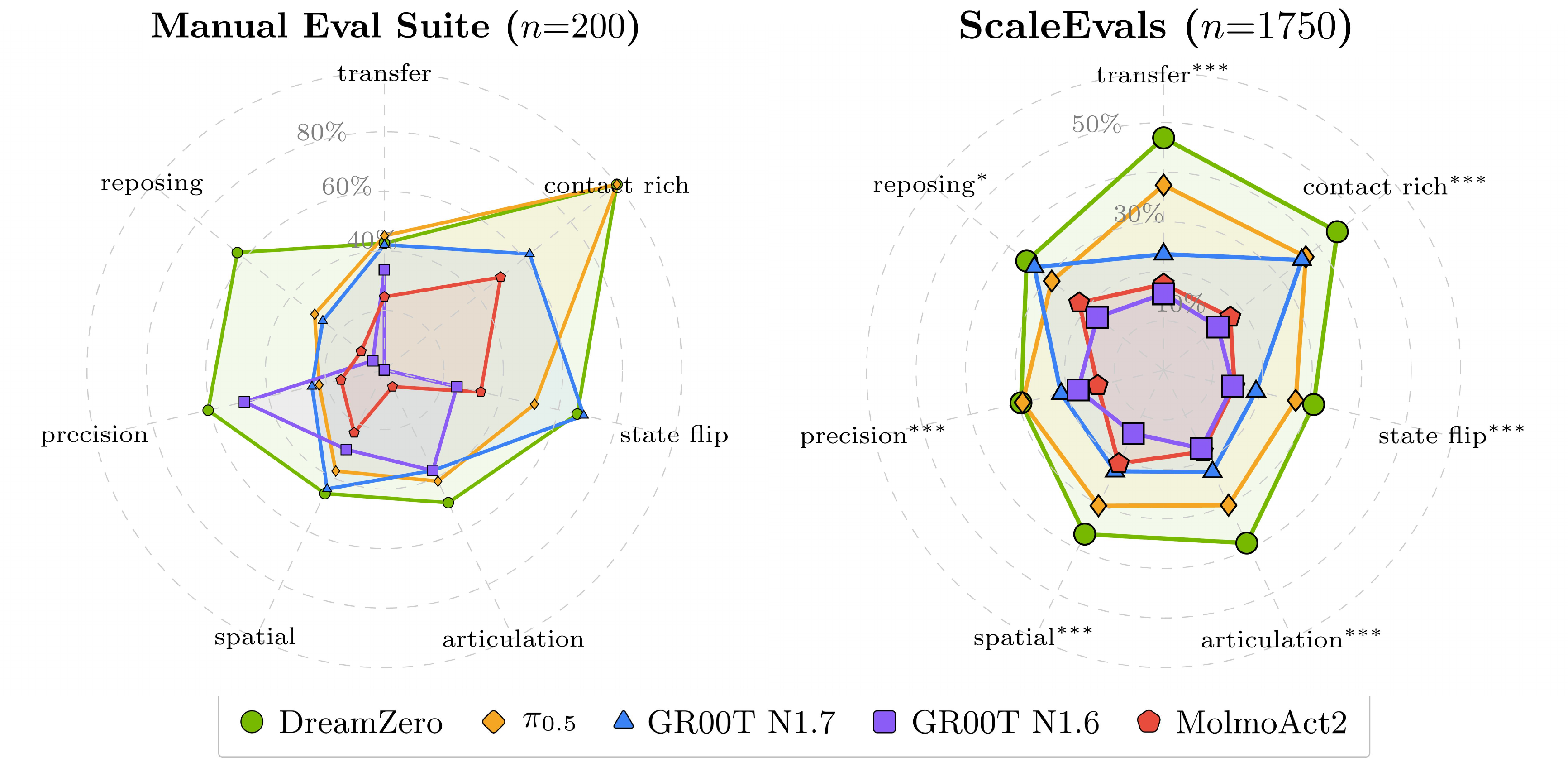}
  \caption{\revised{\textbf{Skill-tag-conditioned capability diagnostics.}
  Per-tag mean task-progress profiles for each policy, from reset-free ScaleEvals and the held-out manual suite (up to $40$ rollouts per policy). The panels use different radial ranges, shown on their grids; policy ordering within each panel is maintained. The tag decomposition localizes \emph{where} policies differ, while coverage gaps in the small manual suite appear as axes the reset-free stream fills in.}}
  \label{fig:BothRadars}
\end{figure}

\paragraph{Skill-tag profile agreement.}
Figure~\ref{fig:BothRadars} compares per-tag capability profiles between reset-free ScaleEvals and held-out manual evaluation. The radar view highlights where the two protocols preserve the same relative ordering signal across skill tags and divergence due to coverage differences.

\begin{figure*}[!t]
  \centering
  \includegraphics[width=0.88\linewidth]{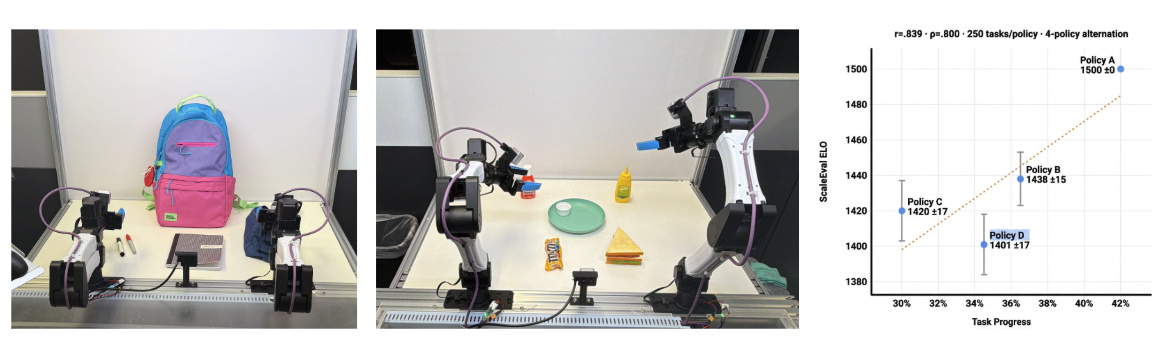}
  \caption{\revised{\textbf{Cross-embodiment evaluation on the YAM bimanual platform.}
  \emph{Left and center}: two views of the YAM dual-arm tabletop setup.
  \emph{Right}: ScaleEvals Bradley--Terry rating versus mean task progress for four anonymized policies (Policy~A--D), with $250$ rollouts per policy under four-policy alternation. Their correlation (Pearson $r{=}0.839$, Spearman $\rho{=}0.800$) supports internal consistency of the reset-free protocol on a second embodiment; YAM does not include an independent manual-evaluation baseline.}}
  \label{fig:yam-crossembodiment}
\end{figure*}

\revised{\paragraph{Cross-embodiment transfer (YAM).}
To test whether \systemname{} generalizes beyond the DROID Franka setup, we run the same reset-free protocol on the YAM bimanual robot, collecting $250$ rollouts per policy across four anonymized policies (Policy~A--D) under continuous $4$-policy alternation. Figure~\ref{fig:yam-crossembodiment} shows that on YAM the ScaleEvals Bradley--Terry rating and mean task progress remain strongly correlated (Pearson $r{=}0.839$, Spearman $\rho{=}0.800$), supporting feasibility of the same protocol on a second embodiment.}

\section{Discussion}

\paragraph{Agreement with manual evaluation.}
Figures~\ref{fig:Comprehensive-convergence} and
\ref{fig:Comprehensive-calibration} show that the Bradley--Terry (BT) ordering
stabilizes by approximately 100 rollouts per policy and correlates strongly with
manual mean task progress (\(r{=}0.94,p{=}0.018\)); raw \systemname{} mean task
progress shows similar agreement (\(r{=}0.92,p{=}0.029\)). The sole rank
disagreement is MolmoAct2 versus GR00T~N1.6. We hypothesize that the fixed manual
suite over-weights skill axes on which N1.6 is stronger, whereas \systemname{}
samples broader skill-tag coverage (Figure~\ref{fig:BothRadars}).

\begin{figure}[!h]
\centering
\begin{minipage}[t]{0.54\linewidth}
\centering
\textbf{From behavioral evidence to a prospective test}\\[0.35em]
\resizebox{\linewidth}{!}{%
\begin{tikzpicture}[
  node distance=3.5mm,
  flow/.style={
    draw=black!45,
    rounded corners=2pt,
    fill=black!2,
    align=center,
    inner sep=3pt,
    font=\scriptsize
  },
  tag/.style={
    flow,
    text width=1.7cm,
    minimum height=1.0cm
  },
  wide/.style={
    flow,
    text width=6.5cm
  },
  arrow/.style={
    ->,
    >=latex,
    line width=0.55pt,
    draw=black!65
  }
]
\node[tag] (skill)
  {\textbf{Skill axis}\\
   \texttt{reposing}\\
   \(\mathrm{N1.7} > \pi_{0.5}\)};
\node[tag, left=3.5mm of skill] (verb)
  {\textbf{Verb axis}\\
   \texttt{rotate}\\
   \(\mathrm{N1.7} > \pi_{0.5}\)};
\node[tag, right=3.5mm of skill] (object)
  {\textbf{Object axis}\\
   \texttt{marker}\\
   \(\mathrm{N1.7} > \pi_{0.5}\)};
\node[wide, below=4.5mm of skill] (aggregate)
  {\textbf{Confidence- and support-weighted Borda aggregation}\\
   \(\displaystyle
   S_i(T)=\sum_{t\in T}c_t\,s_t\,(4-r_{it})\)\\[-0.15em]
   {\tiny \(s_t=\min(n_t/10,1)\)}};
\node[wide, below=3.5mm of aggregate] (test)
  {\textbf{Prospective test frozen before execution}\\
   Prompt: ``Rotate marker \(90^\circ\) clockwise.''\\
   Predicted order:
   \(\mathrm{N1.7} > \pi_{0.5} > \mathrm{N1.6}\)};
\draw[arrow] (verb) -- (aggregate);
\draw[arrow] (skill) -- (aggregate);
\draw[arrow] (object) -- (aggregate);
\draw[arrow] (aggregate) -- (test);
\end{tikzpicture}%
}
\end{minipage}
\hfill
\begin{minipage}[t]{0.43\linewidth}
\centering
\textbf{Prospective physical validation}\\[0.7em]
\small
\setlength{\tabcolsep}{4pt}
\resizebox{\linewidth}{!}{%
\begin{tabular}{lcccc}
\toprule
\textbf{Family}
& \textbf{Predicted order}
& \textbf{\(N\)}
& \textbf{Exact}
& \(\boldsymbol{\bar{\rho}}\) \\
\midrule
Place
& \(\pi_{0.5} > \mathrm{N1.7} > \mathrm{N1.6}\)
& 5 & 5/5 & 0.974 \\
Rotate
& \(\mathrm{N1.7} > \pi_{0.5} > \mathrm{N1.6}\)
& 5 & 4/5 & 0.800 \\
\midrule
\textbf{Overall}
& capability-conditioned
& \textbf{10}
& \textbf{9/10}
& \textbf{0.887} \\
\bottomrule
\end{tabular}
}
\end{minipage}
\caption{\textbf{Tag recomposition enables prospective policy-order prediction.}
\emph{Left}: per-tag BT orderings for the active motion, skill, and object
factors are combined using confidence- and support-weighted Borda aggregation.
The predictor scores candidate tag sets, and a fixed template renders the
selected composition as a natural-language prompt. The prompt and predicted
policy order are frozen before execution. \emph{Right}: prospective validation
recovers 9/10 exact orders over 240 rollouts, with 96\% pairwise concordance.
Appendix Table~\ref{tab:Predict} reports task-level results, and Appendix~B.3
analyzes the Red~5 discrepancy.}
\label{fig:prediction-summary}
\end{figure}

\paragraph{Capability-conditioned prediction.}
The skill-tag profiles in Figure~\ref{fig:BothRadars} localize capability
differences hidden by aggregate scores. We aggregate tag-conditioned BT
orderings over motion verbs, skill tags, and primary objects to score candidate
tag compositions. A prompt template realizes the selected composition as a
natural-language instruction, while its predicted policy order is frozen before
execution. This prediction is nontrivial because \(\pi_{0.5}\) and GR00T N1.7
are adjacent in aggregate BT rank, yet reverse order between the placement and
rotation families. Across 10 prompts and 240 physical rollouts,
Table~\ref{tab:Predict} reports mean Spearman \(\rho{=}0.887\), 90\%
exact-order accuracy, and 96\% pairwise concordance.

These results show that the diagnostics are actionable rather than merely
descriptive. For closely performing policies, aggregate rankings can conceal
conditional strengths, while recombining motion-, skill-, and
object-conditioned evidence predicts where their ordering should persist or
reverse. We call this operational predictive reuse of capability evidence
\emph{behavioral compositionality}. These results suggest stable behavioral regularities without establishing their underlying internal mechanisms. Future work can test whether the
observed regularities correspond to reusable latent policy features using
activation-level probes and controlled causal interventions.

\section{Limitations}
\label{sec:limitations}

\textbf{VLM perception and rubric quality.}
AutoGrader depends on VLM perception and rubric specificity; occlusion,
ambiguous states, and underspecified rubrics can confound progress, with errors
tending to over-credit. Failures concentrate in articulation and rotation tasks,
where depth ordering, contact, or final object state can be ambiguous. AutoGrader
achieves 90.7\% criterion accuracy (86.2\% macro F1, 66.7\% precision, 94.5\%
recall), so periodic auditing remains important. Human correction preserves
ordering (Pearson \(r{=}0.99\), Spearman \(\rho{=}1.00\);
Figure~\ref{fig:grader-score-slope}), while proprietary VLMs limit exact
reproducibility.

\textbf{Scene drift, task-stream noise, and matching.}
Continuous play makes future tasks endogenous and difficulty sensitive to pose,
occlusion, and reachability; string uniqueness also does not ensure semantic
diversity. Matching and 530 interleaved rollouts reduce but do not remove this
dependence. Results remain deployment-specific; YAM lacks a manual baseline and
diversity statistics are descriptive. Moderate drift is useful as a robustness
stress test, but can confound fine-grained comparisons under sparse tag coverage.
Accordingly, \systemname{} complements controlled manual evaluation: able to screen or "vibe-check"
generalist policies to prioritize manual human evaluation, where controlled initial-state tests
remain necessary.

\textbf{Prediction scope.}
Behavioral compositionality is validated prospectively on ten prompts and three
target policies. Establishing its breadth requires more prompt families,
embodiments, and laboratories; composed predictions should prioritize rather
than replace controlled physical validation.


\clearpage
\onecolumn
\acknowledgments{
We thank Mengda Xu, Ryan Punamiya, Jonathan Tremblay, Yu Qi, and Tavish Chen for feedback and valuable discussions, as well as Danyi Chen, Jeremy Chimienti, Mimoza Huynh, Nick Aguilar, Zizheng Wang, Amanpreet Singh, Cristaldo Campos, Gloria Whang, Leilee Naderi, Alec Nagal, Anshuman Sahu, Amy Nguyen, Caleb Geballe, and Matin Nikoui. This work was done at NVIDIA.
}


\bibliography{references}  
\newpage

\appendix
\raggedbottom

\section*{Appendix Overview}
Appendix~A documents AutoPrompter examples, difficulty calibration, accepted
task coverage, scene-aware tagging, and the full task-generation prompt.
Appendix~B reports capability diagnostics, sample-efficiency and throughput
accounting, prospective prediction details, and the held-out manual suite.
Appendix~C contains the AutoGrader audit, balanced-leaderboard calibration, and
extended capability profiles.

\section{AutoPrompter Few-Shot Task Generation}
\label{sec:LLMdetails}

AutoPrompter generates instructions from the current scene while preserving enough structure for grading and comparison. Rather than relying on unconstrained prompting, we ground generation with a curated few-shot library of tasks. Each few-shot example specifies the task language, required scene objects, difficulty annotation, and a stepwise grading rubric expressed as yes/no visual checks. Table~\ref{tab:autoprompter_examples} shows representative examples from this library, and Table~\ref{tab:autoprompter_difficulty} summarizes how the examples calibrate the difficulty labels used downstream for matching.

\subsection{Few-Shot Examples, Difficulty, and Rubrics}
The few-shot library spans the canonical capability tags used in our analysis: transfer, spatial reasoning, precision, reposing, state flip, contact-rich manipulation, and articulation. These examples discourage low-quality prompt distributions dominated by simple pick-and-place, calibrate task difficulty, and demonstrate how generated instructions should be paired with decomposed grading rubrics. Skill tags are assigned after a task is accepted, using the generated instruction and scene context; the tags are used for downstream capability analysis rather than as hard labels during task generation.

\begin{table}[h]
\centering
\small
\begin{adjustbox}{max width=\linewidth}
\begin{tabular}{lll}
\toprule
Example instruction & Skill / role & Difficulty role \\
\midrule
Put the banana on the green plate & transfer & easy single-step placement \\
Center the cup on the plate & spatial & medium placement with alignment \\
Stack the red cube on the yellow cube & precision & medium tight-tolerance placement \\
Turn the apple so its stem points up & reposing & hard orientation control \\
Push the yellow cube into the black square & contact-rich & easy non-prehensile motion \\
Open the drawer fully & articulation & easy articulated-object interaction \\
Open drawer, place marker inside, close it, then stack cubes & composite & expert multi-skill sequence \\
\bottomrule
\end{tabular}
\end{adjustbox}
\caption{Representative AutoPrompter few-shot examples. The library covers the skill tags used for downstream capability analysis and gives the model concrete examples of how task difficulty and rubric decomposition should be expressed.}
\label{tab:autoprompter_examples}
\end{table}

AutoPrompter outputs a 1--10 difficulty score with a corresponding label; Table~\ref{tab:autoprompter_difficulty} summarizes the calibration used for matching and aggregation.

\begin{table}[h]
\centering
\small
\begin{adjustbox}{max width=\linewidth}
\begin{tabular}{lll}
\toprule
Difficulty & Score range & Few-shot calibration \\
\midrule
Easy & 1.0--4.0 & One direct manipulation goal, e.g., transfer or simple push \\
Medium & 4.1--6.0 & Spatial, precision, or sustained-contact requirement \\
Hard & 6.1--8.0 & Three-step composition, articulation, or fragile orientation control \\
Expert & 8.1--10.0 & Four or more ordered subgoals or mixed-skill sequence \\
\bottomrule
\end{tabular}
\end{adjustbox}
\caption{AutoPrompter difficulty calibration. Difficulty labels are grounded by few-shot examples and used for matching and aggregation rather than as a standalone measure of task quality.}
\label{tab:autoprompter_difficulty}
\end{table}

For every generated instruction, AutoPrompter also produces a milestone rubric. Rubric milestones are written as yes/no visual questions and weighted to sum to one. For example, a transfer task may ask whether the object was lifted and whether it reached the target receptacle, while a long-horizon task allocates credit across each subgoal. This makes generated tasks compatible with AutoGrader and avoids reducing evaluation to binary success.

\subsection{Observed Output Distribution}
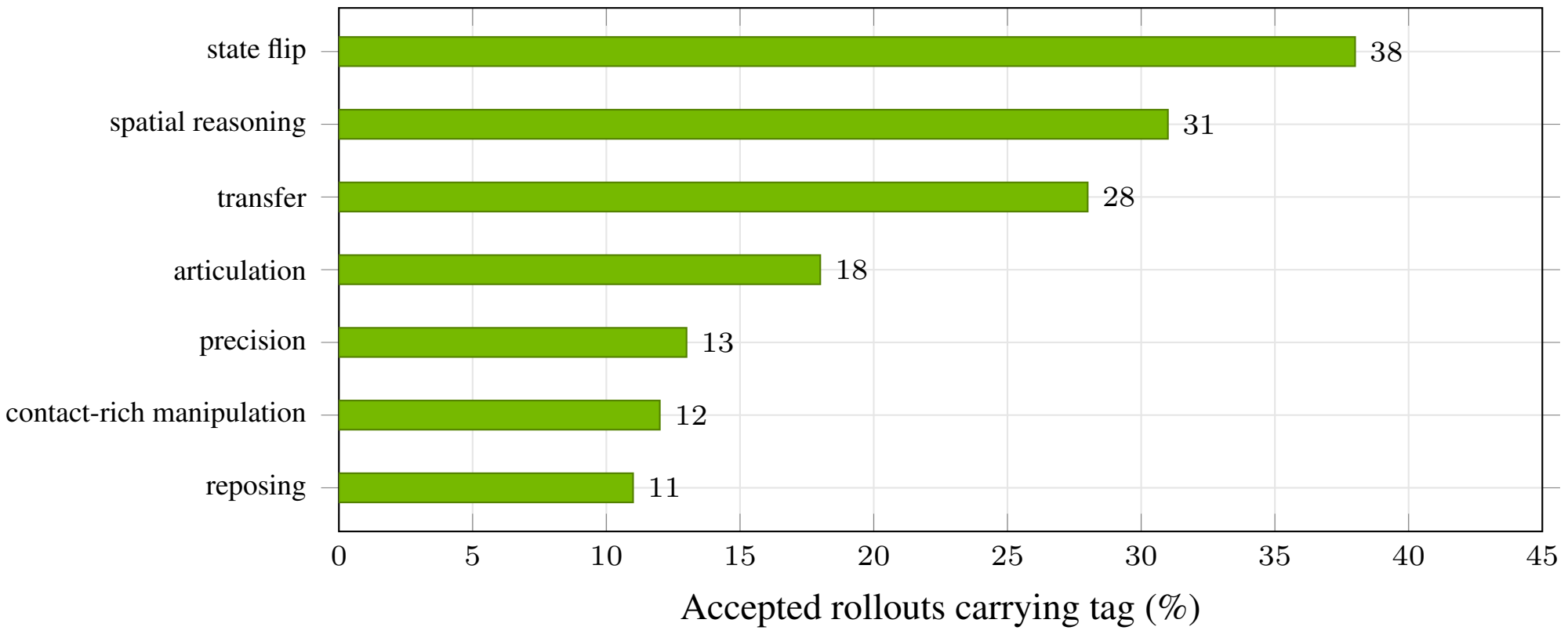
\begin{figure}[h]
\centering
\begin{tikzpicture}
\begin{axis}[
  xbar,
  width=0.84\linewidth,
  height=6.0cm,
  xmin=0,
  xmax=45,
  xlabel={Accepted rollouts carrying tag (\%)},
  symbolic y coords={reposing,contact-rich manipulation,precision,articulation,transfer,spatial reasoning,state flip},
  ytick=data,
  bar width=7pt,
  nodes near coords,
  every node near coord/.append style={font=\scriptsize},
  tick label style={font=\scriptsize},
  label style={font=\small},
  grid=major,
  major grid style={gray!20},
]
\addplot[fill=radarDream, draw=radarDream!70!black] coordinates {
  (11,reposing)
  (12,contact-rich manipulation)
  (13,precision)
  (18,articulation)
  (28,transfer)
  (31,spatial reasoning)
  (38,state flip)
};
\end{axis}
\end{tikzpicture}
\caption{\textbf{Observed AutoPrompter skill-tag distribution.}
Across 2,486 accepted DROID rollouts, generated prompts contain 26 distinct
action verbs and reference 32 objects. Tags are multi-label, so percentages
sum to more than 100\%. The distribution documents task coverage but is not a
standardized semantic-diversity metric.}
\label{fig:autoprompter-output-distribution}
\end{figure}
\FloatBarrier

\begin{figure}[htbp]
  \centering
  \includegraphics[width=0.86\linewidth]{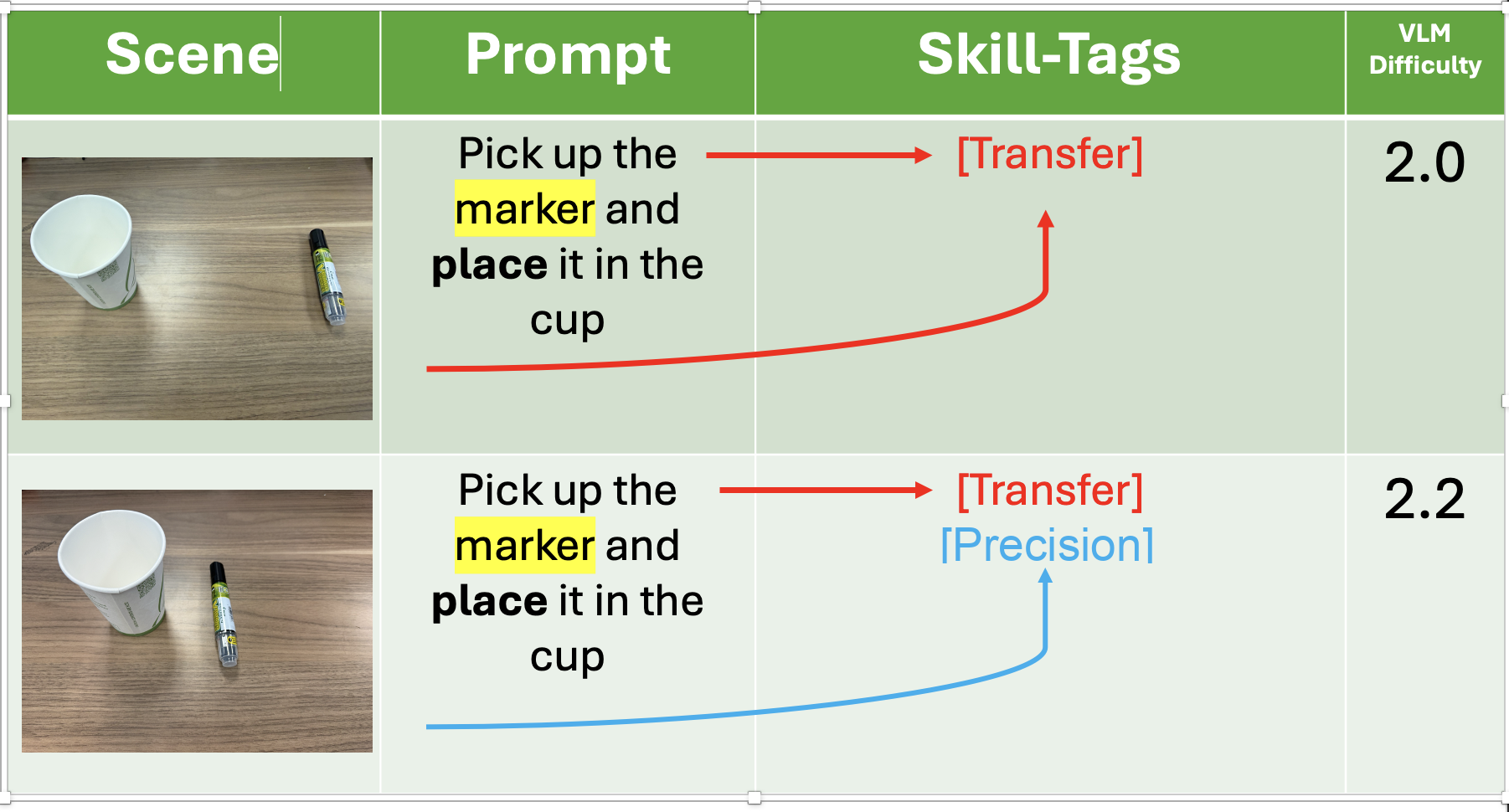}
  \caption{\textbf{Scene-aware skill tagging and difficulty.}
  Both prompts require object transfer. In the second scene, the
  marker-to-cup alignment adds a precision demand, illustrating that skill tags
  depend on both instruction semantics and scene context. VLM-estimated
  difficulty, action verbs, and primary objects provide additional metadata for
  matched Bradley--Terry rankings and prospective prompt composition
  (Table~\ref{tab:Predict}).}
  \label{fig:scene-aware-tagging}
\end{figure}
\FloatBarrier

\subsection{Prompt Template}
The AutoPrompter prompt is instantiated with scene-specific objects, recent task history, and few-shot rubric examples. We reproduce the compact template below.

\begin{tcolorbox}[
    colback=gray!5,
    colframe=gray!50,
    title=AutoPrompter task-generation prompt,
    breakable,
    enhanced
]
\begin{Verbatim}[
    breaklines,
    breakanywhere,
    breaksymbolleft=,
    breakindent=2em,
    fontsize=\scriptsize,
    baselinestretch=0.84,
    fontfamily=tt
]
You are a robot task generator for a single-arm tabletop manipulation robot (Franka FR3).
AVAILABLE OBJECTS IN THE SCENE (with notes):
{object_details_text}
OBJECTS SAFE TO LIFT:
{safe_to_lift_json}
HEAVY OBJECTS (DO NOT ATTEMPT TO LIFT - robot payload limit is 0.5kg):
{heavy_objects_json}
For heavy objects, you can ONLY: push, slide, or interact with items inside/on top of them.
EXAMPLE TASKS (for reference on style and complexity):
{example_task_text}
PREVIOUSLY ATTEMPTED TASKS (DO NOT REPEAT THESE - choose something DIFFERENT):
{previously_attempted_tasks}
IMPORTANT: The tasks above have already been attempted in this session. You MUST generate a DIFFERENT task.
Even if the scene looks similar, pick a different object or a different action. Be creative and vary your choices.
ACTION VARIETY - Use diverse verbs and motions. Consider these actions:
  - MANIPULATION: pick, place, grasp, release, lift, lower, hold
  - MOVEMENT: push, slide, drag, move, shift, reposition, transfer
  - ROTATION: rotate, turn, flip, spin, orient, tilt, tip
  - ARRANGEMENT: stack, align, center, match, group, separate, spread
  - CONTAINER: put into, take out of, deposit, extract, scoop
  - ORGANIZATION: clear, gather, arrange, sort, line up
FORBIDDEN ACTIONS (the robot physically CANNOT do these - NEVER generate tasks involving them):
  - cutting, slicing, chopping (no blade/knife capability)
  - pouring, filling, emptying liquids (no liquid handling)
  - throwing, tossing, catching (no ballistic control)
  - folding, unfolding (no deformable-object manipulation)
  - tying, untying, knotting (no rope/string manipulation)
  - pressing buttons (too precise for the gripper)
  - using two hands (single-arm robot)
Note: object NAMES like "watermelon slice" are fine. The word "slice" in a name is not a slicing action.
RULES FOR TASK GENERATION:
1. ONLY use objects that you can ACTUALLY SEE in the camera image.
2. Use the EXACT object names from the AVAILABLE OBJECTS list. Do not add adjectives, modifiers, or synonyms.
3. NEVER attempt to lift heavy objects marked as unsafe. Only push, slide, or interact with items inside/on top of them.
4. Tasks must be achievable by a single-arm robot on a tabletop.
5. Tasks should be one clear sentence in natural language.
6. Task timeout MUST match the difficulty SCORE:
   score <= 4  -> 30s
   score <= 6  -> 60s
   score <= 8  -> 90s
   score <= 10 -> 120s
7. Avoid multi-step tasks that require more than 2-3 actions unless the requested difficulty is hard or expert.
8. Avoid tasks that would require two hands or complex tool use.
9. Be creative with actions. Do not generate only pick-and-place tasks.
10. Do not repeat any previously attempted task.
EXAMPLE TASK RUBRICS:
{few_shot_rubric_examples}
DIFFICULTY CLASSIFICATION (you MUST classify the task on a 1-10 scale, NOT 0-1):
- easy (score 1.0-4.0): Single-step pick-and-place, simple slides/pushes, basic object identification.
- medium (score 4.1-6.0): Spatial reasoning, alignment ("center", "between"), precision insertion/stacking, sustained contact-rich manipulation.
- hard (score 6.1-8.0): Multi-step tasks with 3 or more actions, mixed-skill chains, articulation, dependent sequencing.
- expert (score 8.1-10.0): Long ordered sequences, complex multi-step tasks with precision subtasks, tool-like interaction.
The score MUST be between 1 and 10. Do NOT use a 0-1 scale. The label MUST match the score band.
GRADING RUBRIC GENERATION:
You must also generate a grading rubric with 2-4 milestones that an autograder can use to score the task.
Each milestone should award partial credit for progress toward the goal.
Every sub-task or milestone description must be phrased as a yes/no question, for example:
"Did the robot pick up the banana?"
"Did the robot place the fruit on the plate?"
"Did the drawer end up mostly open?"
Milestones should be ordered from easiest to hardest, with weights summing to 1.0.
RESPOND IN JSON FORMAT:
{
    "language": "<task description in natural language>",
    "timeout": <MUST match difficulty SCORE: score<=4 -> 30, score<=6 -> 60, score<=8 -> 90, score<=10 -> 120>,
    "objects_used": ["<exact object names from the AVAILABLE OBJECTS list that this task involves>"],
    "reasoning": "<brief explanation of why this task is appropriate given the current scene>",
    "difficulty": {
        "label": "<easy|medium|hard|expert>",
        "score": <1.0-10.0 within the difficulty range>,
        "reasoning": "<why this difficulty level is appropriate>"
    },
    "grading_rubric": {
        "type": "milestone_stepwise",
        "max_score": 1.0,
        "milestones": [
            {
                "weight": <0.0-1.0>,
                "description": "<yes/no question a grader or autograder answers to award this credit>"
            },
            {
                "weight": <0.0-1.0>,
                "description": "<yes/no question a grader or autograder answers to award this credit>"
            }
        ]
    }
}
\end{Verbatim}

\end{tcolorbox}

\FloatBarrier
\Needspace{10\baselineskip}
\section{Capability Diagnostics and Robustness}
\subsection{Skill-Conditioned Capability Analysis}

\begin{figure}[H]
\centering
\providecolor{radarDream}{HTML}{76B900}   
\providecolor{radarPi}{HTML}{F5A623}      
\providecolor{radarN17}{HTML}{3B82F6}     
\providecolor{radarMolmo}{HTML}{E74C3C}   
\providecolor{radarN16}{HTML}{8B5CF6}     
%
\begin{minipage}{\linewidth}
\centering
\begin{tikzpicture}
\begin{polaraxis}[
    width=7.5cm,
    height=7.5cm,
    xmin=0, xmax=360,
    ymin=0, ymax=60,
    xtick={90.0000,38.5714,347.1429,295.7143,244.2857,192.8571,141.4286},
    xticklabels={%
        transfer$^{***}$,%
        contact rich$^{***}$,%
        state flip$^{***}$,%
        articulation$^{***}$,%
        spatial$^{***}$,%
        precision$^{***}$,%
        reposing$^{*}$%
    },
    xticklabel style={font=\scriptsize, align=center, anchor=center},
    ytick={10,20,30,40,50,60},
    yticklabel=\empty,
    ytick distance=10,
    grid=both,
    major grid style={dashed, line width=0.3pt, draw=gray!40},
    minor grid style={line width=0.2pt, draw=gray!15},
    axis line style={draw=none},
    tick style={draw=none},
    axis on top,
    legend style={
        font=\scriptsize,
        at={(0.5,-0.06)}, anchor=north,
        legend columns=5,
        draw=gray!50,
        inner xsep=2pt, inner ysep=1.5pt,
        nodes={inner sep=1pt},
        /tikz/every even column/.append style={column sep=0.25em},
    },
    legend cell align=left,
    legend image post style={mark=*, mark size=1.6pt},
    clip=false,
]
\addplot+[color=radarDream, fill=radarDream, fill opacity=0.08, draw=none, mark=none, forget plot]
    coordinates {(90.0000,46.92) (38.5714,44.87) (347.1429,31.08) (295.7143,38.72) (244.2857,36.71) (192.8571,29.55) (141.4286,35.38) (90.0000,46.92)};
\addplot+[color=radarPi, fill=radarPi, fill opacity=0.08, draw=none, mark=none, forget plot]
    coordinates {(90.0000,37.38) (38.5714,36.74) (347.1429,27.37) (295.7143,30.23) (244.2857,30.35) (192.8571,29.23) (141.4286,28.90) (90.0000,37.38)};
\addplot+[color=radarN17, fill=radarN17, fill opacity=0.08, draw=none, mark=none, forget plot]
    coordinates {(90.0000,23.45) (38.5714,35.71) (347.1429,19.14) (295.7143,22.77) (244.2857,22.63) (192.8571,21.23) (141.4286,33.39) (90.0000,23.45)};
\addplot+[color=radarMolmo, fill=radarMolmo, fill opacity=0.08, draw=none, mark=none, forget plot]
    coordinates {(90.0000,17.39) (38.5714,17.24) (347.1429,14.68) (295.7143,18.23) (244.2857,20.90) (192.8571,13.67) (141.4286,21.83) (90.0000,17.39)};
\addplot+[color=radarN16, fill=radarN16, fill opacity=0.08, draw=none, mark=none, forget plot]
    coordinates {(90.0000,15.45) (38.5714,14.04) (347.1429,14.31) (295.7143,17.51) (244.2857,14.17) (192.8571,17.73) (141.4286,17.16) (90.0000,15.45)};
\addplot+[color=radarDream, line width=1.2pt, mark=*, mark size=3pt,
    mark options={draw=black, line width=0.5pt, fill=radarDream}]
    coordinates {(90.0000,46.92) (38.5714,44.87) (347.1429,31.08) (295.7143,38.72) (244.2857,36.71) (192.8571,29.55) (141.4286,35.38) (90.0000,46.92)};
\addlegendentry{DreamZero}
\addplot+[color=radarPi, line width=1.2pt, mark=diamond*, mark size=3pt,
    mark options={draw=black, line width=0.5pt, fill=radarPi}]
    coordinates {(90.0000,37.38) (38.5714,36.74) (347.1429,27.37) (295.7143,30.23) (244.2857,30.35) (192.8571,29.23) (141.4286,28.90) (90.0000,37.38)};
\addlegendentry{$\pi_{0.5}$}
\addplot+[color=radarN17, line width=1.2pt, mark=triangle*, mark size=3pt,
    mark options={draw=black, line width=0.5pt, fill=radarN17}]
    coordinates {(90.0000,23.45) (38.5714,35.71) (347.1429,19.14) (295.7143,22.77) (244.2857,22.63) (192.8571,21.23) (141.4286,33.39) (90.0000,23.45)};
\addlegendentry{GR00T N1.7}
\addplot+[color=radarMolmo, line width=1.2pt, mark=pentagon*, mark size=3pt,
    mark options={draw=black, line width=0.5pt, fill=radarMolmo}]
    coordinates {(90.0000,17.39) (38.5714,17.24) (347.1429,14.68) (295.7143,18.23) (244.2857,20.90) (192.8571,13.67) (141.4286,21.83) (90.0000,17.39)};
\addlegendentry{MolmoAct2}
\addplot+[color=radarN16, line width=1.2pt, mark=square*, mark size=3pt,
    mark options={draw=black, line width=0.5pt, fill=radarN16}]
    coordinates {(90.0000,15.45) (38.5714,14.04) (347.1429,14.31) (295.7143,17.51) (244.2857,14.17) (192.8571,17.73) (141.4286,17.16) (90.0000,15.45)};
\addlegendentry{GR00T N1.6}
\node[anchor=south west, font=\scriptsize, gray] at (axis cs:115,10) {10\%};
\node[anchor=south west, font=\scriptsize, gray] at (axis cs:115,30) {30\%};
\node[anchor=south west, font=\scriptsize, gray] at (axis cs:115,50) {50\%};
\end{polaraxis}
\end{tikzpicture}
\\[0.2em]
\subcaption{Skill radar per policy. Spoke labels carry FDR-significance stars only; effect sizes $\eta^{2}$ are in the table below.}
\label{fig:skill-radar}
\end{minipage}
\vspace{0.8em}
\begin{minipage}{\linewidth}
\centering
\small
\setlength{\tabcolsep}{6pt}
\renewcommand{\arraystretch}{1.15}
\begin{tabular}{@{}l r r r r c c c c c@{}}
\toprule
\textbf{Skill tag} & $\boldsymbol{n}$ & $\boldsymbol{F}$ & $\boldsymbol{p}$ & $\boldsymbol{\eta^{2}}$ & \textbf{DZ} & $\boldsymbol{\pi_{0.5}}$ & \textbf{N1.7} & \textbf{MA2} & \textbf{N1.6} \\
\midrule
transfer$^{\dagger}$      & 707 & 22.0           & $<\!0.001$ & 0.11 & \textbf{47} & 37 & 23 & 17 & 15 \\
contact\_rich$^{\dagger}$ & 298 & \phantom{0}9.0 & $<\!0.001$ & 0.11 & \textbf{45} & 37 & 36 & 17 & 14 \\
state\_flip$^{\dagger}$   & 952 & 17.2           & $<\!0.001$ & 0.07 & \textbf{31} & 27 & 19 & 15 & 14 \\
long\_horizon$^{\dagger}$ & 996 & 19.5           & $<\!0.001$ & 0.07 & \textbf{32} & 25 & 19 & 14 & 15 \\
articulation$^{\dagger}$  & 448 & \phantom{0}7.9 & $<\!0.001$ & 0.07 & \textbf{39} & 30 & 23 & 18 & 18 \\
spatial$^{\dagger}$       & 761 & 13.3           & $<\!0.001$ & 0.07 & \textbf{37} & 30 & 23 & 21 & 14 \\
precision$^{\dagger}$     & 333 & \phantom{0}5.4 & $<\!0.001$ & 0.06 & \textbf{30} & 29 & 21 & 14 & 18 \\
reposing$^{\dagger}$      & 272 & \phantom{0}3.3 & 0.011      & 0.05 & \textbf{35} & 29 & 33 & 22 & 17 \\
\bottomrule
\end{tabular}
\\[0.4em]
\subcaption{Tag-stratified one-way ANOVA, skill dimension + long\_horizon (matched-only, 5 policies). $^{\dagger}$\,Significant under Benjamini--Hochberg FDR at $\alpha{=}0.05$. Bold marks the per-row leader.}
\label{tab:tag-anova}
\end{minipage}
\caption{\textbf{Skill profile and per-tag ANOVA.} \textbf{(a)}~Per-skill mean task progress on matched-pair tasks across all five policies. Spokes are sorted by one-way-ANOVA effect size ($\eta^2$, policy across TP) descending, starting from the top and going clockwise. Radii are raw mean TP in percent (radial gridlines at 10, 20, 30, 40, 50, 60\%). Significance stars use Benjamini--Hochberg FDR-corrected $p$-values ($^{*}p{<}0.05$, $^{**}p{<}0.01$, $^{***}p{<}0.001$). Color palette matches the PlayArena dashboard (positional by BT rank). \textbf{(b)}~Same data in tabular form, augmented with the binary \texttt{longHorizon} dimension; cells are per-policy mean TP in \%. DreamZero now leads on all eight tags (narrowest margins: precision 30 vs $\pi_{0.5}$ 29, reposing 35 vs GR00T N1.7 33); the previously reported precision/reposing inversions disappeared as more matched comparisons accrued. \emph{Caveat (radar).} Radar charts encode value as radius but readers perceive area ($\propto r^2$), which inflates apparent differences; the table is the quantitative source of record.}
\label{fig:Anova}
\end{figure}
\FloatBarrier
\FloatBarrier

\Needspace{10\baselineskip}
\subsection{Sample Efficiency and Operational Throughput}
\begin{table}[!ht]
  \centering
  \small
  \renewcommand{\arraystretch}{1.15}
  \setlength{\tabcolsep}{5pt}
  \begin{tabular}{@{}lrrrr@{}}
    \toprule
    \textbf{Policy} & \textbf{Full \(n\)} &
    \textbf{BT at \(B{=}100\)}\,$\uparrow$ &
    \textbf{TP\,\%} & \textbf{SR\,\%} \\
    \midrule
    DreamZero             & 385 & 1594\,$\pm$\,15 & 39.1 & 14.0 \\
    $\pi_{0.5}$ (anchor)  & 681 & 1524\,$\pm$\,30 & 30.0 & \phantom{0}8.0 \\
    GR00T N1.7            & 601 & 1500\,$\pm$\,\phantom{0}5 & 23.3 & \phantom{0}4.0 \\
    MolmoAct2             & 440 & 1470\,$\pm$\,28 & 17.5 & \phantom{0}2.5 \\
    GR00T N1.6            & 379 & 1453\,$\pm$\,25 & 14.7 & \phantom{0}0.0 \\
    \bottomrule
  \end{tabular}
  \caption{\textbf{Minimum-sample stress test: balanced subset at $B{=}100$ trajectories per policy.}
  The same two-pass balancing protocol as the equal-\(N\), \(B{=}350\) analysis
  trims each policy to \(B{=}100\), the rank lock-in identified in
  Figure~\ref{fig:Comprehensive-convergence}. Values are \(K{=}40\) bootstrap
  medians and the BT \(\pm\) column reports median Hessian SE on the Elo scale.
  The five-way ordering is unchanged; relative to \(B{=}350\), BT shifts are at
  most 15 Elo and TP/SR shifts are below one percentage point.}
  \label{tab:scaleevals100}
\end{table}

\begin{table}[!ht]
  \centering
  \small
  \renewcommand{\arraystretch}{1.15}
  \setlength{\tabcolsep}{7pt}
  \begin{tabular}{@{}lccc@{}}
    \toprule
    \textbf{Evaluation method}
      & \textbf{Net rollouts/h/station}
      & \textbf{Active human time/100}
      & \textbf{Operator scaling} \\
    \midrule
    Manual reset-required baseline & 14.8 & 4 h & 1:1 \\
    \textbf{\systemname{}} & \textbf{23.0} & \textbf{5 min} & \textbf{1:$N$} \\
    \bottomrule
  \end{tabular}
  \caption{\textbf{End-to-end operational throughput and human involvement.}
  Net throughput includes AutoPrompter VLM calls, failed task-generation retries,
  safety and visibility checks, policy execution, and rollout time. The manual baseline
  uses one operator per station for scene resets and grading. \systemname{} reduces
  active human involvement from 4 hours to 5 minutes per 100 rollouts (98\%) and recorded
  five human interventions over the additional 530-rollout temporal-drift evaluation.}
  \label{tab:Time}
\end{table}

\begin{figure}[htbp]
  \centering
  \includegraphics[width=0.72\linewidth]{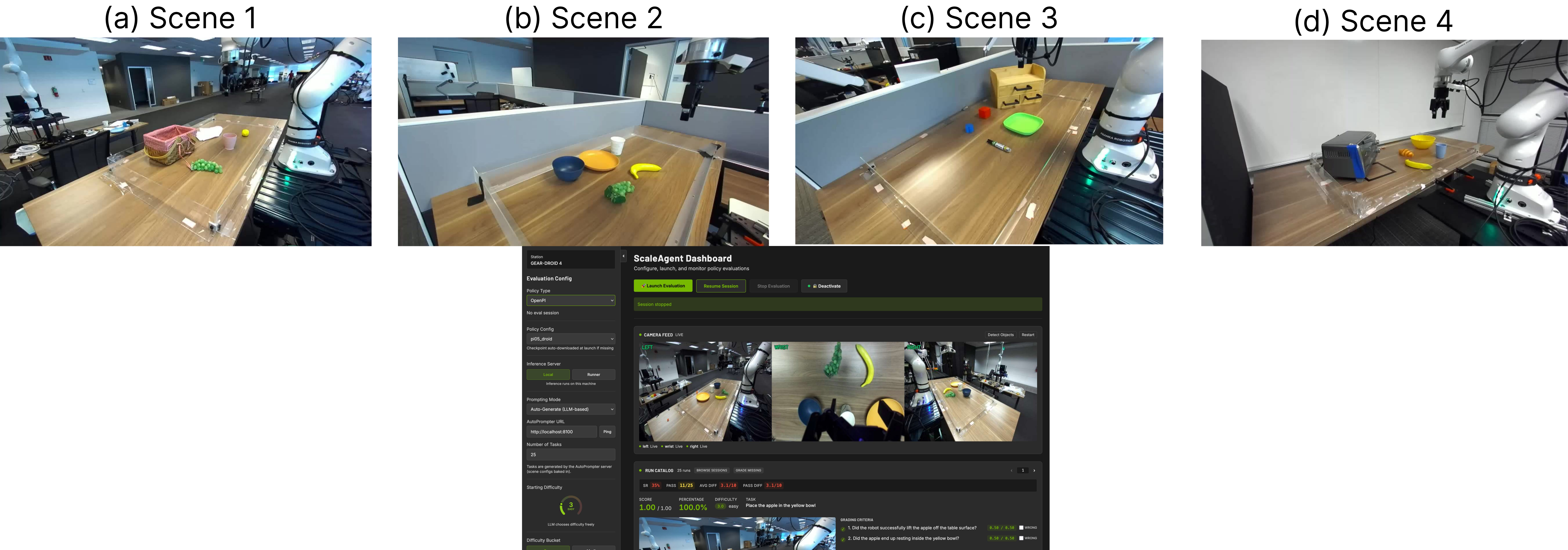}
  \caption{\textbf{ScaleEvals operator dashboard.} The WebUI configures and
  launches evaluations, displays live camera observations, tracks generated
  tasks and rollout history, and exposes AutoGrader rubric-level scores for
  auditing and correction. The displayed camera panes show the original monitored rollout views.}
  \label{fig:operator-dashboard}
\end{figure}

\FloatBarrier

\Needspace{8\baselineskip}
\subsection{Prospective Prediction Details}
For each action-verb, skill-tag, primary-object, and long-horizon value where
available, we fit a tag-conditioned BT ranking from matched rollouts and retain
values for which all three target policies have sufficient evidence. Each
tag-conditioned ordering casts Borda points \(3,2,1\), producing
\[
  S_i(q)=\sum_{t\in T(q)}
  c_t\,\min(n_t/10,1)\,(4-r_{it}),
\]
where \(c_t\in\{1.0,0.7,0.3,0\}\) is the confidence weight for robust, stable,
exploratory, and insufficient evidence; \(n_t\) is the minimum target-policy
pair count; and \(r_{it}\) is policy \(i\)'s rank for tag value \(t\).
The resulting scores order candidate tag compositions. A prompt template then
realizes the selected composition as a natural-language instruction, and both
the prompt and predicted order are frozen before execution. Pairwise tag tests
are Benjamini--Hochberg corrected globally and reported as reliability
diagnostics rather than used as a hard inclusion gate.

The single miss, Red~5, contains conflicting evidence: its rotate and reposing
components favor GR00T N1.7, while its object-conditioned ordering favors
\(\pi_{0.5}\). The object-conditioned signal is based on only 10--15
comparisons and does not pass BH-FDR, so we report it as conflicting
exploratory evidence rather than a statistically established cause of the
miss. All ten prompts and orderings were fixed before physical execution; the
subsequent 240 rollouts were used only for prospective validation.

\begin{table}[!ht]
  \centering
  \small
  \setlength{\tabcolsep}{6pt}
  \renewcommand{\arraystretch}{1.2}
  \begin{adjustbox}{max width=\textwidth}
  \begin{tabular}{@{}p{5.2cm} r c c@{}}
    \toprule
    \textbf{Prompt} & \textbf{Tags} & $\boldsymbol{\rho}$ & \textbf{Exact Order} \\
    \midrule
    \multicolumn{4}{@{}l}{\textit{Place suite} --- predicted $\pi_{0.5} >$ N1.7 $>$ N1.6} \\
    \midrule
    Place banana on orange plate         & {\scriptsize\texttt{(place; transfer; banana)}}         & 1.00 & \checkmark \\
    Slide blue bowl to orange plate      & {\scriptsize\texttt{(slide; contact\_rich; blue bowl)}} & 1.00 & \checkmark \\
    Place grapes on blue bowl            & {\scriptsize\texttt{(place; transfer; grapes)}}         & 1.00 & \checkmark \\
    Place paper cup on orange plate      & {\scriptsize\texttt{(place; transfer; paper\_cup)}}     & 1.00 & \checkmark \\
    Place blue bowl on orange plate      & {\scriptsize\texttt{(place; transfer; blue bowl)}}      & 0.87 & \checkmark \\
    \midrule
    \multicolumn{4}{@{}l}{\textit{Rotate suite} --- predicted N1.7 $> \pi_{0.5} >$ N1.6} \\
    \midrule
    Rotate marker 90$^{\circ}$ CW        & {\scriptsize\texttt{(rotate; reposing; marker)}}        & 1.00 & \checkmark \\
    Rotate pink mug 90$^{\circ}$ CW      & {\scriptsize\texttt{(rotate; reposing; pink mug)}}      & 1.00 & \checkmark \\
    Rotate green plate 90$^{\circ}$ CW   & {\scriptsize\texttt{(rotate; reposing; green plate)}}   & 1.00 & \checkmark \\
    Rotate cup 90$^{\circ}$ CW           & {\scriptsize\texttt{(rotate; reposing; cup)}}           & 1.00 & \checkmark \\
    Rotate red 5 90$^{\circ}$ CW         & {\scriptsize\texttt{(rotate; reposing; Red 5)}}         & 0.00 & $\times$ \\
    \midrule
    \multicolumn{4}{@{}l@{}}{\textbf{Summary}: $N{=}240$ trajs (10 prompts $\times$ 8 rollouts $\times$ 3 policies); $\bar{\rho}{=}0.887$; exact-order hit $=90\%$; concordance $=96\%$.} \\
    \bottomrule
  \end{tabular}
  \end{adjustbox}
  \caption{\textbf{Predictive validation of tag-composed prompts.}
  Each prompt is synthesized from a Borda vote (action verb, skill tag(s), primary
  object), shown in the \textbf{Tags} column in \texttt{teletype}. The predicted
  ordering (stated in each section header) is produced \emph{before} the rig is run;
  we then execute $N{=}8$ episodes per cell and grade with the AutoGrader.
  Spearman~$\rho$ is the per-prompt rank correlation between predicted ranks and
  observed mean task-progress scores; \textbf{Exact Order} indicates whether
  the complete predicted three-policy order is observed. The sole miss was the Red~5 prompt. The rotate and reposing factors favored GR00T N1.7, whereas the object-conditioned BT ordering for Red~5 favored
\(\pi_{0.5}\). The confidence- and support-weighted Borda aggregation therefore
predicted \(\mathrm{N1.7}>\pi_{0.5}>\mathrm{N1.6}\), but physical validation
placed \(\pi_{0.5}\) first and tied GR00T N1.7 with GR00T N1.6, yielding
Spearman \(\rho{=}0.00\). The Red~5 object ordering was estimated from only
10--15 comparisons and did not pass BH-FDR, so we treat it as exploratory
conflicting evidence rather than a statistically established object effect.}
  \label{tab:Predict}
\end{table}
\FloatBarrier

\Needspace{10\baselineskip}
\subsection{Tag-Conditioned Statistics and Manual Evaluation}
The held-out suite contains 20 diverse reset-required tasks with two initial
frames per policy/task cell where available. Table~\ref{tab:F6-detailed}
reports the task-level progress values and corresponding initial observations
used by the main calibration and radar analyses.
\begin{table}[h]
\centering
\small
\setlength{\tabcolsep}{4pt}
\renewcommand{\arraystretch}{1.15}
\caption{Manual task-progress results for the 20 diverse manipulation tasks underlying the Figure~6 calibration and Figure~7 radar diagnostics. Each entry is the mean task progress over two trials with different initial frames. Best per row in \textbf{bold}.}
\label{tab:F6-detailed}
\begin{adjustbox}{max width=\textwidth}
\begin{tabular}{p{0.30\linewidth} c ccccc}
\toprule
\textbf{Task (Language Prompt)} & \textbf{Initial frame} & \textbf{GR00T N1.6} & \textbf{GR00T N1.7} & \textbf{$\pi_{0.5}$} & \textbf{DreamZero} & \textbf{MolmoAct~2} \\
\midrule
Put the banana on the blue plate & \includegraphics[width=0.10\linewidth,height=0.035\textheight,keepaspectratio]{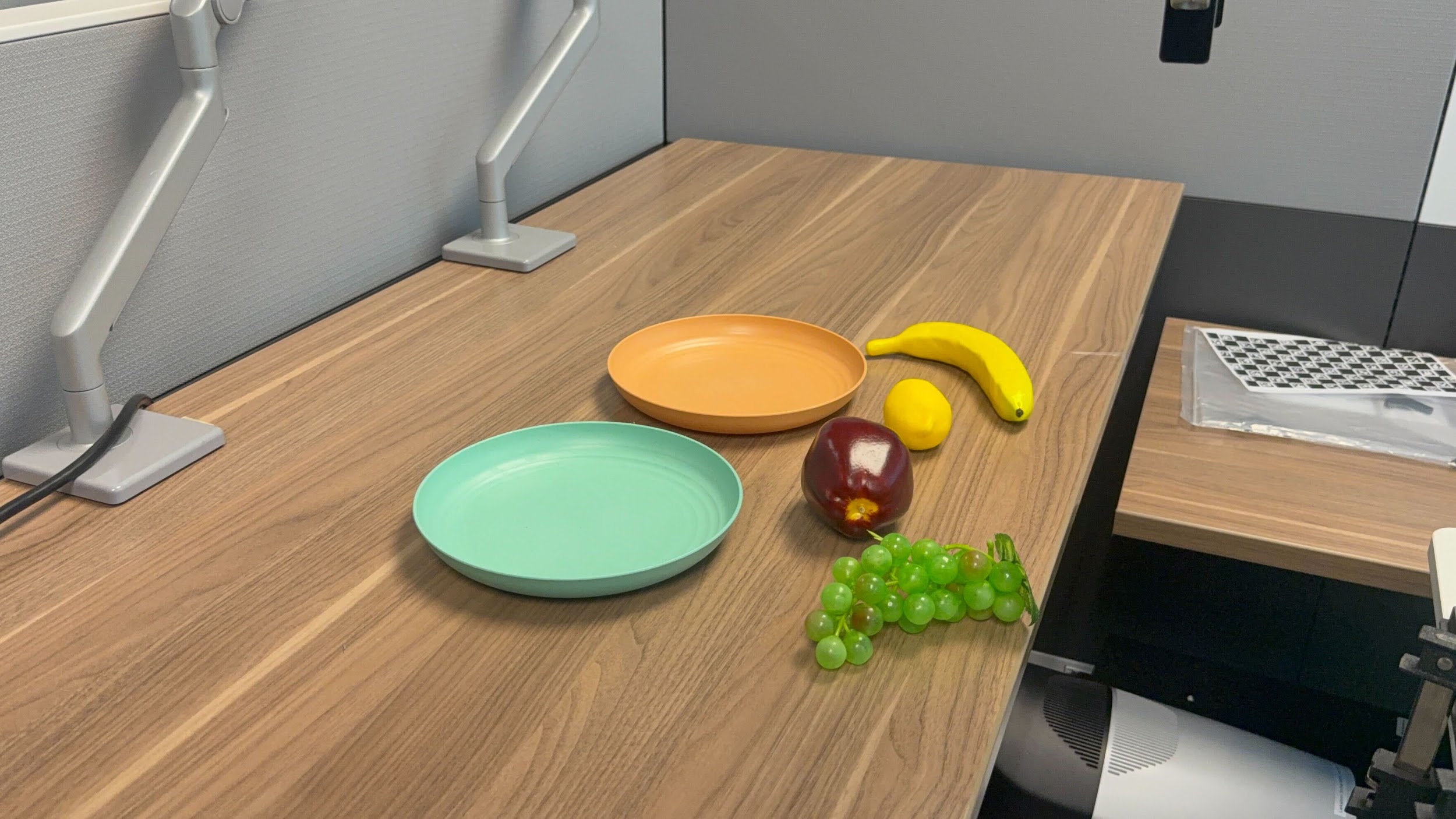} & \textbf{1.00} & \textbf{1.00} & \textbf{1.00} & \textbf{1.00} & \textbf{1.00} \\
Put the `red' cube on `top' of the yellow cube & \includegraphics[width=0.10\linewidth,height=0.035\textheight,keepaspectratio]{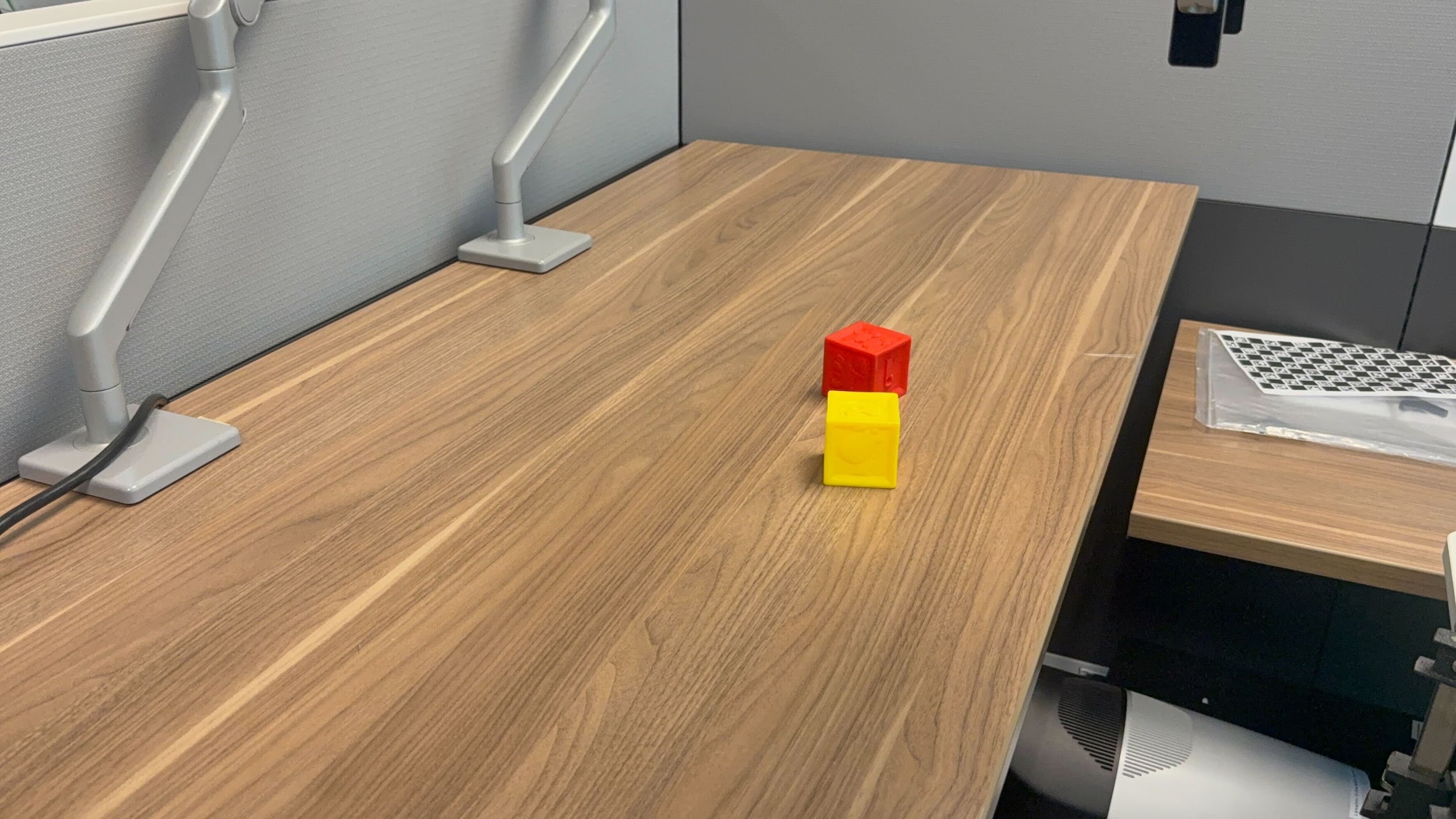} & 0.75 & 0.50 & 0.25 & \textbf{1.00} & 0.25 \\
Open the oven door, pickup the croissant, and place it in the oven & \includegraphics[width=0.10\linewidth,height=0.035\textheight,keepaspectratio]{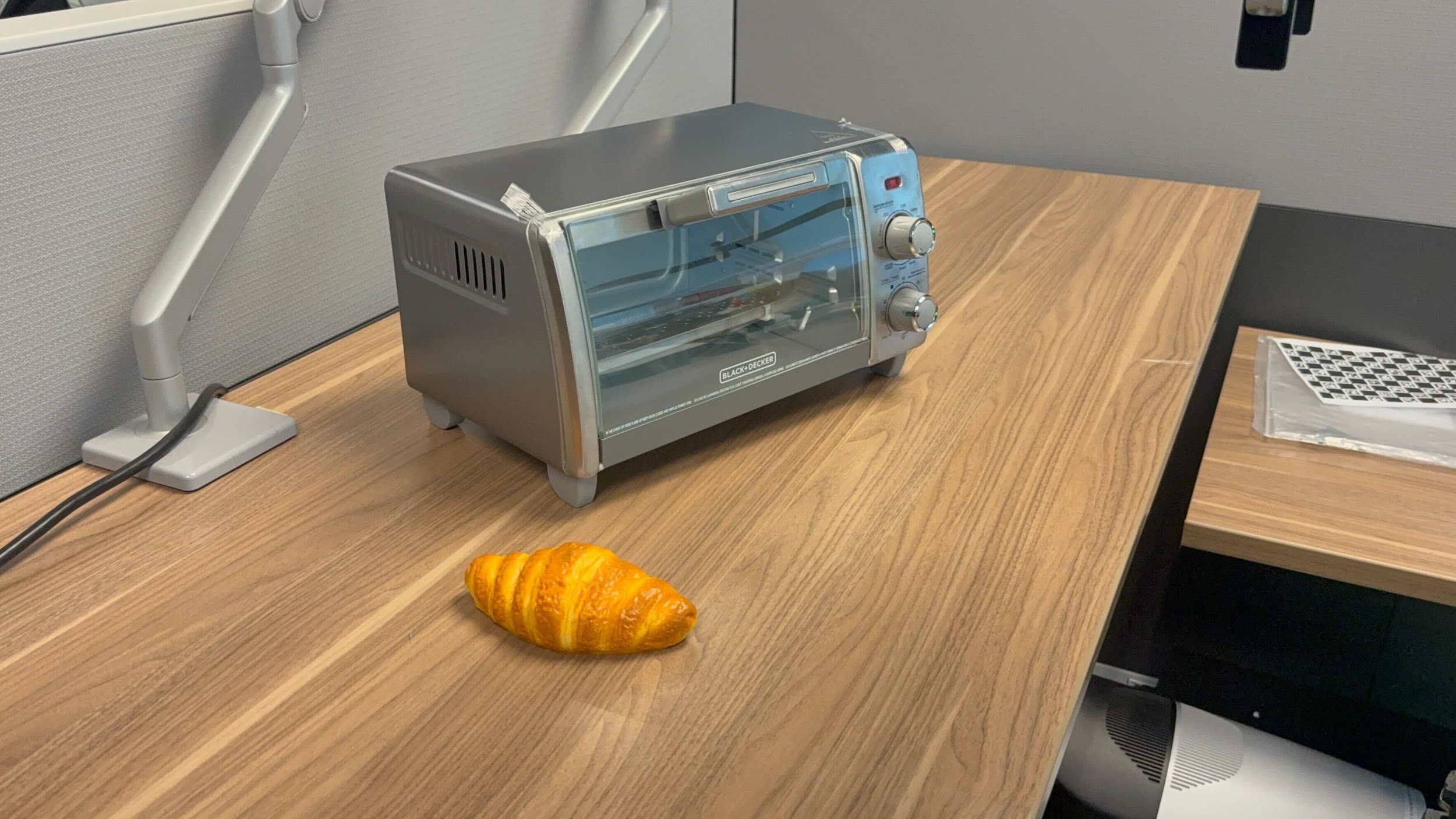} & 0.25 & 0.00 & 0.33 & \textbf{0.49} & 0.00 \\
Pick up bread and place into toaster & \includegraphics[width=0.10\linewidth,height=0.035\textheight,keepaspectratio]{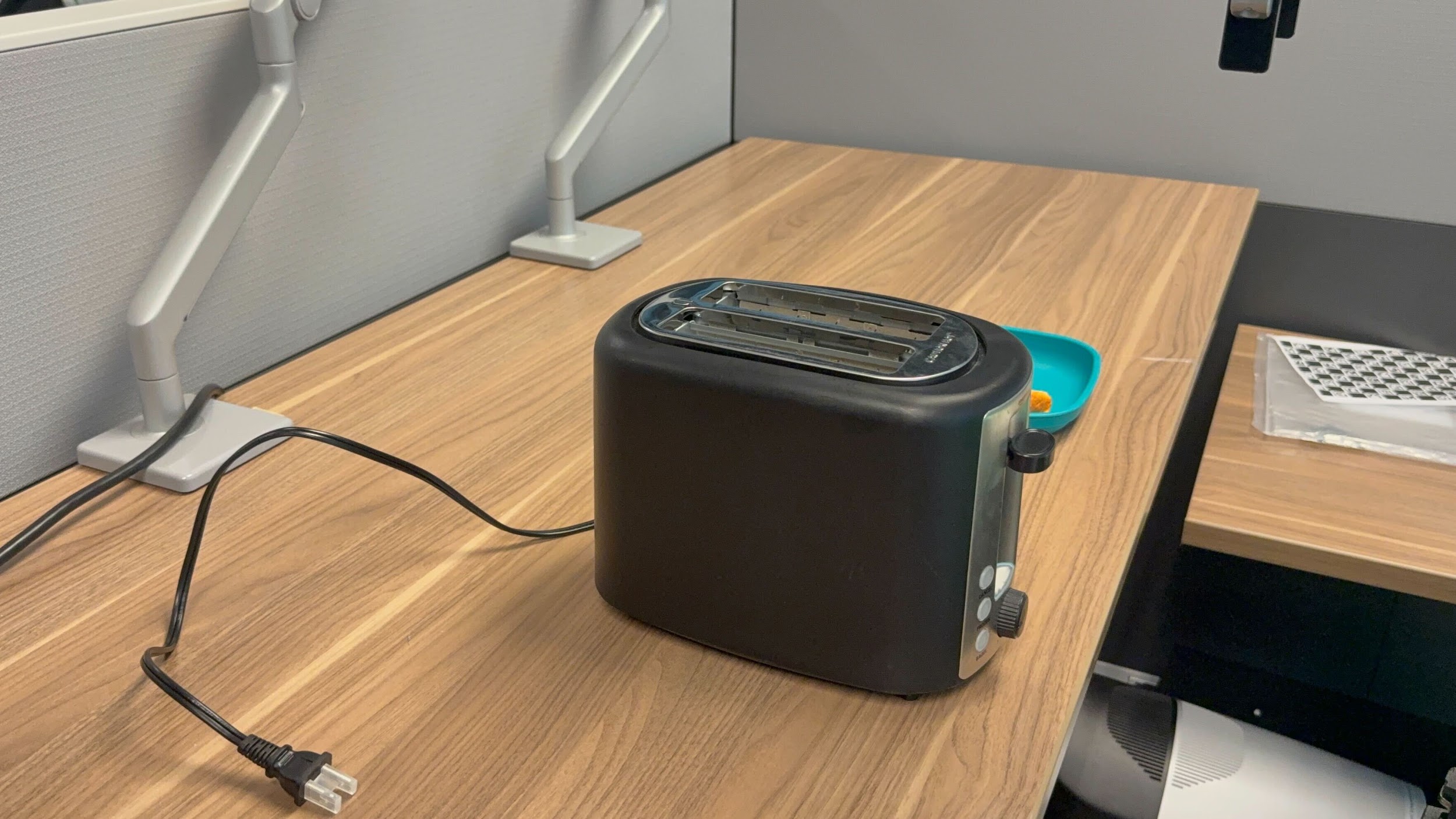} & \textbf{0.75} & 0.25 & 0.25 & 0.62 & 0.00 \\
Put the marker in the drawer and close the drawer & \includegraphics[width=0.10\linewidth,height=0.035\textheight,keepaspectratio]{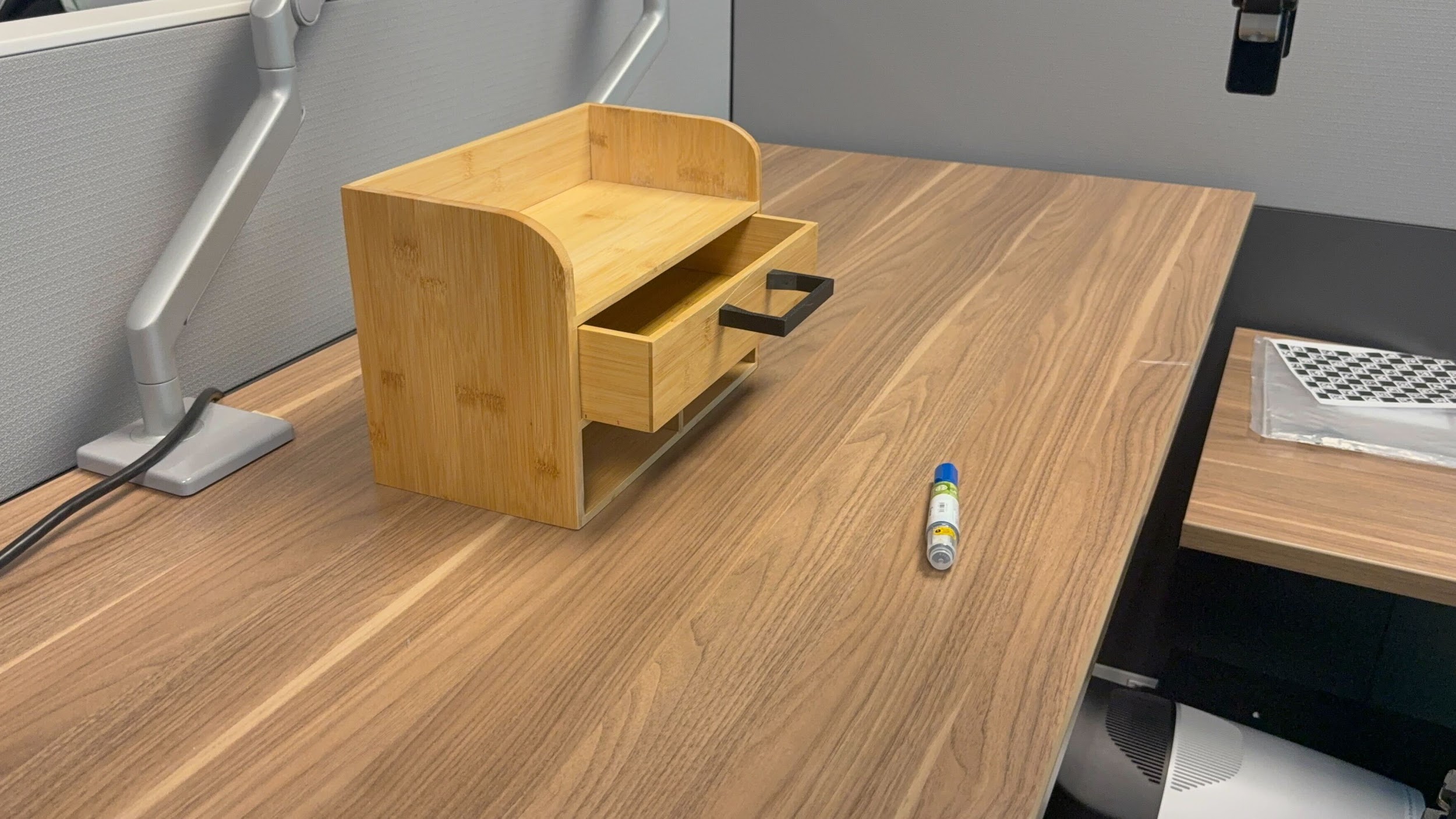} & 0.50 & \textbf{0.75} & 0.50 & -- & 0.12 \\
Flip cup & \includegraphics[width=0.10\linewidth,height=0.035\textheight,keepaspectratio]{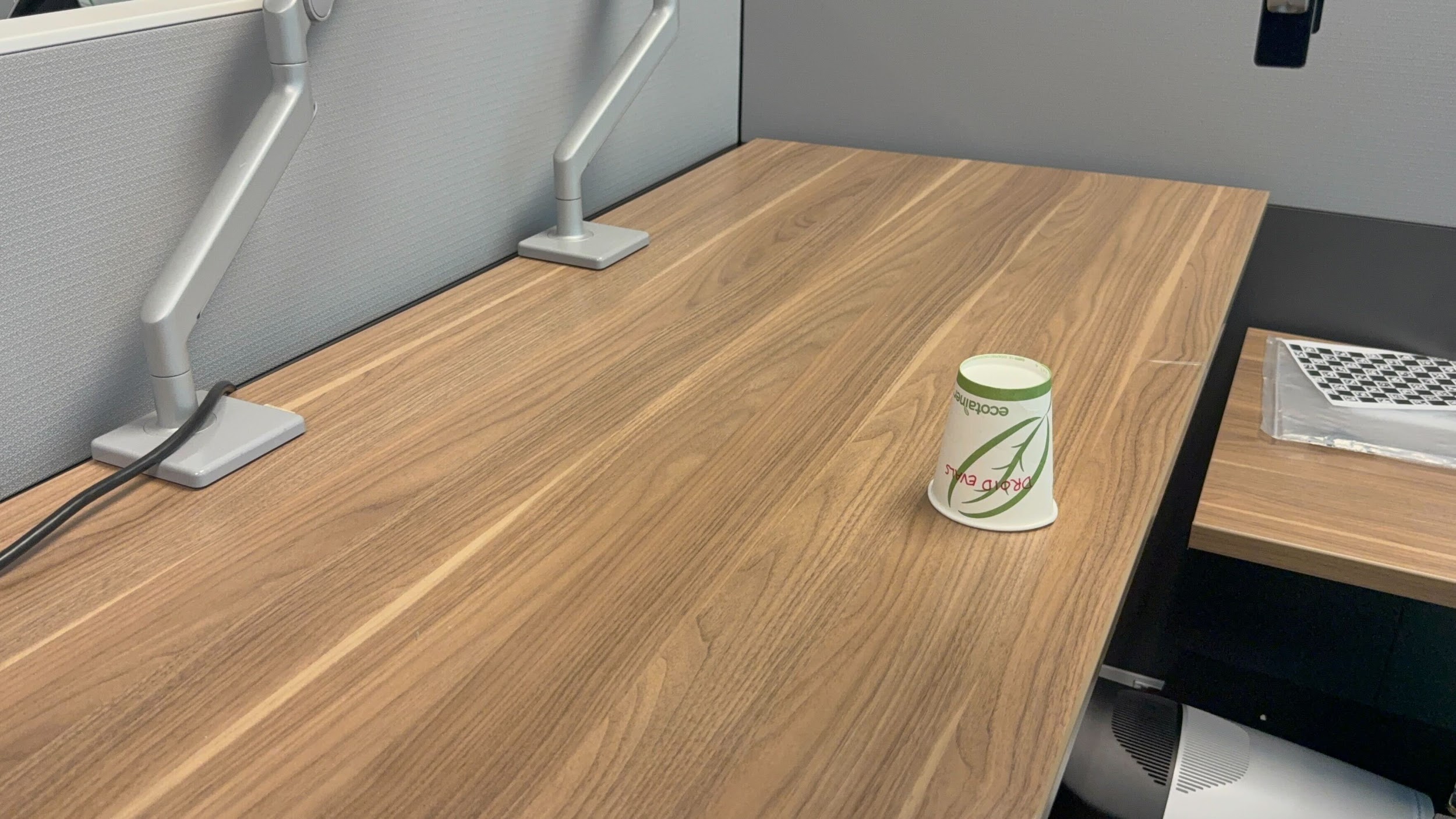} & 0.25 & 0.38 & 0.38 & \textbf{0.50} & \textbf{0.50} \\
Pick up the pink cup and pour it into the blue plate & \includegraphics[width=0.10\linewidth,height=0.035\textheight,keepaspectratio]{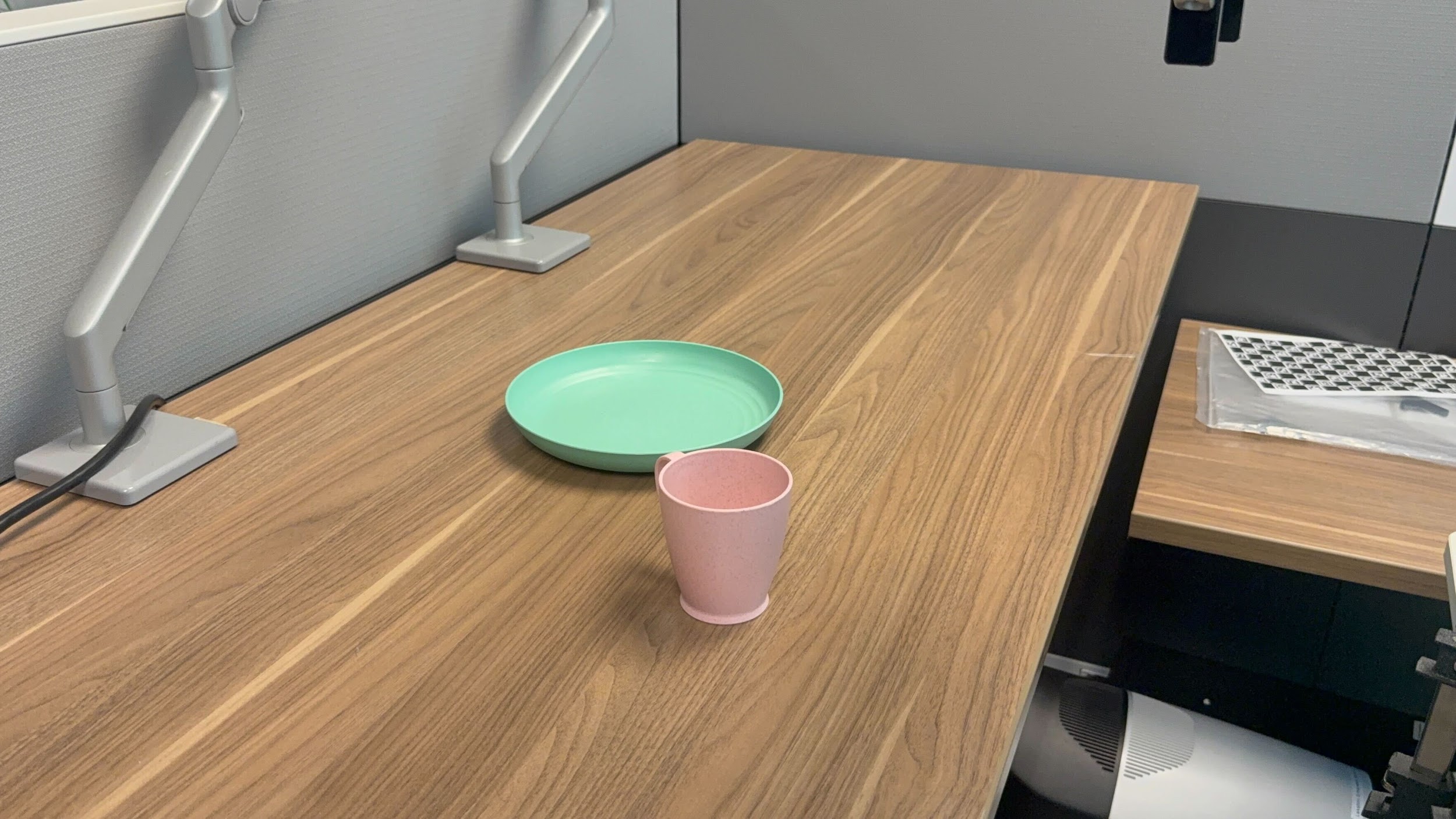} & 0.10 & 0.20 & \textbf{0.60} & \textbf{0.60} & 0.20 \\
slide plate to left side of table & \includegraphics[width=0.10\linewidth,height=0.035\textheight,keepaspectratio]{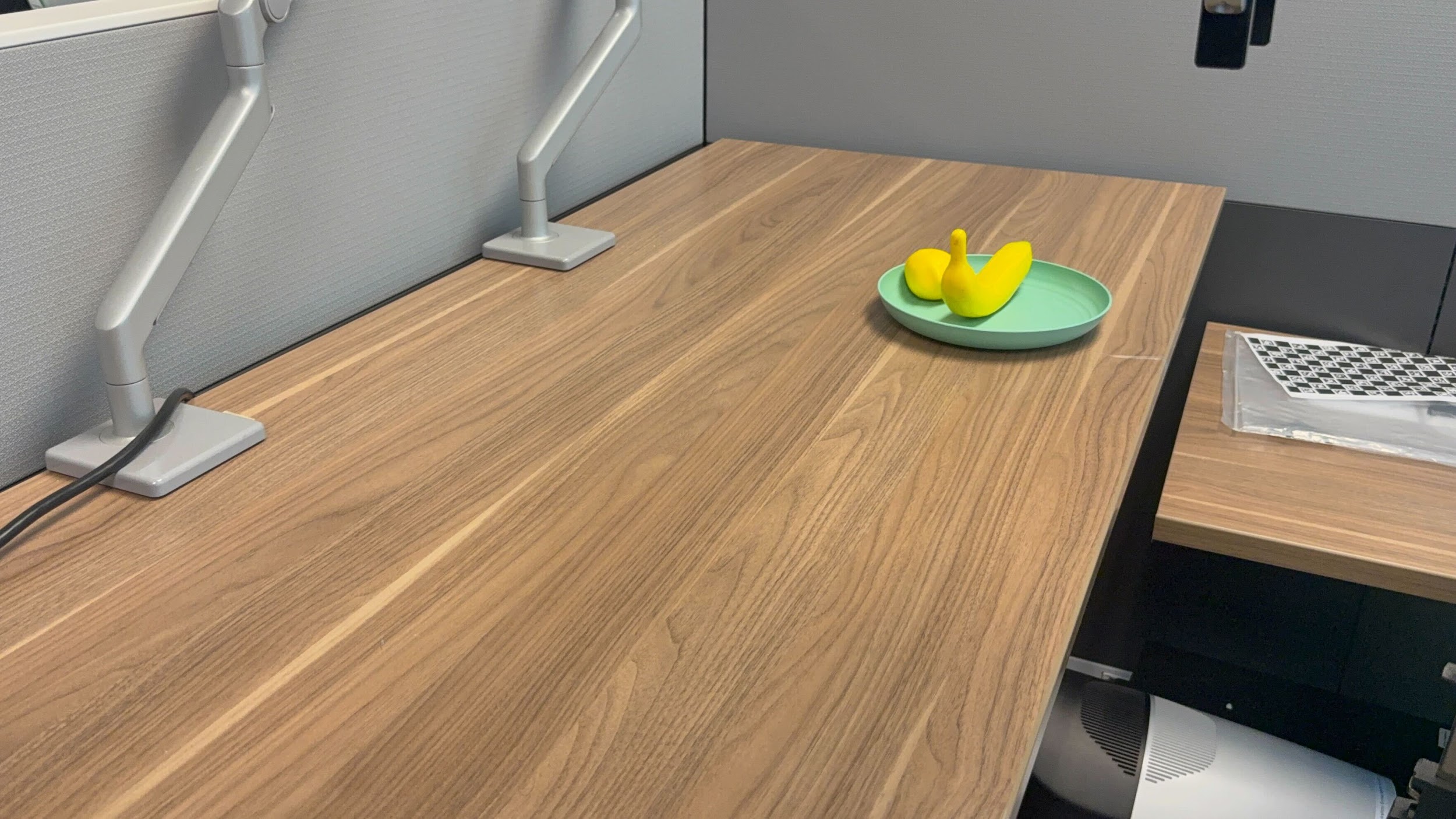} & 0.00 & 0.62 & \textbf{1.00} & \textbf{1.00} & 0.50 \\
Center cup on plate & \includegraphics[width=0.10\linewidth,height=0.035\textheight,keepaspectratio]{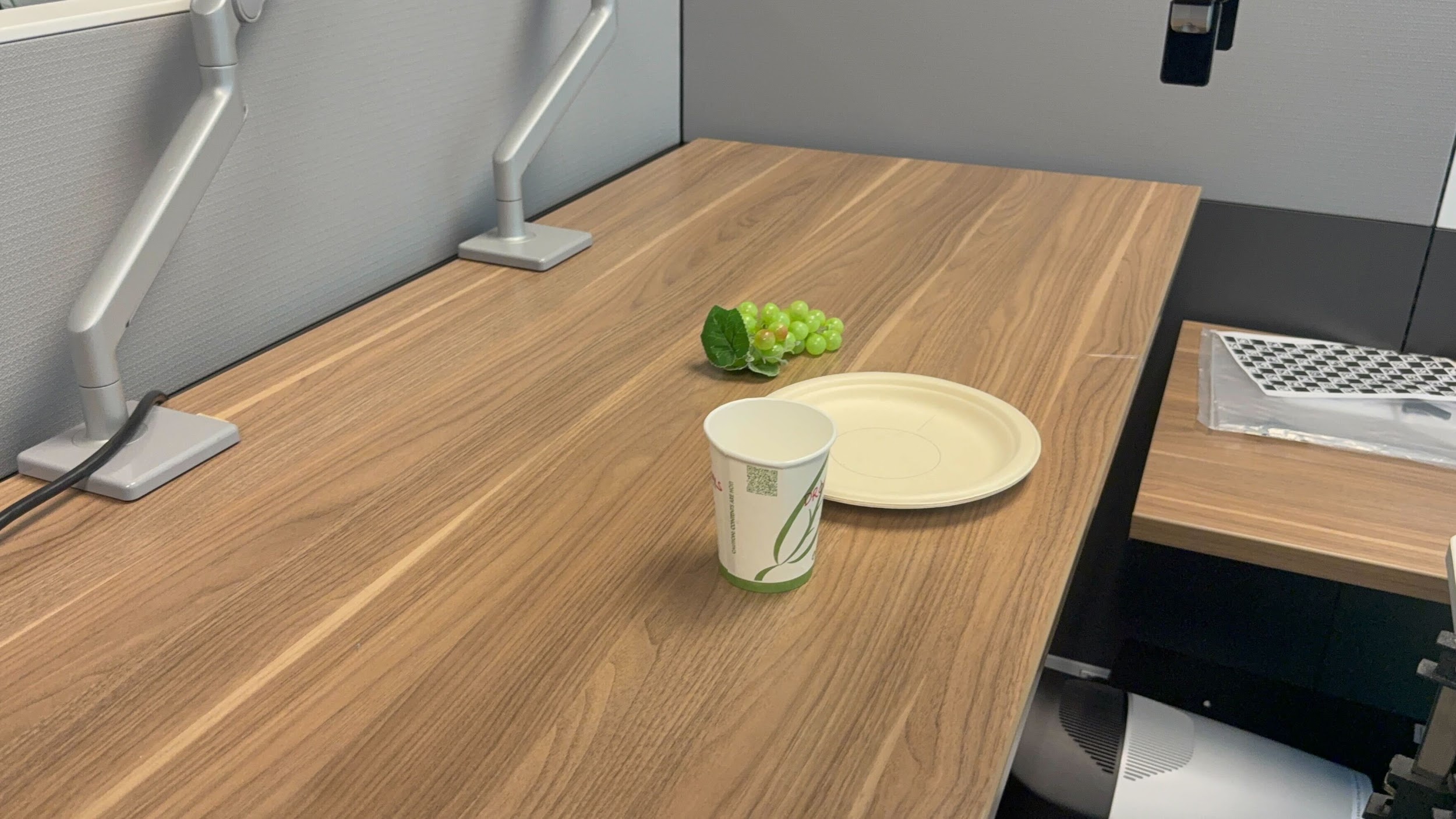} & 0.33 & \textbf{0.66} & 0.33 & \textbf{0.66} & \textbf{0.66} \\
Cover the blue letter with the plate & \includegraphics[width=0.10\linewidth,height=0.035\textheight,keepaspectratio]{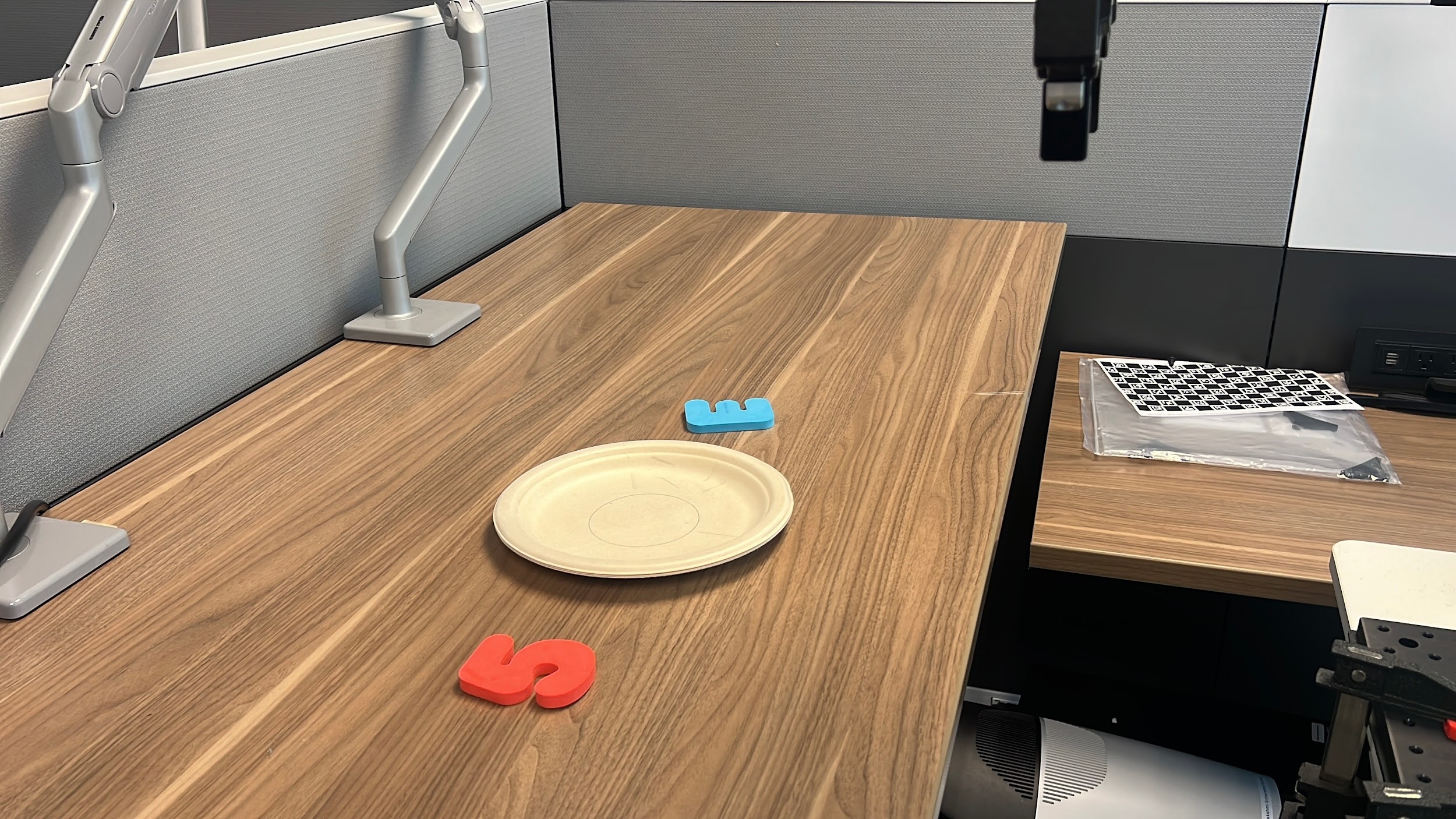} & 0.00 & 0.00 & \textbf{0.33} & 0.00 & 0.00 \\
Remove all objects from basket & \includegraphics[width=0.10\linewidth,height=0.035\textheight,keepaspectratio]{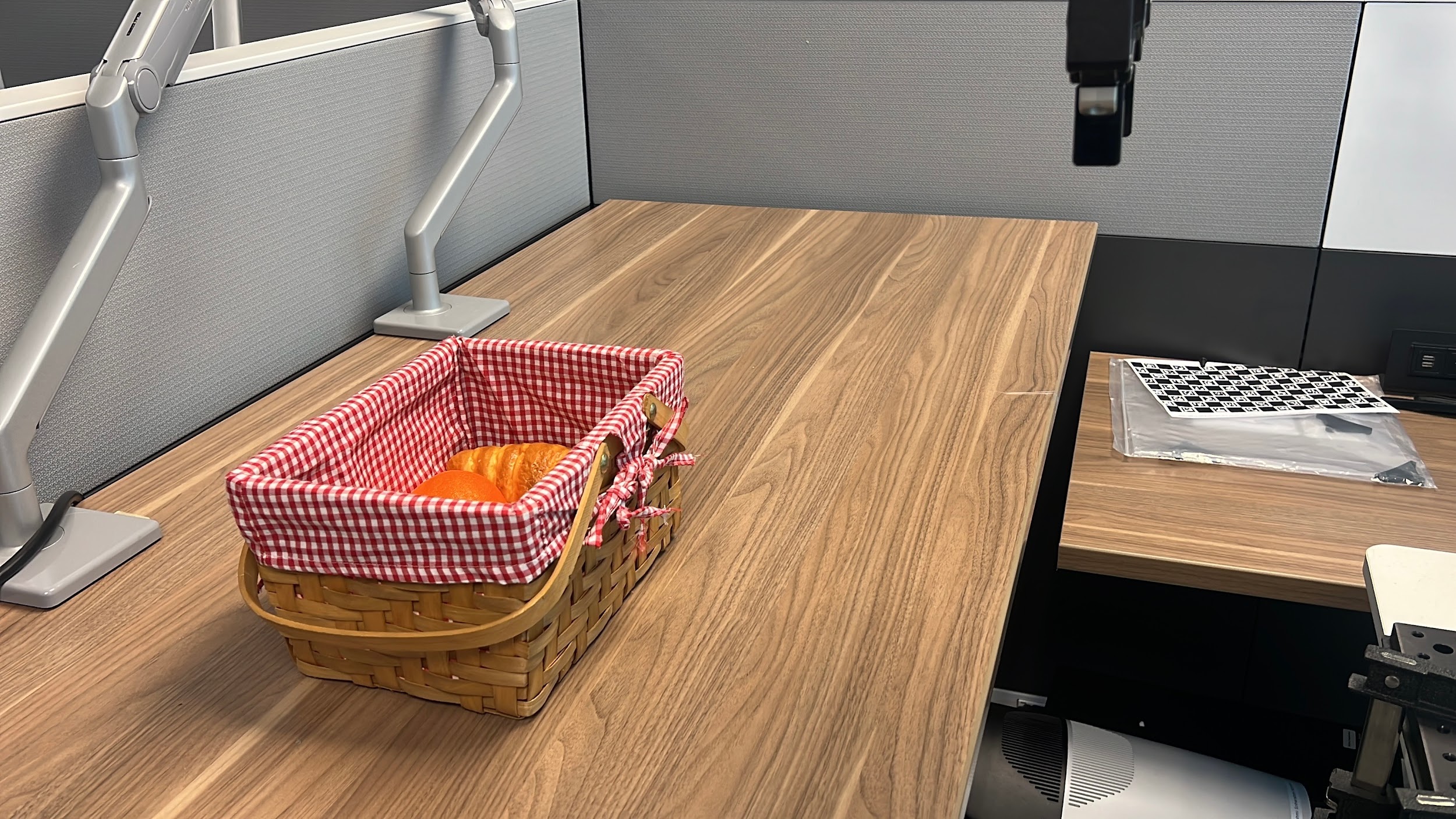} & 0.00 & 0.50 & \textbf{0.75} & 0.38 & 0.62 \\
Separate bowls and place grapes in red bowl & \includegraphics[width=0.10\linewidth,height=0.035\textheight,keepaspectratio]{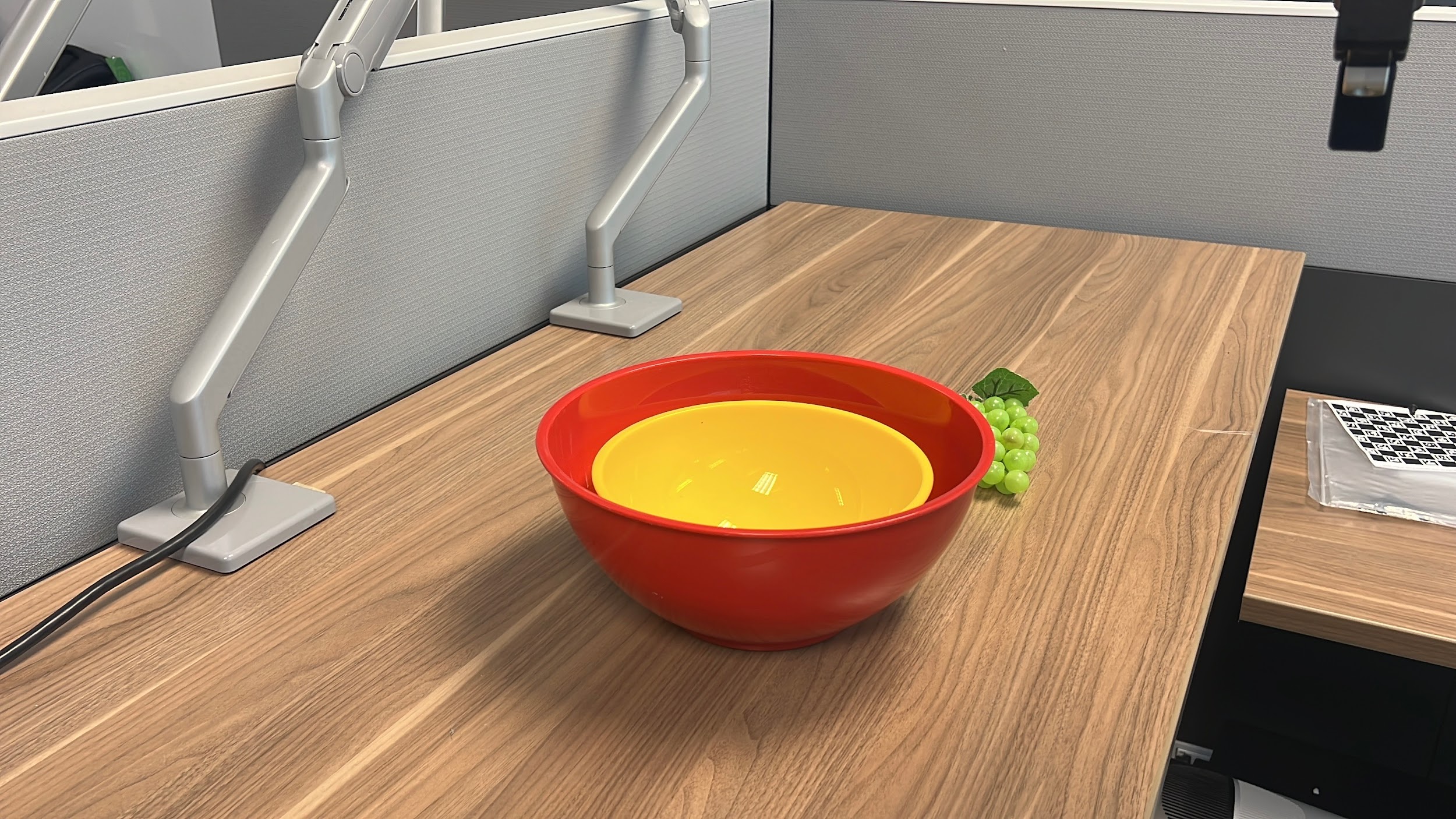} & \textbf{0.25} & \textbf{0.25} & \textbf{0.25} & 0.00 & 0.12 \\
Put all objects into the box & \includegraphics[width=0.10\linewidth,height=0.035\textheight,keepaspectratio]{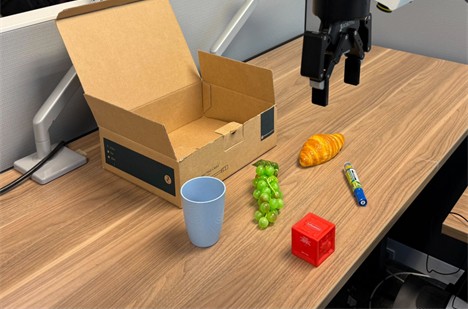} & 0.60 & 0.60 & \textbf{0.70} & 0.50 & 0.00 \\
Place cube on sandwich & \includegraphics[width=0.10\linewidth,height=0.035\textheight,keepaspectratio]{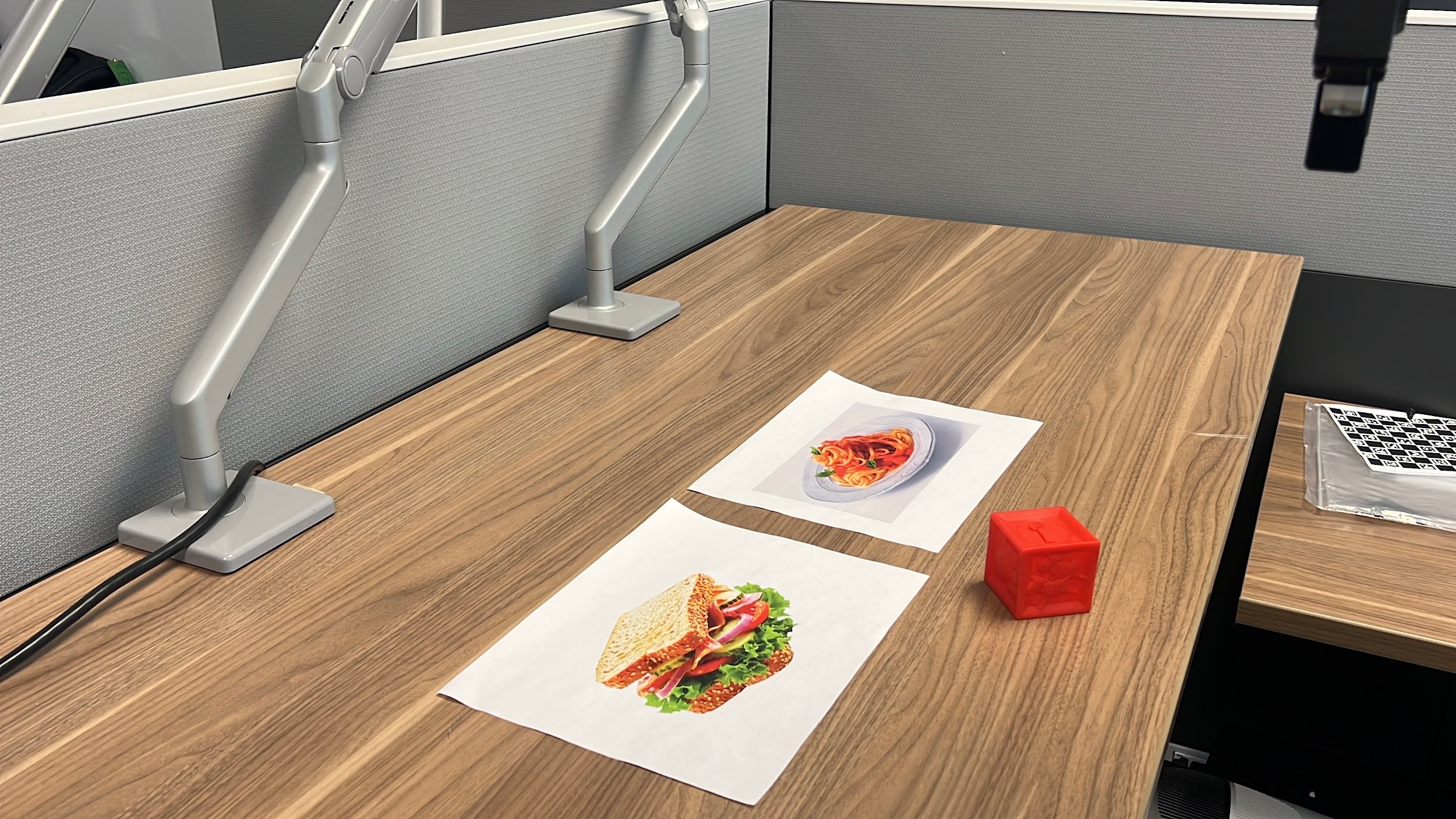} & \textbf{0.67} & 0.49 & 0.00 & \textbf{0.67} & 0.00 \\
Assemble pyramid & \includegraphics[width=0.10\linewidth,height=0.035\textheight,keepaspectratio]{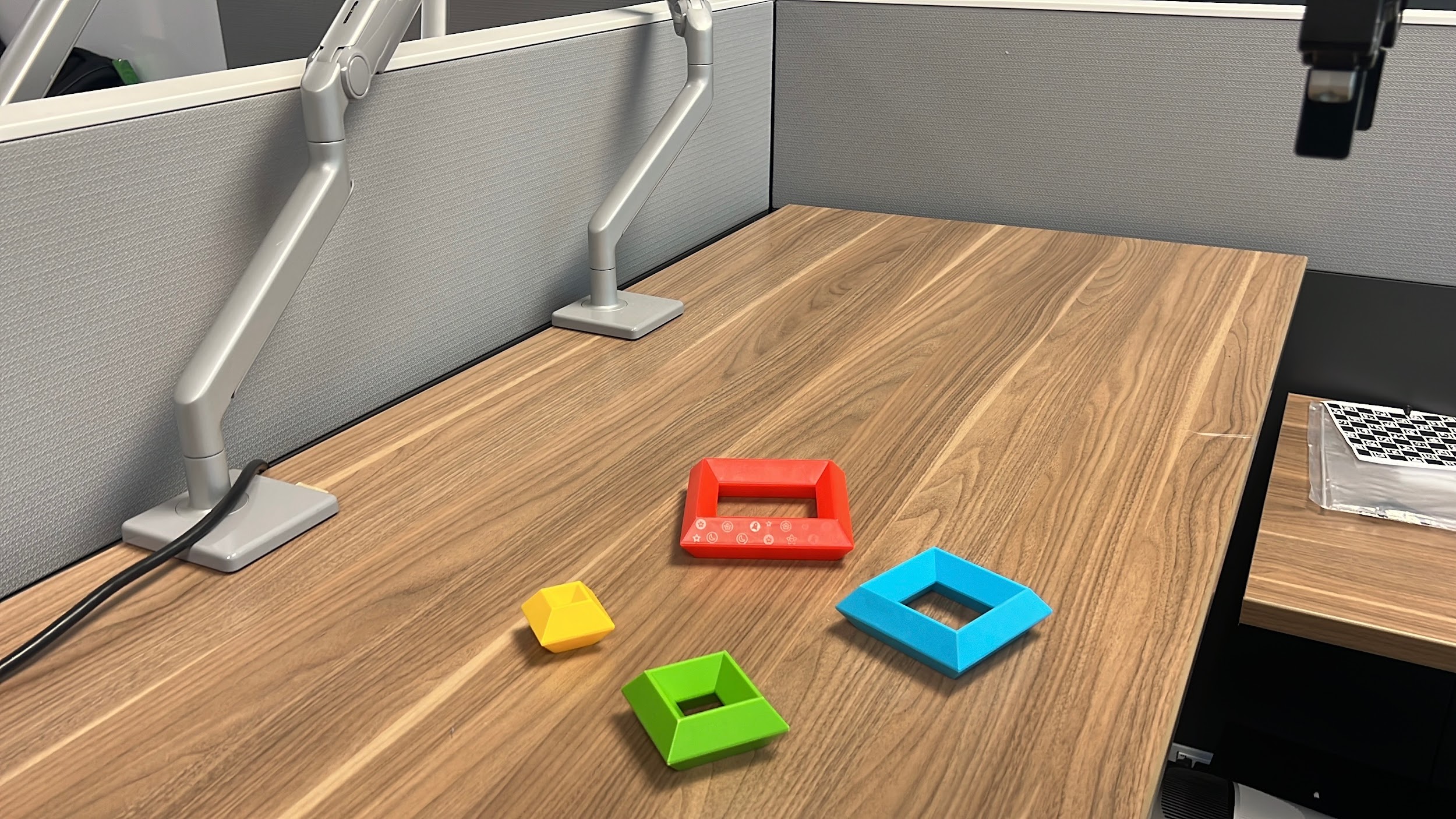} & \textbf{0.25} & \textbf{0.25} & 0.12 & \textbf{0.25} & \textbf{0.25} \\
Disassemble pyramid & \includegraphics[width=0.10\linewidth,height=0.035\textheight,keepaspectratio]{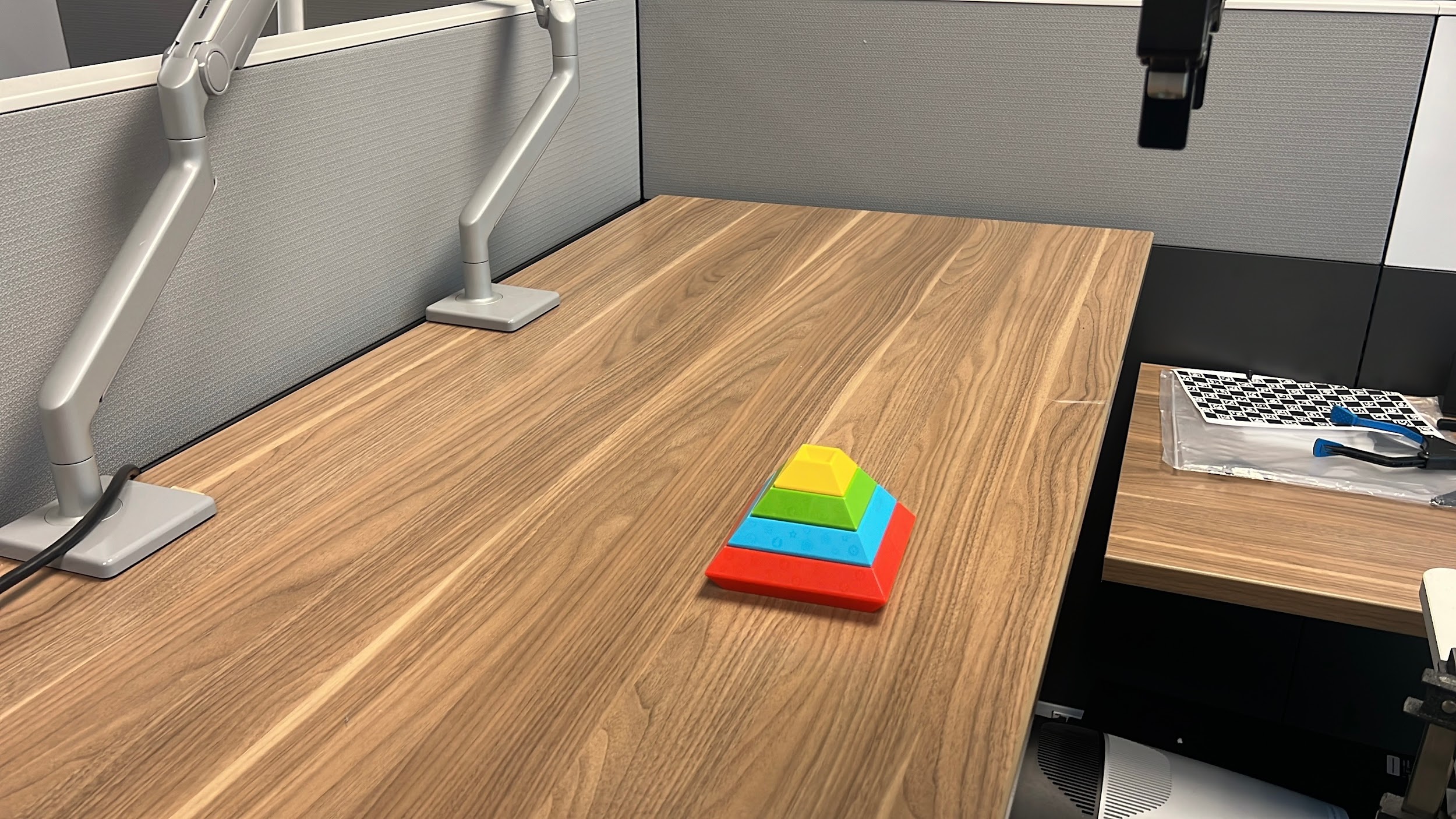} & 0.25 & \textbf{0.75} & 0.25 & 0.00 & 0.62 \\
Fold the towel & \includegraphics[width=0.10\linewidth,height=0.035\textheight,keepaspectratio]{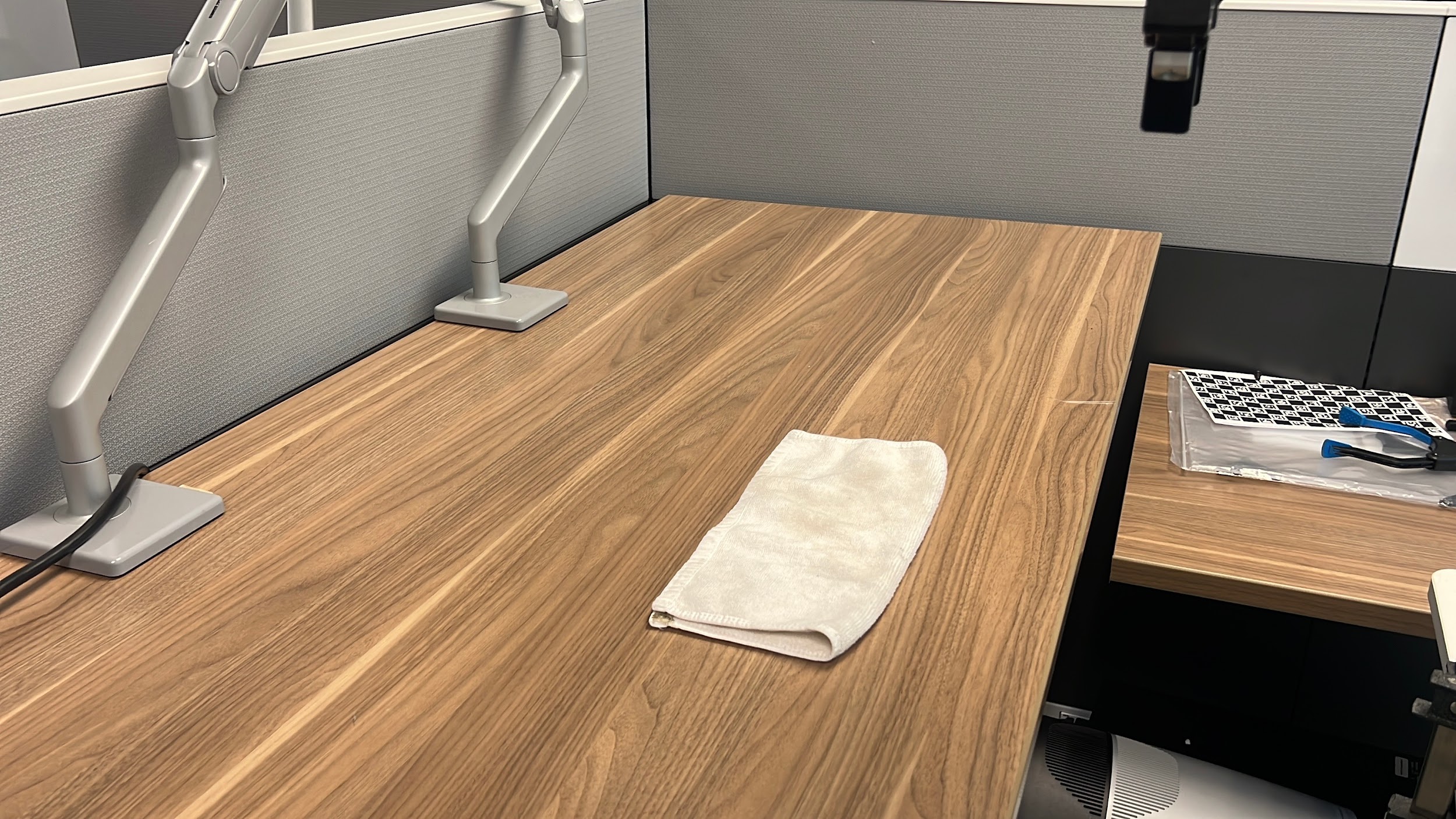} & 0.25 & \textbf{1.00} & 0.66 & 0.83 & 0.17 \\
Mix the vegetables in the bowl with the spoon & \includegraphics[width=0.10\linewidth,height=0.035\textheight,keepaspectratio]{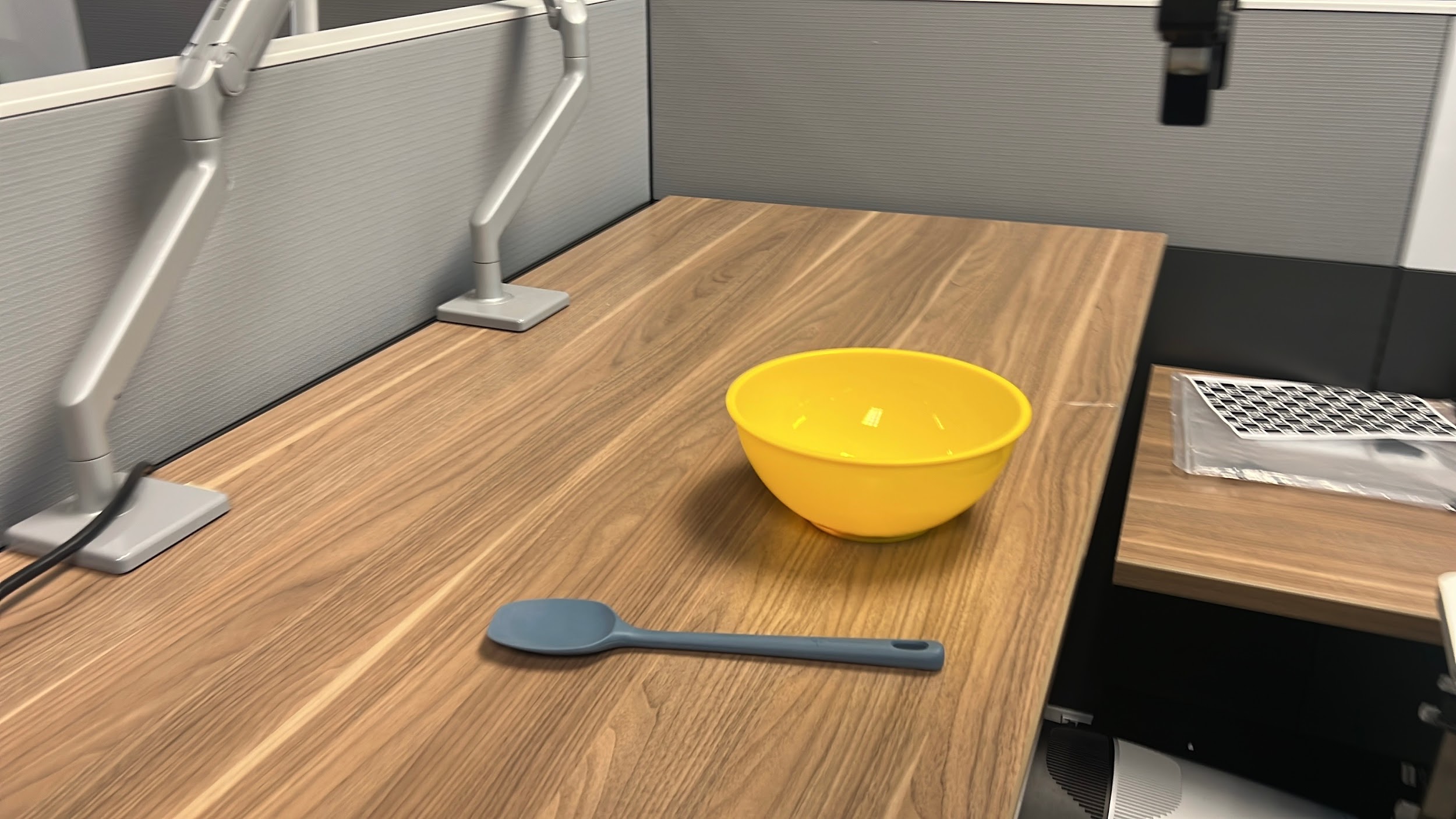} & 0.00 & 0.33 & 0.00 & \textbf{0.67} & 0.00 \\
Assemble burger & \includegraphics[width=0.10\linewidth,height=0.035\textheight,keepaspectratio]{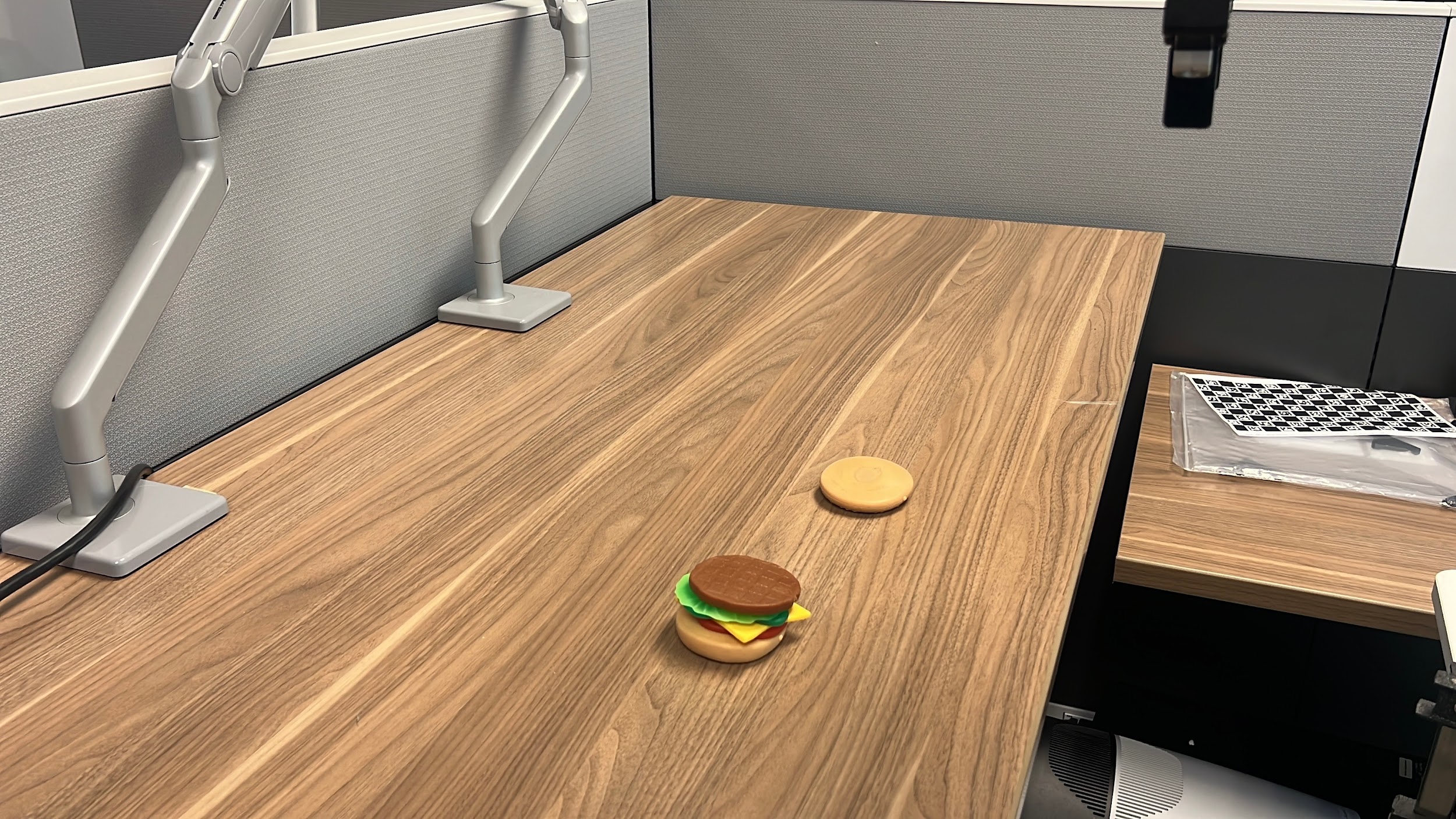} & 0.17 & 0.00 & 0.00 & \textbf{0.67} & 0.00 \\
Put the blocks in the matching hole & \includegraphics[width=0.10\linewidth,height=0.035\textheight,keepaspectratio]{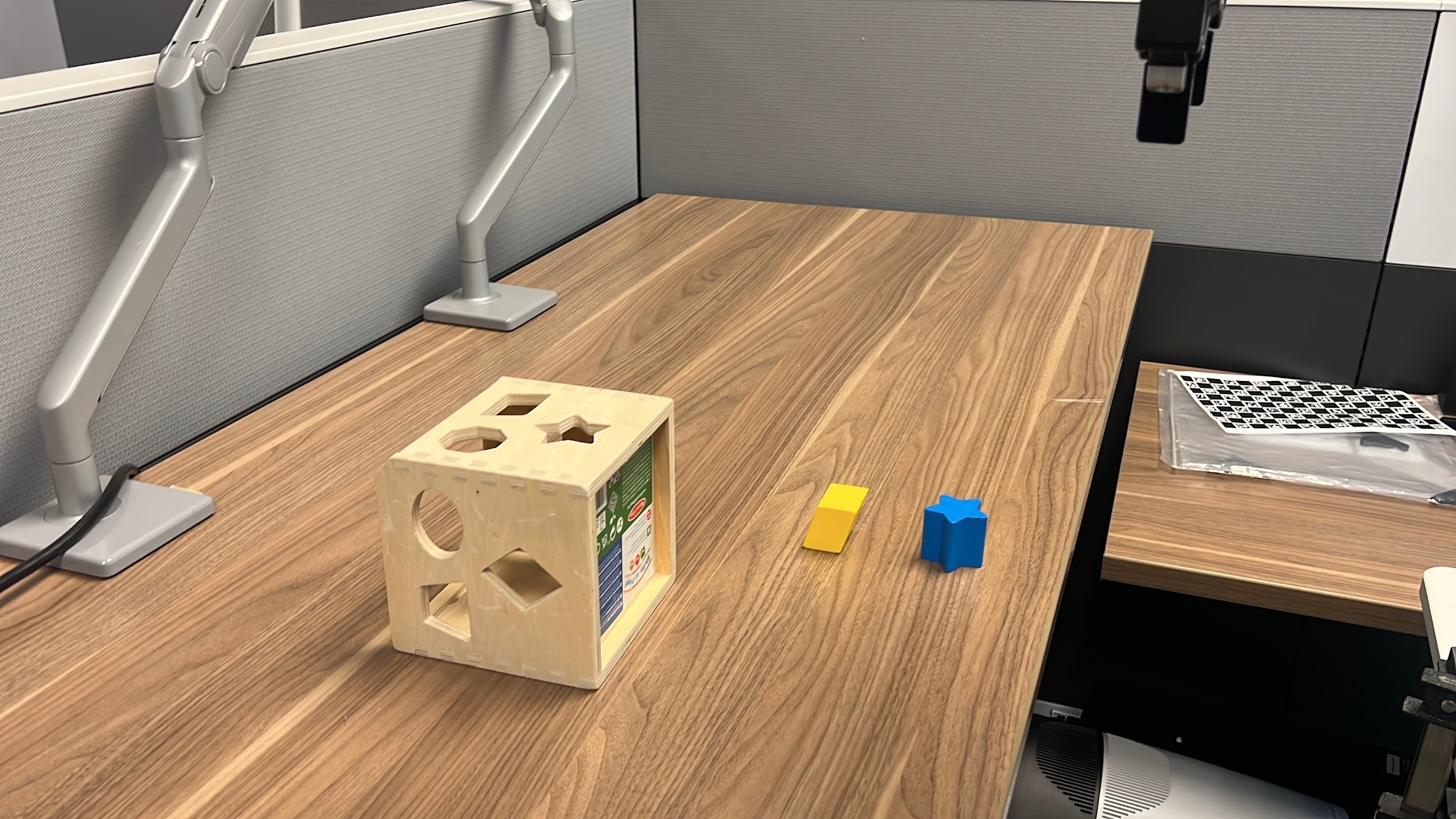} & \textbf{0.50} & 0.25 & \textbf{0.50} & \textbf{0.50} & 0.25 \\
\midrule
\textbf{Mean Task Progress} & & 0.34 & 0.44 & 0.41 & \textbf{0.54} & 0.26 \\
\bottomrule
\end{tabular}
\end{adjustbox}
\end{table}
\FloatBarrier

\FloatBarrier
\Needspace{10\baselineskip}
\section{AutoGrader Reliability and Extended Evaluation}

\subsection{Prompt Template}
\label{sec:autograder-prompt}
AutoGrader scores each rubric milestone with an independent, stateless VLM query. Each query contains the temporally batched frame context followed by a single milestone question from the AutoPrompter rubric; the grader is not shown the task instruction or the other milestones. A ``Yes'' answer awards that milestone's weight, and the trajectory score is the sum of awarded weights. We reproduce the template below.

\begin{tcolorbox}[
    colback=gray!5,
    colframe=gray!50,
    title=AutoGrader milestone-grading prompt,
    breakable,
    enhanced
]
\begin{Verbatim}[
    breaklines,
    breakanywhere,
    breaksymbolleft=,
    breakindent=2em,
    fontsize=\scriptsize,
    baselinestretch=0.84,
    fontfamily=tt
]
I'm showing you video frames organized into labeled batches from a video.

VIDEO BATCHES:
  - Batch 1: {start}s - {end}s ({n} frames)
  - Batch 2: {start}s - {end}s ({n} frames)
  - ... one line per batch ...

INSTRUCTIONS:
- When answering questions, ALWAYS reference the relevant batch number(s)
- Use format "In Batch N (X.Xs - Y.Ys)..." when describing events
- If an action spans multiple batches, mention all relevant batches
- Be specific about which frames/timestamps show key events

The frames follow below, organized by batch:

==================================================
BATCH 1 ({start}s - {end}s)
==================================================
[image frames for batch 1]

==================================================
BATCH 2 ({start}s - {end}s)
==================================================
[image frames for batch 2]

... repeated for every batch ...

Based on the video frames shown, answer the following question.

QUESTION: {rubric milestone description}

IMPORTANT: 
1. Start your response with either "Yes" or "No"
2. Then provide a detailed explanation with SPECIFIC TIMESTAMPS and batch numbers
3. Reference exact batch numbers and time ranges where you observed the action
   (e.g., "In Batch 3 (5.2s - 7.8s), I observed the robot...")
4. Be precise about which frames/timestamps show key events
5. Include timestamps for both when actions started and completed

Your response format:
[Yes/No]. [Detailed explanation with specific batch numbers, timestamps, and frame references]
\end{Verbatim}

\end{tcolorbox}

\subsection{AutoGrader Audit}
\begin{figure}[H]
\centering
\begin{tikzpicture}
\begin{axis}[
    width=10cm, height=6.5cm,
    enlargelimits=false,
    axis on top,
    xmin=-0.5, xmax=6.5, ymin=-0.5, ymax=4.5,
    xtick={0,1,2,3,4,5,6},
    xticklabels={state flip, spatial, transfer, articulation, precision, contact rich, reposing},
    xticklabel style={rotate=45, anchor=north east, font=\footnotesize},
    ytick={0,1,2,3,4},
    yticklabels={Pi05, MolmoAct2, GR00T N1.7, GR00T N1.6, DreamZero},
    yticklabel style={font=\footnotesize},
    tick style={draw=none},
    title={Grader F1 Pass per (policy $\times$ skill)},
    title style={font=\small},
    point meta min=0, point meta max=100,
    colormap={RdYlGn}{
        rgb255=(165,0,38) rgb255=(215,48,39) rgb255=(244,109,67)
        rgb255=(253,174,97) rgb255=(254,224,139) rgb255=(255,255,191)
        rgb255=(217,239,139) rgb255=(166,217,106) rgb255=(102,189,99)
        rgb255=(26,152,80) rgb255=(0,104,55)
    },
    colorbar,
    colorbar style={
        width=0.3cm, font=\footnotesize, ylabel={F1 Pass (0--1)},
        ytick={0,20,40,60,80,100},
        yticklabels={0.0,0.2,0.4,0.6,0.8,1.0},
    },
]
\addplot[matrix plot*, point meta=explicit, mesh/cols=7] table[meta=v] {
x y v
0 4 90
1 4 71
2 4 97
3 4 64
4 4 67
5 4 86
6 4 88
0 3 40
1 3 29
2 3 29
3 3 18
4 3 20
5 3 0
6 3 22
0 2 44
1 2 100
2 2 93
3 2 75
4 2 50
5 2 87
6 2 92
0 1 67
1 1 74
2 1 67
3 1 50
4 1 57
5 1 50
6 1 83
0 0 83
1 0 92
2 0 94
3 0 53
4 0 75
5 0 85
6 0 44
};
\pgfplotsextra{%
\foreach \cx/\cy/\cv in {%
    0/4/90, 1/4/71, 2/4/97, 3/4/64, 4/4/67, 5/4/86, 6/4/88,
    0/3/40, 1/3/29, 2/3/29, 3/3/18, 4/3/20, 5/3/0,  6/3/22,
    0/2/44, 1/2/100,2/2/93, 3/2/75, 4/2/50, 5/2/87, 6/2/92,
    0/1/67, 1/1/74, 2/1/67, 3/1/50, 4/1/57, 5/1/50, 6/1/83,
    0/0/83, 1/0/92, 2/0/94, 3/0/53, 4/0/75, 5/0/85, 6/0/44%
} {%
    \node[font=\footnotesize] at (axis cs:\cx,\cy) {\cv};%
}%
}
\end{axis}
\end{tikzpicture}
\caption{\textbf{Per-policy, per-skill grader F1 (pass class).} Values are \%; color encodes F1 on a $[0,1]$ scale (red--yellow--green).}
\label{fig:graderHeatmap}
\end{figure}
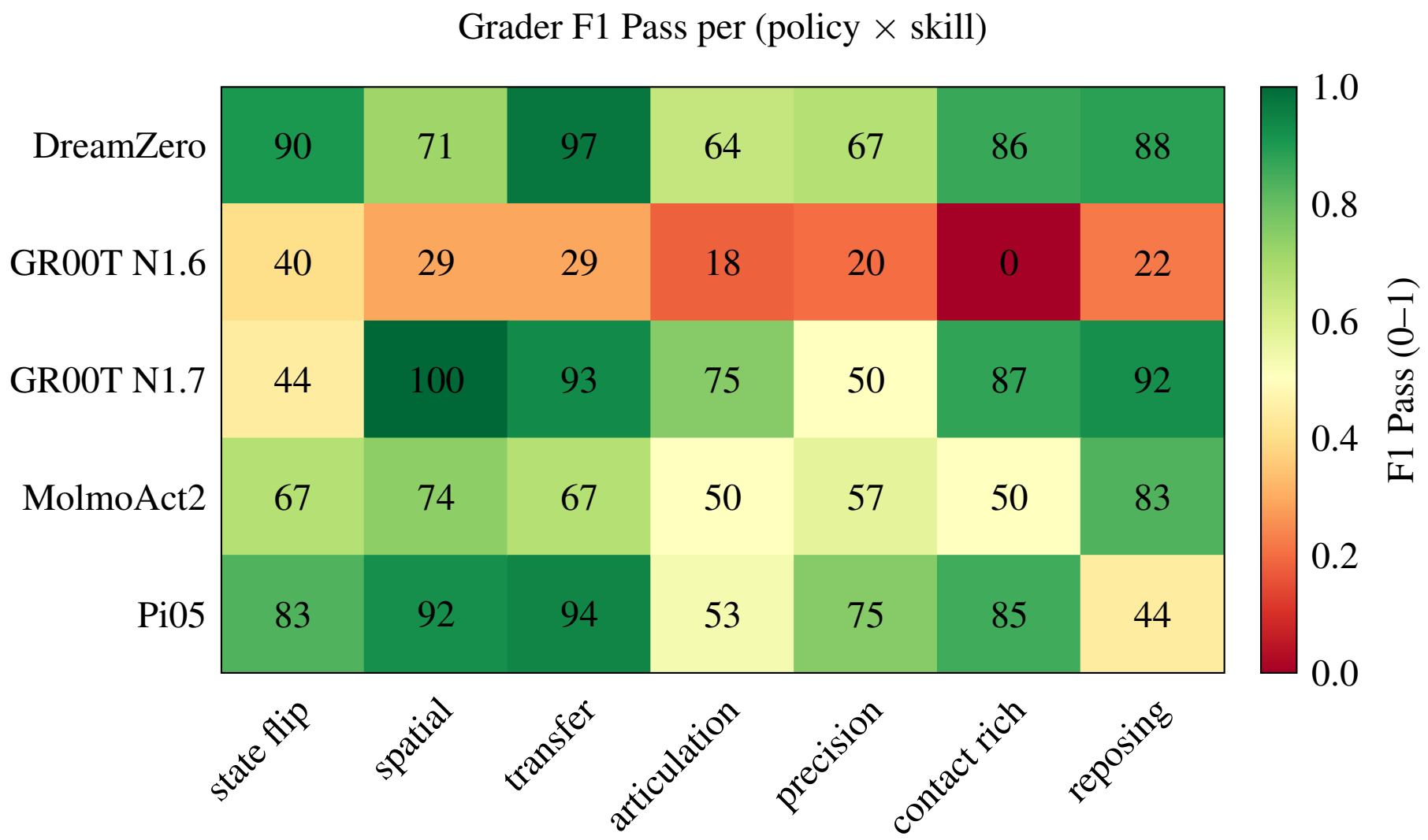

\suppressfloats[t] 

%
\begin{figure}[!htbp]
\centering
\begin{adjustbox}{max width=\linewidth}
\providecolor{rg0}{HTML}{A50026}
\providecolor{rg1}{HTML}{D73027}
\providecolor{rg2}{HTML}{F46D43}
\providecolor{rg3}{HTML}{FDAE61}
\providecolor{rg4}{HTML}{FEE08B}
\providecolor{rg5}{HTML}{FFFFBF}
\providecolor{rg6}{HTML}{D9EF8B}
\providecolor{rg7}{HTML}{A6D96A}
\providecolor{rg8}{HTML}{66BD63}
\providecolor{rg9}{HTML}{1A9850}
\providecolor{rg10}{HTML}{006837}
\begin{tikzpicture}[
  font=\footnotesize,
  cell/.style={
    minimum width=1.55cm, minimum height=1.00cm,
    anchor=center, inner sep=0pt, align=center,
    text=white, draw=white, line width=0.8pt,
  },
]
\matrix[
  matrix of nodes, nodes=cell,
  column sep=-\pgflinewidth, row sep=-\pgflinewidth, inner sep=0pt,
] (HM) {
  |[fill=rg9, text=white]| {\shortstack{\textbf{93}\\[1pt]{\tiny 3/46 wrong}}} & |[fill=rg9, text=white]| {\shortstack{\textbf{89}\\[1pt]{\tiny 5/44 wrong}}} & |[fill=rg9, text=white]| {\shortstack{\textbf{95}\\[1pt]{\tiny 2/38 wrong}}} & |[fill=rg9, text=white]| {\shortstack{\textbf{85}\\[1pt]{\tiny 7/47 wrong}}} & |[fill=rg9, text=white]| {\shortstack{\textbf{86}\\[1pt]{\tiny 6/44 wrong}}} & |[fill=rg9, text=white]| {\shortstack{\textbf{88}\\[1pt]{\tiny 4/34 wrong}}} & |[fill=rg9, text=white]| {\shortstack{\textbf{92}\\[1pt]{\tiny 3/37 wrong}}} \\
  |[fill=rg9, text=white]| {\shortstack{\textbf{88}\\[1pt]{\tiny 5/40 wrong}}} & |[fill=rg9, text=white]| {\shortstack{\textbf{91}\\[1pt]{\tiny 4/43 wrong}}} & |[fill=rg9, text=white]| {\shortstack{\textbf{89}\\[1pt]{\tiny 4/35 wrong}}} & |[fill=rg9, text=white]| {\shortstack{\textbf{91}\\[1pt]{\tiny 4/46 wrong}}} & |[fill=rg9, text=white]| {\shortstack{\textbf{90}\\[1pt]{\tiny 5/48 wrong}}} & |[fill=rg9, text=white]| {\shortstack{\textbf{88}\\[1pt]{\tiny 5/42 wrong}}} & |[fill=rg8, text=black]| {\shortstack{\textbf{84}\\[1pt]{\tiny 7/45 wrong}}} \\
  |[fill=rg9, text=white]| {\shortstack{\textbf{86}\\[1pt]{\tiny 6/44 wrong}}} & |[fill=rg10, text=white]| {\shortstack{\textbf{100}\\[1pt]{\tiny 0/42 wrong}}} & |[fill=rg10, text=white]| {\shortstack{\textbf{98}\\[1pt]{\tiny 1/45 wrong}}} & |[fill=rg9, text=white]| {\shortstack{\textbf{87}\\[1pt]{\tiny 6/46 wrong}}} & |[fill=rg8, text=black]| {\shortstack{\textbf{82}\\[1pt]{\tiny 8/44 wrong}}} & |[fill=rg9, text=white]| {\shortstack{\textbf{92}\\[1pt]{\tiny 3/40 wrong}}} & |[fill=rg9, text=white]| {\shortstack{\textbf{86}\\[1pt]{\tiny 6/42 wrong}}} \\
  |[fill=rg10, text=white]| {\shortstack{\textbf{95}\\[1pt]{\tiny 2/43 wrong}}} & |[fill=rg9, text=white]| {\shortstack{\textbf{88}\\[1pt]{\tiny 5/43 wrong}}} & |[fill=rg9, text=white]| {\shortstack{\textbf{92}\\[1pt]{\tiny 3/37 wrong}}} & |[fill=rg9, text=white]| {\shortstack{\textbf{89}\\[1pt]{\tiny 5/45 wrong}}} & |[fill=rg10, text=white]| {\shortstack{\textbf{95}\\[1pt]{\tiny 2/41 wrong}}} & |[fill=rg10, text=white]| {\shortstack{\textbf{98}\\[1pt]{\tiny 1/44 wrong}}} & |[fill=rg9, text=white]| {\shortstack{\textbf{93}\\[1pt]{\tiny 3/43 wrong}}} \\
  |[fill=rg9, text=white]| {\shortstack{\textbf{94}\\[1pt]{\tiny 3/49 wrong}}} & |[fill=rg10, text=white]| {\shortstack{\textbf{95}\\[1pt]{\tiny 2/42 wrong}}} & |[fill=rg9, text=white]| {\shortstack{\textbf{91}\\[1pt]{\tiny 4/46 wrong}}} & |[fill=rg9, text=white]| {\shortstack{\textbf{85}\\[1pt]{\tiny 7/47 wrong}}} & |[fill=rg8, text=black]| {\shortstack{\textbf{83}\\[1pt]{\tiny 8/48 wrong}}} & |[fill=rg9, text=white]| {\shortstack{\textbf{93}\\[1pt]{\tiny 3/41 wrong}}} & |[fill=rg9, text=white]| {\shortstack{\textbf{86}\\[1pt]{\tiny 6/43 wrong}}} \\
};

  \node[anchor=east, font=\footnotesize] at (HM-1-1.west) {DreamZero};
  \node[anchor=east, font=\footnotesize] at (HM-2-1.west) {Gr00t1.6};
  \node[anchor=east, font=\footnotesize] at (HM-3-1.west) {Gr00t1.7};
  \node[anchor=east, font=\footnotesize] at (HM-4-1.west) {MolmoAct2};
  \node[anchor=east, font=\footnotesize] at (HM-5-1.west) {Pi05};

  \node[anchor=north east, font=\scriptsize, rotate=22] at (HM-5-1.south) {state flip};
  \node[anchor=north east, font=\scriptsize, rotate=22] at (HM-5-2.south) {spatial};
  \node[anchor=north east, font=\scriptsize, rotate=22] at (HM-5-3.south) {transfer};
  \node[anchor=north east, font=\scriptsize, rotate=22] at (HM-5-4.south) {articulation};
  \node[anchor=north east, font=\scriptsize, rotate=22] at (HM-5-5.south) {precision};
  \node[anchor=north east, font=\scriptsize, rotate=22] at (HM-5-6.south) {contact rich};
  \node[anchor=north east, font=\scriptsize, rotate=22] at (HM-5-7.south) {reposing};

\end{tikzpicture}
\end{adjustbox}
\caption{Grader \textbf{Accuracy} of the VLM grader versus human review, per
policy $\times$ skill tag (\(N=525\) audited rollouts, 15 per cell). Each
cell shows the accuracy (bold), with the count of criteria the grader labelled incorrectly below.
Cells are coloured by accuracy (red--yellow--green, 0--1).}
\label{fig:grader-accuracy-per-tag}
\end{figure}
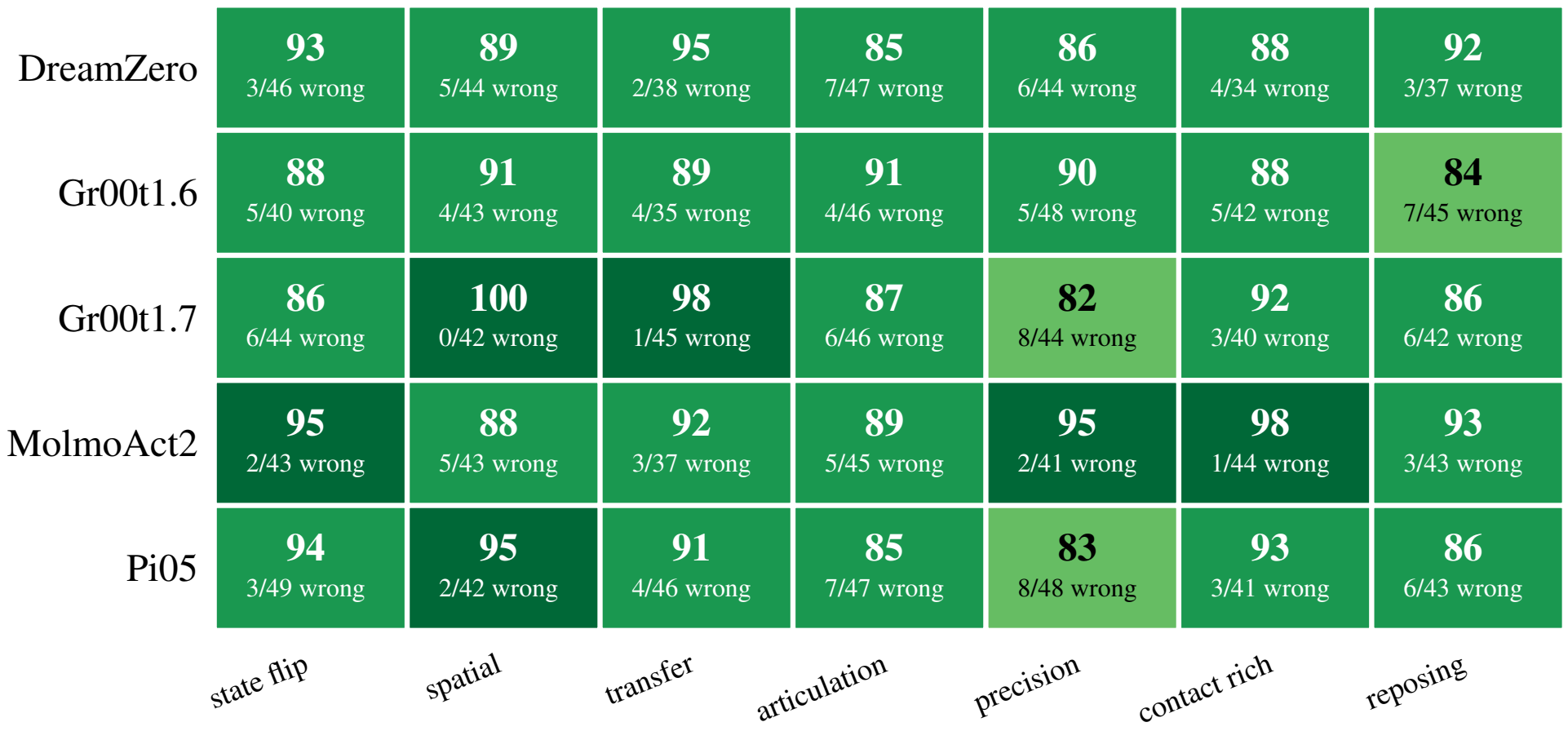

\FloatBarrier

%
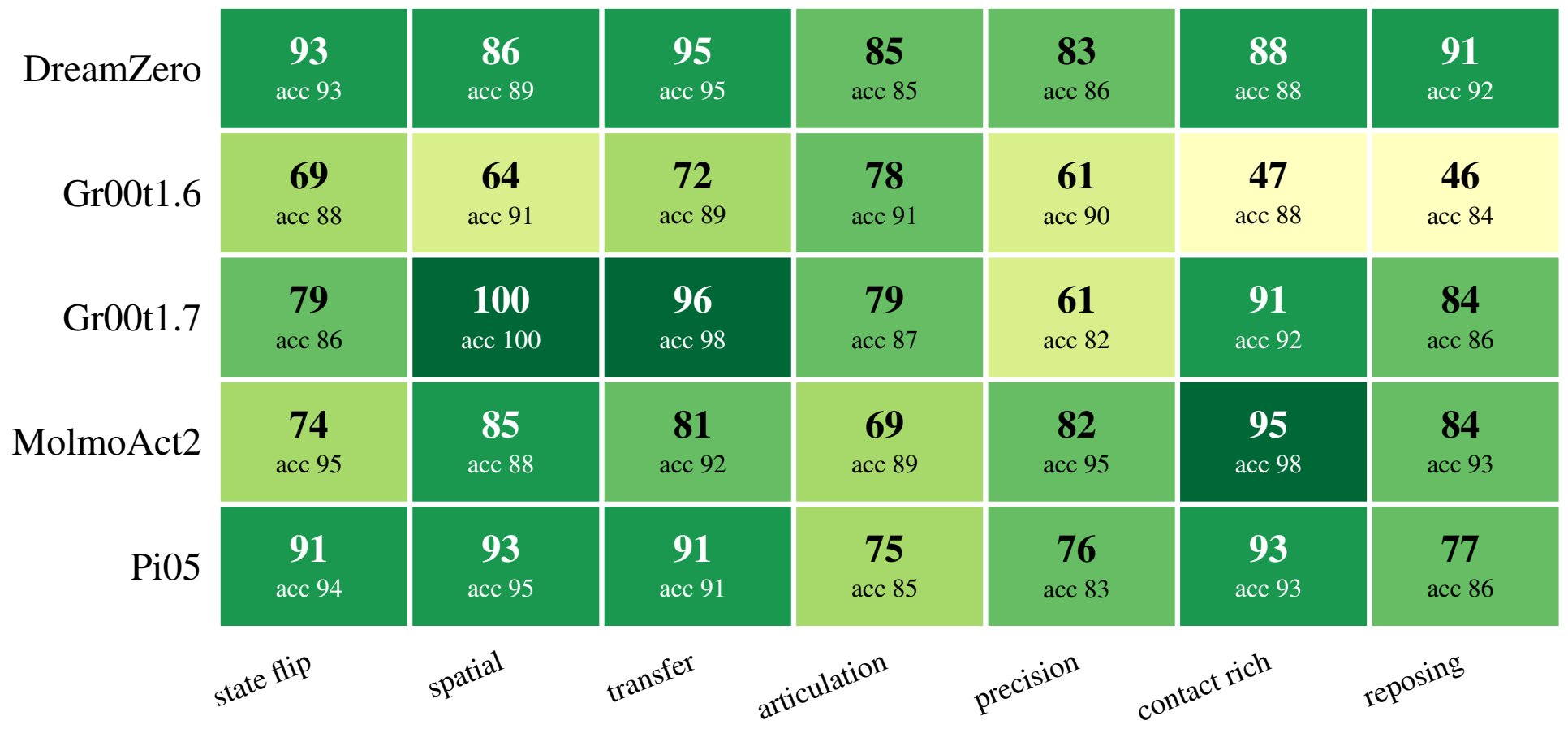
\begin{figure}[!htbp]
\centering
\begin{adjustbox}{max width=\linewidth}
\providecolor{rg0}{HTML}{A50026}
\providecolor{rg1}{HTML}{D73027}
\providecolor{rg2}{HTML}{F46D43}
\providecolor{rg3}{HTML}{FDAE61}
\providecolor{rg4}{HTML}{FEE08B}
\providecolor{rg5}{HTML}{FFFFBF}
\providecolor{rg6}{HTML}{D9EF8B}
\providecolor{rg7}{HTML}{A6D96A}
\providecolor{rg8}{HTML}{66BD63}
\providecolor{rg9}{HTML}{1A9850}
\providecolor{rg10}{HTML}{006837}
\begin{tikzpicture}[
  font=\footnotesize,
  cell/.style={
    minimum width=1.55cm, minimum height=1.00cm,
    anchor=center, inner sep=0pt, align=center,
    text=white, draw=white, line width=0.8pt,
  },
]
\matrix[
  matrix of nodes, nodes=cell,
  column sep=-\pgflinewidth, row sep=-\pgflinewidth, inner sep=0pt,
] (HM) {
  |[fill=rg9, text=white]| {\shortstack{\textbf{93}\\[1pt]{\tiny acc 93}}} & |[fill=rg9, text=white]| {\shortstack{\textbf{86}\\[1pt]{\tiny acc 89}}} & |[fill=rg9, text=white]| {\shortstack{\textbf{95}\\[1pt]{\tiny acc 95}}} & |[fill=rg8, text=black]| {\shortstack{\textbf{85}\\[1pt]{\tiny acc 85}}} & |[fill=rg8, text=black]| {\shortstack{\textbf{83}\\[1pt]{\tiny acc 86}}} & |[fill=rg9, text=white]| {\shortstack{\textbf{88}\\[1pt]{\tiny acc 88}}} & |[fill=rg9, text=white]| {\shortstack{\textbf{91}\\[1pt]{\tiny acc 92}}} \\
  |[fill=rg7, text=black]| {\shortstack{\textbf{69}\\[1pt]{\tiny acc 88}}} & |[fill=rg6, text=black]| {\shortstack{\textbf{64}\\[1pt]{\tiny acc 91}}} & |[fill=rg7, text=black]| {\shortstack{\textbf{72}\\[1pt]{\tiny acc 89}}} & |[fill=rg8, text=black]| {\shortstack{\textbf{78}\\[1pt]{\tiny acc 91}}} & |[fill=rg6, text=black]| {\shortstack{\textbf{61}\\[1pt]{\tiny acc 90}}} & |[fill=rg5, text=black]| {\shortstack{\textbf{47}\\[1pt]{\tiny acc 88}}} & |[fill=rg5, text=black]| {\shortstack{\textbf{46}\\[1pt]{\tiny acc 84}}} \\
  |[fill=rg8, text=black]| {\shortstack{\textbf{79}\\[1pt]{\tiny acc 86}}} & |[fill=rg10, text=white]| {\shortstack{\textbf{100}\\[1pt]{\tiny acc 100}}} & |[fill=rg10, text=white]| {\shortstack{\textbf{96}\\[1pt]{\tiny acc 98}}} & |[fill=rg8, text=black]| {\shortstack{\textbf{79}\\[1pt]{\tiny acc 87}}} & |[fill=rg6, text=black]| {\shortstack{\textbf{61}\\[1pt]{\tiny acc 82}}} & |[fill=rg9, text=white]| {\shortstack{\textbf{91}\\[1pt]{\tiny acc 92}}} & |[fill=rg8, text=black]| {\shortstack{\textbf{84}\\[1pt]{\tiny acc 86}}} \\
  |[fill=rg7, text=black]| {\shortstack{\textbf{74}\\[1pt]{\tiny acc 95}}} & |[fill=rg9, text=white]| {\shortstack{\textbf{85}\\[1pt]{\tiny acc 88}}} & |[fill=rg8, text=black]| {\shortstack{\textbf{81}\\[1pt]{\tiny acc 92}}} & |[fill=rg7, text=black]| {\shortstack{\textbf{69}\\[1pt]{\tiny acc 89}}} & |[fill=rg8, text=black]| {\shortstack{\textbf{82}\\[1pt]{\tiny acc 95}}} & |[fill=rg10, text=white]| {\shortstack{\textbf{95}\\[1pt]{\tiny acc 98}}} & |[fill=rg8, text=black]| {\shortstack{\textbf{84}\\[1pt]{\tiny acc 93}}} \\
  |[fill=rg9, text=white]| {\shortstack{\textbf{91}\\[1pt]{\tiny acc 94}}} & |[fill=rg9, text=white]| {\shortstack{\textbf{93}\\[1pt]{\tiny acc 95}}} & |[fill=rg9, text=white]| {\shortstack{\textbf{91}\\[1pt]{\tiny acc 91}}} & |[fill=rg7, text=black]| {\shortstack{\textbf{75}\\[1pt]{\tiny acc 85}}} & |[fill=rg8, text=black]| {\shortstack{\textbf{76}\\[1pt]{\tiny acc 83}}} & |[fill=rg9, text=white]| {\shortstack{\textbf{93}\\[1pt]{\tiny acc 93}}} & |[fill=rg8, text=black]| {\shortstack{\textbf{77}\\[1pt]{\tiny acc 86}}} \\
};

  \node[anchor=east, font=\footnotesize] at (HM-1-1.west) {DreamZero};
  \node[anchor=east, font=\footnotesize] at (HM-2-1.west) {Gr00t1.6};
  \node[anchor=east, font=\footnotesize] at (HM-3-1.west) {Gr00t1.7};
  \node[anchor=east, font=\footnotesize] at (HM-4-1.west) {MolmoAct2};
  \node[anchor=east, font=\footnotesize] at (HM-5-1.west) {Pi05};

  \node[anchor=north east, font=\scriptsize, rotate=22] at (HM-5-1.south) {state flip};
  \node[anchor=north east, font=\scriptsize, rotate=22] at (HM-5-2.south) {spatial};
  \node[anchor=north east, font=\scriptsize, rotate=22] at (HM-5-3.south) {transfer};
  \node[anchor=north east, font=\scriptsize, rotate=22] at (HM-5-4.south) {articulation};
  \node[anchor=north east, font=\scriptsize, rotate=22] at (HM-5-5.south) {precision};
  \node[anchor=north east, font=\scriptsize, rotate=22] at (HM-5-6.south) {contact rich};
  \node[anchor=north east, font=\scriptsize, rotate=22] at (HM-5-7.south) {reposing};

\end{tikzpicture}
\end{adjustbox}
\caption{Grader \textbf{Macro F1} of the VLM grader versus human review, per
policy $\times$ skill tag (\(N=525\) audited rollouts, 15 per cell). Each
cell shows the macro f1 (bold), with grader--human agreement below.
Cells are coloured by macro f1 (red--yellow--green, 0--1).}
\label{fig:grader-macro-f1-per-tag}
\end{figure}

\FloatBarrier

%
%
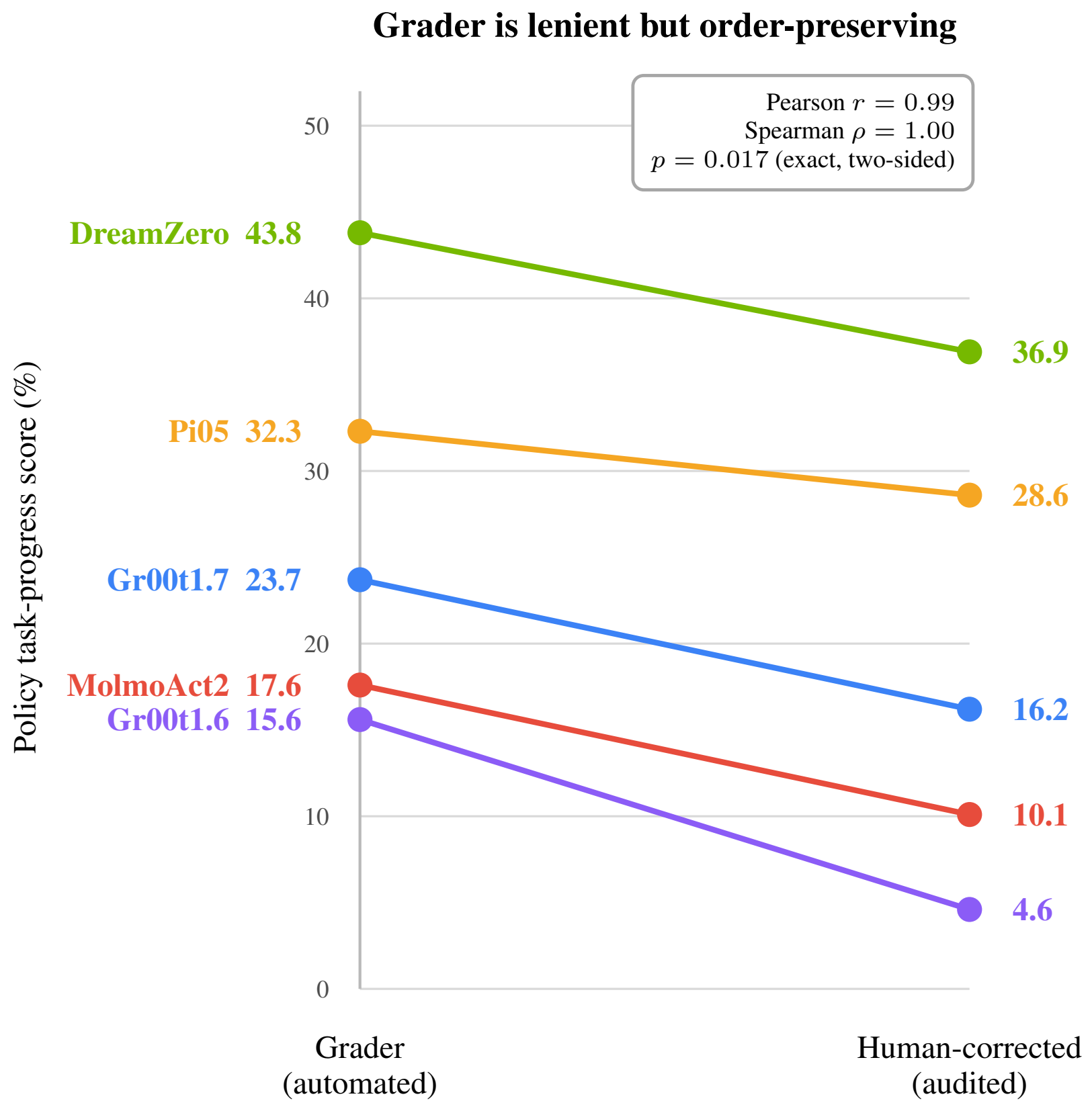
\begin{figure}[t]
\centering
\begin{adjustbox}{max width=0.85\linewidth}
\providecolor{slopeDreamZero}{HTML}{76B900}
\providecolor{slopePi05}{HTML}{F5A623}
\providecolor{slopeGrooo17}{HTML}{3B82F6}
\providecolor{slopeMolmo}{HTML}{E74C3C}
\providecolor{slopeGrooo16}{HTML}{8B5CF6}
\begin{tikzpicture}[
  x=1cm, y=0.17cm, font=\footnotesize,
  slope/.style={line width=1.6pt},
  dot/.style={circle, fill, inner sep=0pt, minimum size=6pt,
              draw=white, line width=1pt},
]
\def\xL{0}     
\def\xR{6}     

\foreach \v in {0,10,20,30,40,50}{
  \draw[gray!30, line width=0.6pt] (\xL,\v) -- (\xR,\v);
  \node[anchor=east, gray!60!black, font=\scriptsize] at (\xL-0.18,\v) {\v};
}
\draw[gray!55, line width=0.8pt] (\xL,0) -- (\xL,52);

\foreach \c/\name/\sL/\sR in {%
  slopeDreamZero/DreamZero/43.8/36.9,%
  slopePi05/Pi05/32.3/28.6,%
  slopeGrooo17/Gr00t1.7/23.7/16.2,%
  slopeMolmo/MolmoAct2/17.6/10.1,%
  slopeGrooo16/Gr00t1.6/15.6/4.6%
}{
  \draw[slope, \c] (\xL,\sL) -- (\xR,\sR);
  \node[dot, \c] at (\xL,\sL) {};
  \node[dot, \c] at (\xR,\sR) {};
  \node[anchor=east, \c, font=\footnotesize\bfseries] at (\xL-0.45,\sL) {\name\ \ \sL};
  \node[anchor=west, \c, font=\footnotesize\bfseries] at (\xR+0.30,\sR) {\sR};
}

\node[anchor=north, align=center, font=\small] at (\xL,-2.0) {Grader\\(automated)};
\node[anchor=north, align=center, font=\small] at (\xR,-2.0) {Human-corrected\\(audited)};

\node[rotate=90, anchor=south, font=\small] at (\xL-3.0,25)
  {Policy task-progress score (\%)};

\node[anchor=south, font=\normalsize\bfseries] at (3,54)
  {Grader is lenient but order-preserving};

\node[anchor=north east, align=right, font=\scriptsize,
      draw=gray!70, line width=0.9pt, rounded corners=3pt,
      fill=white, inner sep=5pt] at (\xR+0.05,53) {%
  Pearson $r = 0.99$\\
  Spearman $\rho = 1.00$\\
  $p = 0.017$ (exact, two-sided)%
};

\end{tikzpicture}
\end{adjustbox}
\caption{The VLM grader is lenient (every policy's task-progress
score drops after human correction) yet order-preserving as no lines
cross. Grader and human-corrected policy scores are near-perfectly correlated
(Pearson $r=0.99$, Spearman $\rho=1.00$; exact two-sided permutation $p=0.017$, $n=5$ policies over 525 audited rollouts), so the automated grader recovers the same policy ranking as audited
human review.}
\label{fig:grader-score-slope}
\end{figure}

\FloatBarrier
\Needspace{10\baselineskip}
\subsection{Balanced Leaderboard and Calibration}

\begin{table}[!ht]
  \centering
  \small
  \setlength{\tabcolsep}{6pt}
  \renewcommand{\arraystretch}{1.15}
  \begin{tabular}{@{}l r r r r r r@{}}
    \toprule
    \textbf{Policy} & \textbf{Rank} & $\boldsymbol{\theta}\text{ (BT)}$ & $\boldsymbol{\sigma_{\theta}}$ & $\boldsymbol{\overline{\mathrm{TP}}}$ \textbf{(\%)} & \textbf{SR (\%)} & \textbf{Pairs} \\
    \midrule
    DreamZero            & 1 & 1580 & 16 & 38.5 & 13.1 & 989 \\
    Pi05$^{\dagger}$     & 2 & 1539 &  0 & 30.7 &  7.7 & 748 \\
    Gr00tN1.7            & 3 & 1500 & 17 & 23.2 &  3.4 & 809 \\
    MolmoAct2            & 4 & 1456 & 16 & 17.3 &  2.6 & 950 \\
    Gr00tN1.6            & 5 & 1453 & 16 & 15.5 &  0.6 & 970 \\
    \bottomrule
  \end{tabular}
  \caption{\textbf{Bradley--Terry leaderboard on the balanced 1750 subset
  ($n{=}350$ rollouts per policy).} Soft wins use raw
  task-progress score, pair weights track the matched-tag count.
  $^{\dagger}$\,Pi05 is the BT anchor ($\sigma_\theta{=}0$ by construction).
  $\overline{\mathrm{TP}}$ is mean task progress across the 350 rollouts;
  SR is the perfect-rollout rate.}
  \label{tab:bt-balanced1750}
\end{table}

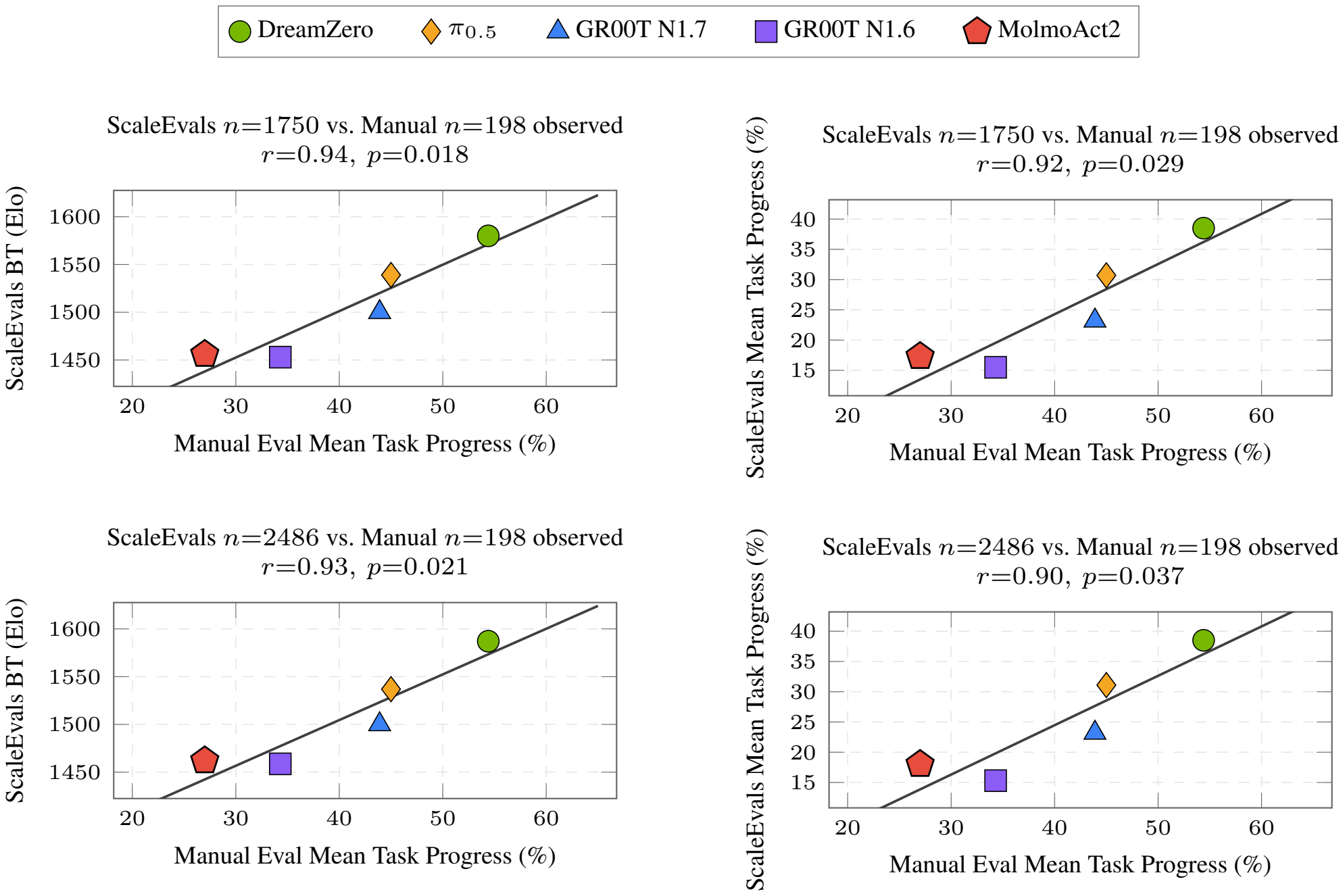
\begin{figure}[!t]
  \centering
  \pgfplotsset{
    /pgfplots/calibrationpanel/.style={
      width=0.95\linewidth, height=3.5cm,
      xlabel style={font=\scriptsize, yshift=1pt},
      ylabel style={font=\scriptsize, yshift=-1pt},
      title style={font=\scriptsize, yshift=-3pt, align=center},
      tick label style={font=\tiny},
      label style={font=\scriptsize},
      grid=both,
      major grid style={dashed, gray!20, line width=0.3pt},
      axis line style={draw=black!60, line width=0.4pt},
      tick style={draw=black!60},
      enlarge x limits=0.04, enlarge y limits=0.04,
    },
  }
  \begin{tikzpicture}
    \begin{axis}[
      hide axis,
      width=0.95\linewidth, height=8mm,
      scale only axis,
      xmin=0, xmax=1, ymin=0, ymax=1,
      legend columns=5,
      legend style={
        at={(0.5,0.5)}, anchor=center,
        draw=black!50, line width=0.3pt,
        font=\scriptsize,
        /tikz/every even column/.append style={column sep=0.4cm},
      },
    ]
      \addlegendimage{only marks, mark=*, mark size=3pt,
        color=radarDream,
        mark options={fill=radarDream,draw=black,line width=0.3pt}}
      \addlegendentry{DreamZero}
      \addlegendimage{only marks, mark=diamond*, mark size=3.5pt,
        color=radarPi,
        mark options={fill=radarPi,draw=black,line width=0.3pt}}
      \addlegendentry{$\pi_{0.5}$}
      \addlegendimage{only marks, mark=triangle*, mark size=3.5pt,
        color=radarN17,
        mark options={fill=radarN17,draw=black,line width=0.3pt}}
      \addlegendentry{GR00T N1.7}
      \addlegendimage{only marks, mark=square*, mark size=3pt,
        color=radarN16,
        mark options={fill=radarN16,draw=black,line width=0.3pt}}
      \addlegendentry{GR00T N1.6}
      \addlegendimage{only marks, mark=pentagon*, mark size=4pt,
        color=radarMolmo,
        mark options={fill=radarMolmo,draw=black,line width=0.5pt}}
      \addlegendentry{MolmoAct2}
    \end{axis}
  \end{tikzpicture}
  \\[0.2em]
  \begin{minipage}[t]{0.49\linewidth}
    \vspace{0pt}\centering
    \begin{tikzpicture}
      \begin{axis}[calibrationpanel,
        xlabel={Manual Eval Mean Task Progress (\%)},
        ylabel={ScaleEvals BT (Elo)},
        title={ScaleEvals $n{=}1750$ vs.\ Manual $n{=}198$ observed \\ $r{=}0.94,\,p{=}0.018$},
        xmin=20, xmax=65, ymin=1430, ymax=1620,
        xtick={20,30,40,50,60},
        ytick={1450,1500,1550,1600},
        yticklabel style={/pgf/number format/1000 sep={}},
      ]
        \addplot[black!75, line width=0.7pt, forget plot]
          coordinates {(20, 1403.98) (65, 1622.66)};
        \addplot[only marks, mark=*, mark size=3pt,
          color=radarDream, mark options={fill=radarDream,draw=black,line width=0.3pt}]
          coordinates {(54.4, 1580)};
        \addplot[only marks, mark=diamond*, mark size=3.5pt,
          color=radarPi, mark options={fill=radarPi,draw=black,line width=0.3pt}]
          coordinates {(45.0, 1539)};
        \addplot[only marks, mark=triangle*, mark size=3.5pt,
          color=radarN17, mark options={fill=radarN17,draw=black,line width=0.3pt}]
          coordinates {(43.9, 1500)};
        \addplot[only marks, mark=square*, mark size=3pt,
          color=radarN16, mark options={fill=radarN16,draw=black,line width=0.3pt}]
          coordinates {(34.3, 1453)};
        \addplot[only marks, mark=pentagon*, mark size=4pt,
          color=radarMolmo, mark options={fill=radarMolmo,draw=black,line width=0.5pt}]
          coordinates {(27.0, 1456)};
      \end{axis}
    \end{tikzpicture}
  \end{minipage}\hfill
  \begin{minipage}[t]{0.49\linewidth}
    \vspace{0pt}\centering
    \begin{tikzpicture}
      \begin{axis}[calibrationpanel,
        xlabel={Manual Eval Mean Task Progress (\%)},
        ylabel={ScaleEvals Mean Task Progress (\%)},
        title={ScaleEvals $n{=}1750$ vs.\ Manual $n{=}198$ observed \\ $r{=}0.92,\,p{=}0.029$},
        xmin=20, xmax=65, ymin=12, ymax=42,
        xtick={20,30,40,50,60},
        ytick={15,20,25,30,35,40},
      ]
        \addplot[black!75, line width=0.7pt, forget plot]
          coordinates {(20, 7.68) (65, 45.01)};
        \addplot[only marks, mark=*, mark size=3pt,
          color=radarDream, mark options={fill=radarDream,draw=black,line width=0.3pt}]
          coordinates {(54.4, 38.5)};
        \addplot[only marks, mark=diamond*, mark size=3.5pt,
          color=radarPi, mark options={fill=radarPi,draw=black,line width=0.3pt}]
          coordinates {(45.0, 30.7)};
        \addplot[only marks, mark=triangle*, mark size=3.5pt,
          color=radarN17, mark options={fill=radarN17,draw=black,line width=0.3pt}]
          coordinates {(43.9, 23.2)};
        \addplot[only marks, mark=square*, mark size=3pt,
          color=radarN16, mark options={fill=radarN16,draw=black,line width=0.3pt}]
          coordinates {(34.3, 15.5)};
        \addplot[only marks, mark=pentagon*, mark size=4pt,
          color=radarMolmo, mark options={fill=radarMolmo,draw=black,line width=0.5pt}]
          coordinates {(27.0, 17.3)};
      \end{axis}
    \end{tikzpicture}
  \end{minipage}
  \\[0.6em]
  \begin{minipage}[t]{0.49\linewidth}
    \vspace{0pt}\centering
    \begin{tikzpicture}
      \begin{axis}[calibrationpanel,
        xlabel={Manual Eval Mean Task Progress (\%)},
        ylabel={ScaleEvals BT (Elo)},
        title={ScaleEvals $n{=}2486$ vs.\ Manual $n{=}198$ observed \\ $r{=}0.93,\,p{=}0.021$},
        xmin=20, xmax=65, ymin=1430, ymax=1620,
        xtick={20,30,40,50,60},
        ytick={1450,1500,1550,1600},
        yticklabel style={/pgf/number format/1000 sep={}},
      ]
        \addplot[black!75, line width=0.7pt, forget plot]
          coordinates {(20, 1408.91) (65, 1624.08)};
        \addplot[only marks, mark=*, mark size=3pt,
          color=radarDream, mark options={fill=radarDream,draw=black,line width=0.3pt}]
          coordinates {(54.4, 1587.1)};
        \addplot[only marks, mark=diamond*, mark size=3.5pt,
          color=radarPi, mark options={fill=radarPi,draw=black,line width=0.3pt}]
          coordinates {(45.0, 1536.9)};
        \addplot[only marks, mark=triangle*, mark size=3.5pt,
          color=radarN17, mark options={fill=radarN17,draw=black,line width=0.3pt}]
          coordinates {(43.9, 1500.0)};
        \addplot[only marks, mark=square*, mark size=3pt,
          color=radarN16, mark options={fill=radarN16,draw=black,line width=0.3pt}]
          coordinates {(34.3, 1458.7)};
        \addplot[only marks, mark=pentagon*, mark size=4pt,
          color=radarMolmo, mark options={fill=radarMolmo,draw=black,line width=0.5pt}]
          coordinates {(27.0, 1462.0)};
      \end{axis}
    \end{tikzpicture}
  \end{minipage}\hfill
  \begin{minipage}[t]{0.49\linewidth}
    \vspace{0pt}\centering
    \begin{tikzpicture}
      \begin{axis}[calibrationpanel,
        xlabel={Manual Eval Mean Task Progress (\%)},
        ylabel={ScaleEvals Mean Task Progress (\%)},
        title={ScaleEvals $n{=}2486$ vs.\ Manual $n{=}198$ observed \\ $r{=}0.90,\,p{=}0.037$},
        xmin=20, xmax=65, ymin=12, ymax=42,
        xtick={20,30,40,50,60},
        ytick={15,20,25,30,35,40},
      ]
        \addplot[black!75, line width=0.7pt, forget plot]
          coordinates {(20, 8.14) (65, 44.88)};
        \addplot[only marks, mark=*, mark size=3pt,
          color=radarDream, mark options={fill=radarDream,draw=black,line width=0.3pt}]
          coordinates {(54.4, 38.5)};
        \addplot[only marks, mark=diamond*, mark size=3.5pt,
          color=radarPi, mark options={fill=radarPi,draw=black,line width=0.3pt}]
          coordinates {(45.0, 31.1)};
        \addplot[only marks, mark=triangle*, mark size=3.5pt,
          color=radarN17, mark options={fill=radarN17,draw=black,line width=0.3pt}]
          coordinates {(43.9, 23.2)};
        \addplot[only marks, mark=square*, mark size=3pt,
          color=radarN16, mark options={fill=radarN16,draw=black,line width=0.3pt}]
          coordinates {(34.3, 15.3)};
        \addplot[only marks, mark=pentagon*, mark size=4pt,
          color=radarMolmo, mark options={fill=radarMolmo,draw=black,line width=0.5pt}]
          coordinates {(27.0, 18.0)};
      \end{axis}
    \end{tikzpicture}
  \end{minipage}
  \caption{\textbf{ScaleEvals aligns with manual evaluation, at both data scales.}
  Per-policy ScaleEvals Bradley--Terry (\emph{left column}) and raw mean
  task-progress (\emph{right column}) plotted against the held-out manual
  eval suite's mean TP. The manual suite contains $n{=}198$ observed rollouts
  (40 reset-required rollouts for four policies and 38 for DreamZero),
  versus $n{=}1750$ ScaleEvals rollouts in the balanced subset
  (\emph{top row}, $350$ per policy) and $n{=}2486$ in the full corpus
  (\emph{bottom row}); ScaleEvals therefore uses roughly $8.75\times$
  and $12.4\times$ more rollouts than the manual suite, respectively.
  Pearson $r$ with exact two-tailed $p$ from the closed-form
  $t(\text{df}{=}3)$ CDF. All four cells clear $\alpha{=}0.05$; the BT
  correlation is uniformly tighter than the raw-TP correlation on both
  data slices, indicating that the soft-win Bradley--Terry aggregation
  removes some of the per-policy idiosyncratic noise present in the raw
  mean-TP. Spearman $\rho{=}0.90$ in every panel; the only rank
  disagreement with the manual suite is
  MolmoAct2~$\leftrightarrow$~GR00T~N1.6 (manual ranks N1.6 above
  MolmoAct2; ScaleEvals reverses the pair on both TP and BT).}
  \label{fig:Comprehensive-calibration-appendix}
\end{figure}

\FloatBarrier
\Needspace{15\baselineskip}
\subsection{Extended Capability Profiles}

\begin{figure}[H]
\centering

\pgfplotsset{
  radarpanel/.style={
    width=7cm, height=7cm,
    xmin=0, xmax=360, ymin=0, ymax=100,
    xtick={90.0000,38.5714,347.1429,295.7143,244.2857,192.8571,141.4286},
    xticklabel style={font=\scriptsize, align=center, anchor=center},
    ytick={20,40,60,80,100}, yticklabel=\empty, ytick distance=20,
    grid=both,
    major grid style={dashed, line width=0.3pt, draw=gray!40},
    minor grid style={line width=0.2pt, draw=gray!15},
    axis line style={draw=none}, tick style={draw=none}, axis on top,
    clip=false,
  },
  radarpanelSE/.style={
    radarpanel,
    ymax=60,
    ytick={10,20,30,40,50,60}, ytick distance=10,
  },
}
%
%
\begin{minipage}{0.5\linewidth}
\centering
\begin{tikzpicture}
\begin{polaraxis}[radarpanel,
    xticklabels={%
        transfer,%
        contact rich,%
        state flip,%
        articulation,%
        spatial,%
        precision,%
        reposing%
    },
]
\addplot+[color=radarDream, fill=radarDream, fill opacity=0.08, draw=none, mark=none, forget plot]
    coordinates {(90.0000,42.60) (38.5714,100.00) (347.1429,66.50) (295.7143,49.50) (244.2857,46.10) (192.8571,60.80) (141.4286,63.30) (90.0000,42.60)};
\addplot+[color=radarPi, fill=radarPi, fill opacity=0.08, draw=none, mark=none, forget plot]
    coordinates {(90.0000,45.10) (38.5714,100.00) (347.1429,51.80) (295.7143,41.50) (244.2857,37.70) (192.8571,22.50) (141.4286,30.00) (90.0000,45.10)};
\addplot+[color=radarN17, fill=radarN17, fill opacity=0.08, draw=none, mark=none, forget plot]
    coordinates {(90.0000,42.10) (38.5714,62.50) (347.1429,68.80) (295.7143,37.50) (244.2857,44.40) (192.8571,25.00) (141.4286,26.50) (90.0000,42.10)};
\addplot+[color=radarMolmo, fill=radarMolmo, fill opacity=0.08, draw=none, mark=none, forget plot]
    coordinates {(90.0000,24.50) (38.5714,50.00) (347.1429,33.20) (295.7143,6.20) (244.2857,23.40) (192.8571,15.00) (141.4286,10.00) (90.0000,24.50)};
\addplot+[color=radarN16, fill=radarN16, fill opacity=0.08, draw=none, mark=none, forget plot]
    coordinates {(90.0000,33.60) (38.5714,0.00) (347.1429,25.00) (295.7143,37.50) (244.2857,29.60) (192.8571,48.30) (141.4286,5.00) (90.0000,33.60)};
\addplot+[color=radarDream, line width=1.2pt, mark=*, mark size=3pt, mark options={draw=black, line width=0.5pt, fill=radarDream}]
    coordinates {(90.0000,42.60) (38.5714,100.00) (347.1429,66.50) (295.7143,49.50) (244.2857,46.10) (192.8571,60.80) (141.4286,63.30) (90.0000,42.60)};
\addplot+[color=radarPi, line width=1.2pt, mark=diamond*, mark size=3pt, mark options={draw=black, line width=0.5pt, fill=radarPi}]
    coordinates {(90.0000,45.10) (38.5714,100.00) (347.1429,51.80) (295.7143,41.50) (244.2857,37.70) (192.8571,22.50) (141.4286,30.00) (90.0000,45.10)};
\addplot+[color=radarN17, line width=1.2pt, mark=triangle*, mark size=3pt, mark options={draw=black, line width=0.5pt, fill=radarN17}]
    coordinates {(90.0000,42.10) (38.5714,62.50) (347.1429,68.80) (295.7143,37.50) (244.2857,44.40) (192.8571,25.00) (141.4286,26.50) (90.0000,42.10)};
\addplot+[color=radarMolmo, line width=1.2pt, mark=pentagon*, mark size=3pt, mark options={draw=black, line width=0.5pt, fill=radarMolmo}]
    coordinates {(90.0000,24.50) (38.5714,50.00) (347.1429,33.20) (295.7143,6.20) (244.2857,23.40) (192.8571,15.00) (141.4286,10.00) (90.0000,24.50)};
\addplot+[color=radarN16, line width=1.2pt, mark=square*, mark size=3pt, mark options={draw=black, line width=0.5pt, fill=radarN16}]
    coordinates {(90.0000,33.60) (38.5714,0.00) (347.1429,25.00) (295.7143,37.50) (244.2857,29.60) (192.8571,48.30) (141.4286,5.00) (90.0000,33.60)};
\node[anchor=south west, font=\scriptsize, gray] at (axis cs:115,20) {20\%};
\node[anchor=south west, font=\scriptsize, gray] at (axis cs:115,60) {60\%};
\node[anchor=south west, font=\scriptsize, gray] at (axis cs:115,100) {100\%};
\end{polaraxis}
\end{tikzpicture}
\subcaption{Manual eval --- held-out reset-required suite ($n{=}198$ observed; $40$ rollouts for four policies and $38$ for DreamZero).}
\label{fig:skill-radar-manual}
\end{minipage}
\\[0.8em]
%
\begin{minipage}[t]{0.495\linewidth}
\vspace{0pt}\centering
\begin{tikzpicture}
\begin{polaraxis}[radarpanelSE,
    xticklabels={%
        \shortstack{transfer\\$\eta^2{=}0.11^{***}$},%
        \shortstack{contact rich\\$\eta^2{=}0.11^{***}$},%
        \shortstack{state flip\\$\eta^2{=}0.07^{***}$},%
        \shortstack{articulation\\$\eta^2{=}0.07^{***}$},%
        \shortstack{spatial\\$\eta^2{=}0.06^{***}$},%
        \shortstack{precision\\$\eta^2{=}0.06^{**}$},%
        \shortstack{reposing\\$\eta^2{=}0.05^{**}$}%
    },
]
\addplot+[color=radarDream, fill=radarDream, fill opacity=0.08, draw=none, mark=none, forget plot]
    coordinates {(90.0000,46.92) (38.5714,44.87) (347.1429,31.08) (295.7143,38.72) (244.2857,36.36) (192.8571,29.55) (141.4286,35.38) (90.0000,46.92)};
\addplot+[color=radarPi, fill=radarPi, fill opacity=0.08, draw=none, mark=none, forget plot]
    coordinates {(90.0000,37.77) (38.5714,37.44) (347.1429,27.22) (295.7143,29.58) (244.2857,30.29) (192.8571,27.77) (141.4286,29.86) (90.0000,37.77)};
\addplot+[color=radarN17, fill=radarN17, fill opacity=0.08, draw=none, mark=none, forget plot]
    coordinates {(90.0000,22.75) (38.5714,36.98) (347.1429,18.71) (295.7143,21.34) (244.2857,22.63) (192.8571,19.91) (141.4286,32.42) (90.0000,22.75)};
\addplot+[color=radarMolmo, fill=radarMolmo, fill opacity=0.08, draw=none, mark=none, forget plot]
    coordinates {(90.0000,17.39) (38.5714,17.24) (347.1429,14.68) (295.7143,18.23) (244.2857,20.90) (192.8571,13.67) (141.4286,21.83) (90.0000,17.39)};
\addplot+[color=radarN16, fill=radarN16, fill opacity=0.08, draw=none, mark=none, forget plot]
    coordinates {(90.0000,17.47) (38.5714,14.04) (347.1429,14.31) (295.7143,17.51) (244.2857,14.43) (192.8571,17.73) (141.4286,15.98) (90.0000,17.47)};
\addplot+[color=radarDream, line width=1.2pt, mark=*, mark size=3pt, mark options={draw=black, line width=0.5pt, fill=radarDream}]
    coordinates {(90.0000,46.92) (38.5714,44.87) (347.1429,31.08) (295.7143,38.72) (244.2857,36.36) (192.8571,29.55) (141.4286,35.38) (90.0000,46.92)};
\addplot+[color=radarPi, line width=1.2pt, mark=diamond*, mark size=3pt, mark options={draw=black, line width=0.5pt, fill=radarPi}]
    coordinates {(90.0000,37.77) (38.5714,37.44) (347.1429,27.22) (295.7143,29.58) (244.2857,30.29) (192.8571,27.77) (141.4286,29.86) (90.0000,37.77)};
\addplot+[color=radarN17, line width=1.2pt, mark=triangle*, mark size=3pt, mark options={draw=black, line width=0.5pt, fill=radarN17}]
    coordinates {(90.0000,22.75) (38.5714,36.98) (347.1429,18.71) (295.7143,21.34) (244.2857,22.63) (192.8571,19.91) (141.4286,32.42) (90.0000,22.75)};
\addplot+[color=radarMolmo, line width=1.2pt, mark=pentagon*, mark size=3pt, mark options={draw=black, line width=0.5pt, fill=radarMolmo}]
    coordinates {(90.0000,17.39) (38.5714,17.24) (347.1429,14.68) (295.7143,18.23) (244.2857,20.90) (192.8571,13.67) (141.4286,21.83) (90.0000,17.39)};
\addplot+[color=radarN16, line width=1.2pt, mark=square*, mark size=3pt, mark options={draw=black, line width=0.5pt, fill=radarN16}]
    coordinates {(90.0000,17.47) (38.5714,14.04) (347.1429,14.31) (295.7143,17.51) (244.2857,14.43) (192.8571,17.73) (141.4286,15.98) (90.0000,17.47)};
\node[anchor=south west, font=\scriptsize, gray] at (axis cs:115,10) {10\%};
\node[anchor=south west, font=\scriptsize, gray] at (axis cs:115,30) {30\%};
\node[anchor=south west, font=\scriptsize, gray] at (axis cs:115,50) {50\%};
\end{polaraxis}
\end{tikzpicture}
\subcaption{ScaleEvals --- full corpus ($n{=}2486$ rollouts).}
\label{fig:skill-radar-2486}
\end{minipage}\hfill
%
\begin{minipage}[t]{0.495\linewidth}
\vspace{0pt}\centering
\begin{tikzpicture}
\begin{polaraxis}[radarpanelSE,
    xticklabels={%
        \shortstack{transfer\\$\eta^2{=}0.13^{***}$},%
        \shortstack{contact rich\\$\eta^2{=}0.13^{***}$},%
        \shortstack{state flip\\$\eta^2{=}0.07^{***}$},%
        \shortstack{articulation\\$\eta^2{=}0.07^{**}$},%
        \shortstack{spatial\\$\eta^2{=}0.08^{***}$},%
        \shortstack{precision\\$\eta^2{=}0.07^{**}$},%
        \shortstack{reposing\\$\eta^2{=}0.05$}%
    },
]
\addplot+[color=radarDream, fill=radarDream, fill opacity=0.08, draw=none, mark=none, forget plot]
    coordinates {(90.0000,46.16) (38.5714,41.76) (347.1429,32.46) (295.7143,37.64) (244.2857,36.95) (192.8571,30.31) (141.4286,32.80) (90.0000,46.16)};
\addplot+[color=radarPi, fill=radarPi, fill opacity=0.08, draw=none, mark=none, forget plot]
    coordinates {(90.0000,38.19) (38.5714,38.11) (347.1429,26.59) (295.7143,25.54) (244.2857,32.55) (192.8571,28.21) (141.4286,24.44) (90.0000,38.19)};
\addplot+[color=radarN17, fill=radarN17, fill opacity=0.08, draw=none, mark=none, forget plot]
    coordinates {(90.0000,23.53) (38.5714,31.43) (347.1429,20.52) (295.7143,24.09) (244.2857,20.38) (192.8571,22.07) (141.4286,34.33) (90.0000,23.53)};
\addplot+[color=radarMolmo, fill=radarMolmo, fill opacity=0.08, draw=none, mark=none, forget plot]
    coordinates {(90.0000,15.16) (38.5714,16.73) (347.1429,14.30) (295.7143,18.04) (244.2857,20.86) (192.8571,13.79) (141.4286,22.34) (90.0000,15.16)};
\addplot+[color=radarN16, fill=radarN16, fill opacity=0.08, draw=none, mark=none, forget plot]
    coordinates {(90.0000,17.65) (38.5714,12.77) (347.1429,14.57) (295.7143,18.37) (244.2857,14.54) (192.8571,17.31) (141.4286,17.16) (90.0000,17.65)};
\addplot+[color=radarDream, line width=1.2pt, mark=*, mark size=3pt, mark options={draw=black, line width=0.5pt, fill=radarDream}]
    coordinates {(90.0000,46.16) (38.5714,41.76) (347.1429,32.46) (295.7143,37.64) (244.2857,36.95) (192.8571,30.31) (141.4286,32.80) (90.0000,46.16)};
\addplot+[color=radarPi, line width=1.2pt, mark=diamond*, mark size=3pt, mark options={draw=black, line width=0.5pt, fill=radarPi}]
    coordinates {(90.0000,38.19) (38.5714,38.11) (347.1429,26.59) (295.7143,25.54) (244.2857,32.55) (192.8571,28.21) (141.4286,24.44) (90.0000,38.19)};
\addplot+[color=radarN17, line width=1.2pt, mark=triangle*, mark size=3pt, mark options={draw=black, line width=0.5pt, fill=radarN17}]
    coordinates {(90.0000,23.53) (38.5714,31.43) (347.1429,20.52) (295.7143,24.09) (244.2857,20.38) (192.8571,22.07) (141.4286,34.33) (90.0000,23.53)};
\addplot+[color=radarMolmo, line width=1.2pt, mark=pentagon*, mark size=3pt, mark options={draw=black, line width=0.5pt, fill=radarMolmo}]
    coordinates {(90.0000,15.16) (38.5714,16.73) (347.1429,14.30) (295.7143,18.04) (244.2857,20.86) (192.8571,13.79) (141.4286,22.34) (90.0000,15.16)};
\addplot+[color=radarN16, line width=1.2pt, mark=square*, mark size=3pt, mark options={draw=black, line width=0.5pt, fill=radarN16}]
    coordinates {(90.0000,17.65) (38.5714,12.77) (347.1429,14.57) (295.7143,18.37) (244.2857,14.54) (192.8571,17.31) (141.4286,17.16) (90.0000,17.65)};
\node[anchor=south west, font=\scriptsize, gray] at (axis cs:115,10) {10\%};
\node[anchor=south west, font=\scriptsize, gray] at (axis cs:115,30) {30\%};
\node[anchor=south west, font=\scriptsize, gray] at (axis cs:115,50) {50\%};
\end{polaraxis}
\end{tikzpicture}
\subcaption{ScaleEvals --- balanced subset ($n{=}1750$; $350$/policy).}
\label{fig:skill-radar-1750}
\end{minipage}
%
\\[0.6em]
\begin{tikzpicture}
\begin{axis}[
    hide axis, scale only axis,
    width=0pt, height=0pt,
    xmin=0, xmax=1, ymin=0, ymax=1,
    legend style={
        font=\scriptsize, draw=gray!50,
        legend columns=5,
        at={(0,0)}, anchor=center,
        inner xsep=4pt, inner ysep=3pt,
        nodes={inner sep=2pt},
        /tikz/every even column/.append style={column sep=0.9em},
    },
    legend cell align=left,
]
\addlegendimage{color=radarDream, line width=1.2pt, mark=*, mark size=3pt, mark options={draw=black, line width=0.5pt, fill=radarDream}}
\addlegendentry{DreamZero}
\addlegendimage{color=radarPi, line width=1.2pt, mark=diamond*, mark size=3pt, mark options={draw=black, line width=0.5pt, fill=radarPi}}
\addlegendentry{$\pi_{0.5}$}
\addlegendimage{color=radarN17, line width=1.2pt, mark=triangle*, mark size=3pt, mark options={draw=black, line width=0.5pt, fill=radarN17}}
\addlegendentry{GR00T N1.7}
\addlegendimage{color=radarMolmo, line width=1.2pt, mark=pentagon*, mark size=3pt, mark options={draw=black, line width=0.5pt, fill=radarMolmo}}
\addlegendentry{MolmoAct2}
\addlegendimage{color=radarN16, line width=1.2pt, mark=square*, mark size=3pt, mark options={draw=black, line width=0.5pt, fill=radarN16}}
\addlegendentry{GR00T N1.6}
\end{axis}
\end{tikzpicture}
\caption{\textbf{Skill radar across three evaluation slices.} Per-skill mean task progress (\%) for each policy, rendered as polar profiles. The manual panel (\subref{fig:skill-radar-manual}) spans a $[0,100]$\% radial axis (its per-axis-normalized scores reach 100\%; gridlines every 20\%); the two ScaleEvals panels are normalized to $[0,60]$\% to resolve the AutoGrader task-progress spread (gridlines every 10\%). Spoke positions are anchored to the (\subref{fig:skill-radar-2486}) ScaleEvals full-corpus one-way-ANOVA $\eta^{2}$ ordering and held fixed across all three panels, so polygon shapes can be compared at the same angular position. \subref{fig:skill-radar-manual}: held-out human-graded reset-required suite from the Manual Evaluation workbook (5--11 tasks per skill across all policies; \textit{contact rich} has only 1 task per cell ``slide plate to left side of table,'' so the $\{0, 50, 62.5, 100\}$ values shown are single-task outcomes). $\eta^{2}$ is omitted from \subref{fig:skill-radar-manual} because at this $n$ the per-skill ANOVA is underpowered (no skill clears BH-FDR at $\alpha{=}0.05$). ScaleEvals AutoGrader scores on the unbalanced full corpus and on the per-tag-balanced subset; spoke labels carry each panel's own $\eta^{2}$ with Benjamini--Hochberg-adjusted significance stars ($^{*}q{<}0.05$, $^{**}q{<}0.01$, $^{***}q{<}0.001$). The radars demonstrate that trajectories per skill-tag should be balanced for improved coverage of generalizability.}
\label{fig:skill-radar-three-panel}
\end{figure}

\end{document}